\documentclass[preprint, number]{elsarticle}
\usepackage{graphicx}
\usepackage{hyperref}
\usepackage{amsmath}
\usepackage{multicol}
\usepackage{amssymb}
\usepackage{caption}
\usepackage{subcaption}
\usepackage{wrapfig}
\usepackage{tabularx}
\usepackage{amsfonts}
\usepackage{dsfont}
\usepackage{dirtytalk}
\usepackage{booktabs}
\usepackage{makecell}
\usepackage{multirow}
\usepackage{xcolor}

\newcolumntype{Y}{>{\centering\arraybackslash}X}
\newcommand{\scenario}[1]{\texttt{#1}}
\newcommand{\meanstd}[2]{${#1}_{\pm #2}$}
\newcommand{\boldmeanstd}[2]{$\mathbf{{#1}_{\pm #2}}$}

\title{Continual Learning for Recurrent Neural Networks: an Empirical Evaluation}

\author[1,2]{Andrea Cossu \corref{cor1}\fnref{cor2}}
\ead{andrea.cossu@sns.it}

\author[1]{Antonio Carta \fnref{cor2}}
\ead{antonio.carta@di.unipi.it}

\author[1]{Vincenzo Lomonaco}
\ead{vincenzo.lomonaco@unipi.it}

\author[1]{Davide Bacciu}
\ead{bacciu@di.unipi.it}

\cortext[cor1]{Corresponding author}
\fntext[cor2]{Equal contribution}

\affiliation[1]{organization={University of Pisa}, addressline={Largo B. Pontecorvo, 3}, postcode={56127}, city={Pisa}, country={Italy}}
\affiliation[2]{organization={Scuola Normale Superiore}, addressline={Piazza dei Cavalieri, 7}, postcode={56126}, city={Pisa}, country={Italy}}

\begin{document}

\begin{abstract}
Learning continuously during all model lifetime is fundamental to deploy machine learning solutions robust to drifts in the data distribution. Advances in Continual Learning (CL) with recurrent neural networks could pave the way to a large number of applications where incoming data is non stationary, like natural language processing and robotics. However, the existing body of work on the topic is still fragmented, with approaches which are application-specific and whose assessment is based on heterogeneous learning protocols and datasets. In this paper, we organize the literature on CL for sequential data processing by providing a categorization of the contributions and a review of the benchmarks. We propose two new benchmarks for CL with sequential data based on existing datasets, whose characteristics resemble real-world applications. \\
We also provide a broad empirical evaluation of CL and Recurrent Neural Networks in class-incremental scenario, by testing their ability to mitigate forgetting with a number of different strategies which are not specific to sequential data processing. Our results highlight the key role played by the sequence length and the importance of a clear specification of the CL scenario.
\end{abstract}


\maketitle

\section{Introduction} \label{sec:intro}
Continual Learning (CL) refers to the ability \say{to learn over time by accommodating new knowledge while retaining previously learned experiences} \cite{parisi2019}. \\ 
Traditionally, Machine Learning (ML) models are trained with a large amount of data to solve a fixed task. The assumption is that, at test time, the model will encounter similar data. Unfortunately, real world scenarios do not satisfy this assumption. Non stationary processes produce gradual or abrupt drifts in the data distribution \cite{ditzler2015, gamaSurveyConceptDrift2014} and the models must be continuously adapted to the new changes. The objective of incrementally learn new concepts \cite{widmer1996} may result in the Catastrophic Forgetting (CF) of previous knowledge \cite{mccloskeyCatastrophicInterferenceConnectionist1989, mcclellandIntegrationNewInformation2020, grossberg1980}. CF causes a deterioration of the performance on old tasks when acquiring new knowledge and it is a direct consequence of the stability-plasticity dilemma \cite{grossberg1980, frenchCatastrophicForgettingConnectionist1999}: the need to maintain adaptive parameters both stable enough to preserve information and plastic enough to adapt to new tasks. Today, CF is considered one of the main problems faced by CL algorithms \cite{toneva2019, nguyen2019, delange2019}. \\
Machine Learning models endowed with CL capabilities would radically change the way a model is deployed, removing the need for a separate offline training phase and allowing continuous training during the entire model lifetime without forgetting previous knowledge. \\
In this paper we focus on continual learning with Recurrent Neural Networks (RNNs) \cite{schaferRecurrentNeuralNetworks2006}. These are adaptive models that can capture the input history within an internal \textit{hidden state} which is updated iteratively. These models are widely used for Sequential Data Processing (SDP) tasks \cite{gravesSequenceTransductionRecurrent2012}, where the input is a sequence of items. \\
Recurrent models are not the only solution for sequential data processing: time-delayed neural networks \cite{waibel1989} and, more recently, Transformers \cite{vaswaniAttentionAllYou2017} are feedforward neural networks able to handle sequential patterns. Transformers have not been investigated much in the context of Continual Learning (with one notable exception \cite{sun2020}), while feedforward models use a fixed context (e.g. sliding windows) to emulate temporal coherence. This may be challenging to implement in Continual Learning, since there may not be the opportunity to collect enough patterns to build the context before making a prediction. \\ 
Ultimately, recurrent models show important differences with respect to both convolutional and feedforward approaches (including Transformers). Recurrent models implement a variable number of hidden layers, due to the unrolling through the entire length of the input sequence. This is not the case for feedforward and convolutional models, whose number of layers is usually fixed a priori. Also, recurrent models leverage weight sharing across time steps. This influences the trajectory of the learning path of recurrent models when training with backpropagation through time. We hypothesize that these two differences may impact on the application of CL strategies on recurrent models. \\

Sequential data is widespread in fundamental machine learning applications such as Natural Language Processing \cite{youngRecentTrendsDeep2018}, Human Activity Recognition \cite{haresn2016,kusupatiFastGRNNFastAccurate2019}, Speech recognition \cite{gravesSequenceTransductionRecurrent2012}, and Robotics \cite{chenRecurrentNeuralNetwork2019}. All these environments are highly non stationary and represent perfect examples of real world CL applications where data comes in the form of sequences. For example, a robot that learns to walk on different terrains or to grasp objects with different shapes will receive input patterns from its sensors as time series. During its deployment, the robot will encounter novel environments/objects and will need to adapt to these novel settings continuously. It is impossible to pretrain the robot in advance on all possible environments/objects. What if, once deployed, it is necessary to adapt to a new terrain or to grasp a new object? Retraining from scratch is an expensive, unnecessary solution since most of the knowledge needed is already contained in the model. Continual Learning represents a better solution which can be combined with SDP techniques to appropriately address sequential data. Human activity recognition \cite{hasanContinuousLearningFramework2015} is another application in which sequential data processing and continual learning interact. The recognition of new classes of activities is typically performed from sequential data (e.g. videos, sensors observations). Therefore, it is fundamental to study how this setting may impact on existing CL strategies. \\
As the examples show, sequential data processing may have a strong impact on Continual Learning, both at the methodological and application level. A true continual learner should be able to deal with temporally correlated patterns, since this is the case in many real world scenarios. This advocates for a more thorough understanding of the behavior of recurrent models in Continual Learning. \\ 

Currently, most of the Continual Learning literature focuses on Computer Vision and Reinforcement Learning problems \cite{parisi2019,lesort2020}, while the study of sequential data processing remains under-documented. The few existing works introduce new approaches tailored to specific tasks and use a diverse set of learning scenarios and experimental protocols. Therefore: i) it remains unclear whether previously existing CL strategies could still work well in SDP environments and ii) the heterogeneity of learning scenarios in the literature makes it difficult to compare experimental results among different papers. In fact, specific continual settings can have different complexity or require domain-specific solutions. \\
In this paper, we provide a systematization and categorization of current literature to highlight the different characteristic of each specific CL strategy and the peculiarities of the data and the continual environment. We adopt the class-incremental scenario \cite{vandeven2018a} as a common test-bed for CL strategies. This scenario involves a stream of data where new steps gradually introduce new classes. Class-incremental scenarios allow to develop and test generic CL strategies for recurrent networks. In our experiments, we verify whether or not CL strategies which are not specifically tailored for recurrent models are still sufficient to endow them with continual learning capabilities. 
To summarize, our main contributions of this work are:
\begin{enumerate}[a)]
    \item the first comprehensive review of the literature on continual learning in RNNs, including a precise categorization of current strategies, datasets and continual environments adopted by the literature (Section \ref{sec:relatedrnn});
    \item the proposal of two novel benchmarks in the class-incremental setting: Synthetic Speech Recognition, based on speech recognition data, and Quickdraw, based on hand-drawn sketches (Section \ref{sec:benchmarks});
    \item an extensive experimental evaluation of $6$ continual learning strategies on several CL environments for sequence classification. To the best of our knowledge, this is currently the most extensive evaluation of CL strategies for recurrent models (Section \ref{sec:protocol});
    \item an experimental comparison that highlights the effect of the recurrence and sequence length on catastrophic forgetting (Section \ref{sec:results}). Also, a comparison between single-head and multi-head models (Section \ref{sec:multihead}) which justifies our choice of class-incremental scenarios.
\end{enumerate}

\section{Continual Learning framework} \label{sec:framework}
In the rest of the paper, we will refer to the notation and the Continual Learning framework defined in this section. 

\subsection{Notation and fundamental concepts}
\begin{wrapfigure}{r}{0.5\textwidth}
    \includegraphics[width=0.5\textwidth]{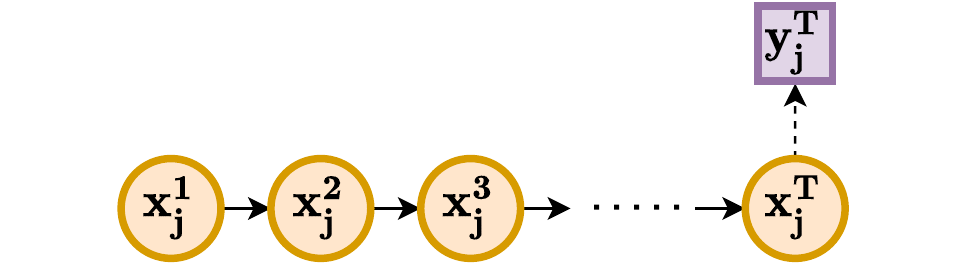}
    \caption{Sequence classification example. The input sequence $\mathbf{x_j}$ is composed by $T$ vectors (circles). The target label (square) is provided only after the last vector $\mathbf{x_j^T}$. Best viewed in color.}
    \label{fig:seqclf}
\end{wrapfigure}

For a formal definition of continual learning environments we follow the framework proposed in \cite{lesort2020}. For simplicity, here we focus on supervised continual learning on a sequence of steps $\mathcal{S} = (S_1, S_2, ...)$, where each step $S_i$ has its own data distribution $D_i$ and belongs to a specific task with label $t_i$. The patterns drawn from each $D_i$ are input-target pairs $(\mathbf{x_j}, \mathbf{y_j})$. We study sequence classification problems in which $\mathbf{y_j}$ is the target class corresponding to the input sequence $\mathbf{x_j}$. Each sequence $\mathbf{x_j}$ is composed of $T$ ordered vectors ($\mathbf{x_j^1}, ..., \mathbf{x_j^T}),\; \mathbf{x_j^i} \in \mathbb{R}^d$. In sequence classification, the target is provided only at the end of the entire sequence, that is after seeing the last vector $\mathbf{x_j^T}$. Also, the sequence length $T$ may vary from one sequence to another, depending on the specific application. Figure \ref{fig:seqclf} provides a representation of a sequence classification task. \\
\begin{figure}[t]
    \centering
    \hspace*{\fill}%
    \begin{subfigure}[t]{0.4\textwidth}
    \centering
    \includegraphics[width=\textwidth]{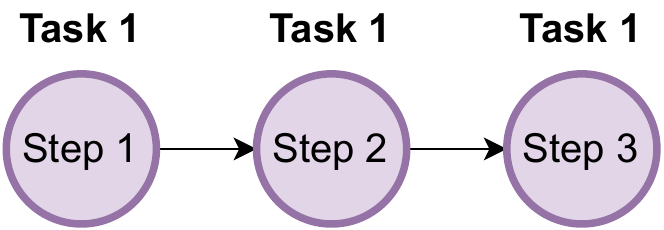}
    \caption{Single Incremental Task}
    \label{fig:sit}
    \end{subfigure}
    \hfill%
    \begin{subfigure}[t]{0.4\textwidth}
    \centering
    \includegraphics[width=\textwidth]{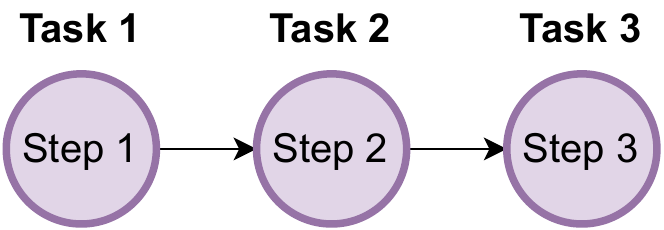}
    \caption{Multi Task}
    \label{fig:mt}
    \end{subfigure}
    \hspace*{\fill}%
    \caption{In Single Incremental Task, each step belongs to the same task label (which is the same as not providing a task label). In Multi Task, each step belongs to a different task label. Such information can be used to select different heads for different tasks both at training and test time. Best viewed in color.}
    \label{fig:mitst}
\end{figure}
At step $i$, a new batch of data becomes available. A CL algorithm $A$ executed at step $i$ leverages the current model $h_i$ to learn the new training data $\text{TR}_i$, drawn from $D_i$. The CL algorithm may also use a memory buffer $M_i$ and a task label $t_i$ associated to the training patterns. The model is then updated using the new data:
\begin{equation}
    A_i : \quad <h_{i-1}, \text{TR}_i, M_{i-1}, t_i>\ \rightarrow\ <h_i, M_i>, \quad \forall D_i \in (D_1, D_2, ...). \label{eq:cl_algo}
\end{equation}
Notice that the buffer $M_i$ and task label $t_i$ are optional and not always available. For example, it may be impossible to store previous data due to privacy concerns, while the task labels may not be available in many real-world scenarios. \\
Notice that the following definition does not pose any computational constraint to continual learning algorithms. However, we are often interested in efficient algorithms. As a results, most algorithms assume to have bounded memory and computational resources. Trivial strategies, such as retraining from scratch using $\bigcup_{i=1}^i TR_i$ as training data, are unfeasible due to their excessive computational cost and bad scaling properties with respect to the number of steps. Unfortunately, training sequentially on $\text{TR}_1, \hdots, TR_n$ will suffer from catastrophic forgetting.\\
A \emph{Continual Learning scenario} defines the properties of the data stream, such as the distinction between different tasks and the properties of new data samples. The work of \cite{maltoni2019} and, subsequently, \cite{lesort2020} introduces a classification for continual learning scenarios according to two properties: the task label sequence and the content of incoming data. The authors distinguished between \emph{Multi Task} (\scenario{MT}) scenarios, where each incoming task is different from the previous ones, \emph{Single Incremental Task} (\scenario{SIT}) in which the task is always the same, and \emph{Multi Incremental Task} (\scenario{MIT}) in which new and old tasks may be interleaved and presented again to the model. 
Each task, identified by its task label, can be composed by one or more steps. Figure \ref{fig:mitst} compares SIT and MT scenarios with respect to the task label information.
In addition, data from new steps may provide \emph{New Classes} (\scenario{NC}), \emph{New Instances} (\scenario{NI}) of previously seen classes or a mix of both, that is \emph{New Instances and Classes} (\scenario{NIC}). A previous categorization, proposed in \cite{vandeven2018a}, focused on three main scenarios: Task incremental, Domain incremental and Class incremental which can find their place in the framework of \cite{maltoni2019, lesort2020}. Table \ref{tab:scenarios} summarizes and compare these different classifications. \\
Task-incremental (\scenario{MT, MIT}) scenarios assume the availability of task labels for each sample. Unfortunately, in many real-world applications it is difficult to obtain explicit task labels for each sample. While sometimes it may be possible to label the data for training, at test time task labels are often not available. This distinction is fundamental since the availability of task labels simplifies the learning problem (see Section \ref{sec:multihead}). For example, in multi-task scenarios it is possible to use a \textit{multi-head} model, in which each task has its own output layer (head). At test time, the model uses the task label to choose the appropriate head and to compute the output.\\

\begin{wrapfigure}{r}{0.5\textwidth}
    \includegraphics[width=0.5\textwidth]{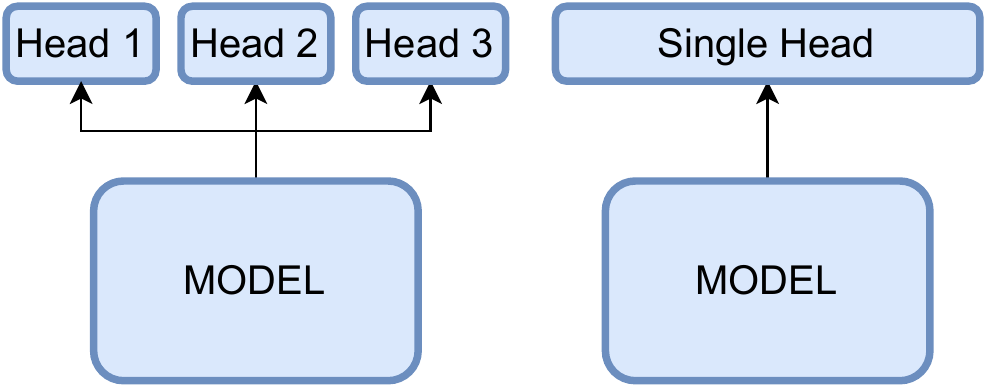}
    \caption{Multi-head model (left) and single-head model (right). In the multi-head model, the output layer allocates a different set of output units (head) for each task. The single-head model use the same output layer for each task.  Best viewed in color.}
    \label{fig:multisinglehead}
\end{wrapfigure}

Most CL systems deployed in the real world operate without task labels, and therefore it is important to develop algorithms that do not require them. Single incremental task scenarios, where the task label is always the same, require to use \textit{single-head} output layers \cite{farquhar2019}. Single-head models do not partition the output space, thus they are more susceptible to catastrophic forgetting (see Section \ref{sec:multihead} for an experimental comparison). Even without the availability of task label, a model could still leverage a multi-head output layer, but then it would have to infer at test time which head to use \cite{cossu2020}. Figure \ref{fig:multisinglehead} shows the difference between multi and single headed models. \\

\begin{table}[t]
    \centering
    \begin{tabular}{c|c|c|c}
        \toprule
         & \thead{\scenario{Multi Task}} & \thead{\scenario{Single}\\ \scenario{Incremental Task}} & \thead{\scenario{Multi} \\\scenario{Incremental Task}}  \\ \midrule
         \scenario{New Classes} & \makecell{Task\\incremental} & \makecell{Class\\incremental} & \\
         \scenario{New Instances} & --- & \makecell{Domain\\incremental} & \\
         \makecell{\scenario{New Instances}\\\scenario{and Classes}} & --- & & \\
         \bottomrule
    \end{tabular}
    \caption{Comparison of different CL scenarios classification. The table highlights the fact that the three scenarios for CL introduced by \cite{vandeven2018a} are not the only possible ones in CL. In fact, the scenario classification in \cite{maltoni2019} provides $7$ scenarios. Dashes indicate unrealistic scenarios: \scenario{Multi Task} is only compatible with \scenario{New Classes}, since new instances would belong to previously encountered tasks. Empty cells indicate scenarios which could occur but are not included in the classification of \cite{vandeven2018a}.}
    \label{tab:scenarios}
\end{table}

The classification proposed above assumes task boundaries are exact and provided by the environment. Task-free or \scenario{online} CL \cite{schlimmerIncrementalLearningNoisy1986, harriesExtractingHiddenContext1998, aljundi2019d, zeno2018} removes such assumption by focusing on a stream of data in which a task label is not provided (\scenario{SIT}) and examples are seen one (or a very small number) at a time. Incremental or online learning (see \cite{ditzler2015} for a review) shares similarities with continual learning. However, online learning usually focuses less on catastrophic forgetting and more on fast learning of new data.  \\ 
In this scenario, old knowledge must be preserved to help with the new step. However, the old steps are not necessarily revisited. This settings is common to many time series applications, such as stock price prediction, or weather and energy consumption prediction \cite{ditzler2015}. \\

\subsection{Taxonomy of Continual Learning Strategies} \label{sec:related}
Continual learning strategies are traditionally divided into three main families \cite{parisi2019}: regularization strategies, architectural strategies and replay (or rehearsal) strategies. \\

\textbf{Regularization strategies} balance plasticity and stability by adding a regularization term to the loss function. For example, Elastic Weight Consolidation (EWC) \cite{kirkpatrick2017} and Synaptic Intelligence \cite{zenke2017} estimate the parameters importance and foster the model stability by penalizing large changes in important parameters. Importance values can be updated after each step, as in EWC and its more efficient version \cite{schwarz2018}, or online after each minibatch, as in Synaptic Intelligence, Memory Aware Synapses \cite{aljundi2018} and EWC++ \cite{chaudhry2018}. \\
The effectiveness of importance-based regularization approaches has been questioned in \cite{lesort2020b} for class incremental scenarios. The authors showed that this family of strategies suffers from complete forgetting. \\
Learning without Forgetting (LwF) \cite{li2016} is a regularization strategy which is not based on importance values. LwF retains the knowledge of previous steps by using knowledge distillation \cite{hintonDistillingKnowledgeNeural2015} to encourage the current model to behave similarly to the models trained on the previous steps. \\
\textbf{Architectural strategies} increment the model plasticity with dynamic modifications to its architecture, for example by adding new components \cite{rusu2016, yoon2018}. Forgetting may be mitigated by freezing previous components \cite{rusu2016}, by reorganizing the architecture via pruning \cite{hung2019} or by using sparse connections \cite{sokar2020}. Due to the very general idea behind this family of strategies, it is quite challenging to organize the existing body of works. Nonetheless, most members of the architectural family share some commonalities: an increasing computational cost with the number of steps due to the incremental model expansion and the need to choose a technique to prevent forgetting on the old model components. \\
\textbf{Replay strategies} leverage an external memory to store previous examples by selecting them randomly or with more ad-hoc criteria \cite{rolnick2019, vandeven2020, aljundi2019b}. Since the amount of storage required for the memory could grow without bounds, several approaches use generative replay to avoid the need to explicitly store patterns \cite{vandeven2018, wang2019}. A generative model replaces the memory, thus bounding the memory cost to the storage of the generative model weights. However, catastrophic forgetting must be still taken into consideration when training the generative model. \\

The three aforementioned families of CL strategies do not capture the entirety of the proposals in CL.
\textbf{Bayesian approaches} are increasingly used to mitigate CF \cite{zeno2018, nguyen2018, li2019, farquhar2018} and they have been successfully combined with regularization \cite{ebrahimi2020, ahn2019}, architectural approaches \cite{mehta2020} and replay \cite{kurle2020}. Bayesian methods in CL consider an initial distribution on the model parameters and iteratively approximate its posterior when new tasks arrive. The approximation usually includes components able to mitigate forgetting of previous tasks: as an example, the learning trajectory can be controlled by the uncertainty associated to each model parameter, computed from its probability distribution. If a parameter has a large uncertainty value, it is deemed not to be relevant for the current task. Therefore, to protect important parameters and mitigate forgetting, the allowed magnitude in parameter change is set to be proportional to its uncertainty \cite{ahn2019, ebrahimi2020}. \\ 
Another line of research explores the use of \textbf{Sparse distributed representations} \cite{ahmad2016}. Machine learning models often produce dense, entangled activations, where a slight change in one parameter may affect the entire representation, causing catastrophic forgetting. In contrast, sparse distributed representations naturally create independent substructures which do not affect each other and are less prone to CF \cite{frenchUsingSemiDistributedRepresentations1991}. As a result, methods that encourage sparsity produce disentangled representations which do not interfere with each other \cite{golkar2019, ororbia2019a, coop2012, aljundi2019c}. The main problem is that sparsity by design is not sufficient: the model has to learn \textit{how} to use sparse connections during training. \\
Since one of the objectives of CL is to quickly learn new information, it seems natural to combine CL and \textbf{Meta Learning} \cite{hospedales2020}. There are already efforts in this direction \cite{beaulieu2020, javed2019, caccia2020}, with a focus on online and task-free scenarios \cite{finn2019, he2019, harrison2019}. 
Finally, CL has also been studied through the lens of graph-structured data: expressing similarities in the input patterns through graphs may help in mitigating forgetting \cite{tang2020}. Also, deep graph networks have been successfully combined with CL approaches on a number of class-incremental graph benchmarks \cite{carta2021}.

In the remainder of this section, we present in more details the methods that we decided to use in our experiments. We focus on regularization and replay approaches for recurrent networks. Architectural approaches are a promising avenue of research but they cannot be easily adapted to different model architectures. As a result, none of them has emerged as a standard in the literature. Therefore, we decided to focus on model-agnostic methods that can be easily adapted to a wide range of models and CL scenarios.\\

\paragraph{Elastic Weight Consolidation}
Elastic Weight Consolidation (EWC) \cite{kirkpatrick2017} is one of the most popular CL strategies. It belongs to the regularization family and it is based on the estimation of parameters importance. At the end of each step, EWC computes an importance vector $\mathbf{\Omega_n}^\mathbf{\Theta}$ for parameters $\mathbf{\Theta}$. The importance vector is computed on the training set $\mathcal{D}$ at step $n$ by approximating the diagonal elements of the Fisher Information Matrix:
\begin{equation} \label{eq:fisher}
\begin{split} 
\mathbf{\Omega_n}^\mathbf{\Theta} &= \mathbb{E}_{(\mathbf{x}, \mathbf{y}) \in \mathcal{D}} \big[\text{diag}((\nabla_{\mathbf{\Theta}} \log p_\mathbf{\Theta}(\mathbf{y}|\mathbf{x}))\:(\nabla_{\mathbf{\Theta}} \log p_\mathbf{\Theta}(\mathbf{y}|\mathbf{x}))^T)\big] \\
&= \mathbb{E}_{(\mathbf{x}, \mathbf{y}) \in \mathcal{D}} \big[(\nabla_{\mathbf{\Theta}} \log p_\mathbf{\Theta}(\mathbf{y}|\mathbf{x}))^2 \big],
\end{split}
\end{equation}
where $p_\theta$ is the output of the model parameterized by $\mathbf{\Theta}$ and $(\mathbf{x},\mathbf{y})$ is the input-target pair.\\
By strictly following Eq. \ref{eq:fisher}, the expectation operator would require to compute the squared gradient on each data sample and then average the result. Since this procedure may be quite slow, a minibatch approach is usually taken. No major difference is experienced between these two versions. \\
During training, the loss function is augmented with a regularization term $\mathcal{R}$ which keeps important parameters close to their previous value:
\begin{equation} \label{eq:ewc}
    \mathcal{R}(\mathbf{\Theta}, \mathbf{\Omega}) = \lambda \sum_{t=1}^{n-1} \sum_{\theta \in \mathbf{\Theta}} \mathbf{\Omega_t}^\theta (\theta_t - \theta_n)^2.
\end{equation}
The hyperparameter $\lambda$ controls the amount of regularization. 

\paragraph{Memory Aware Synapses}
Memory Aware Synapses (MAS) \cite{aljundi2018} is similar to EWC since it is an importance-based regularization method. Unlike EWC, MAS computes importances online in an unsupervised manner. Therefore, MAS keeps only a single importance vector $\mathbf{\Omega}^\mathbf{\Theta}$ and updates it with a running average computed after each pattern:
\begin{equation} \label{eq:mas}
\mathbf{\Omega}^\mathbf{\Theta_{N+1}} = \frac{N \mathbf{\Omega}^\mathbf{\Theta_N} + \big|\nabla_\mathbf{\Theta}{\| p_\mathbf{\Theta}(\mathbf{x_k})\|_2^2} \big|}{N+1} ,
\end{equation}
where $N$ indexes the number of patterns seen so far. To make the update faster, the running average in Eq. \ref{eq:mas} can be computed after each minibatch. The penalization during training is the same of EWC (Eq. \ref{eq:ewc}). 

\paragraph{Learning without Forgetting}
Learning without Forgetting (LwF) \cite{li2016} is a regularization approach based on knowledge distillation \cite{hintonDistillingKnowledgeNeural2015}. At the end of each step, a copy of the current model is saved. During training, the previous model produces its outputs on the current step input. The learning signal for the current model is regularized by a function measuring the distance between the current model output and the previous model output. The objective is to keep the two output distributions close to each other. Formally, when training on step $t$, the total loss $L_t$ is the sum of the classification loss CE (e.g. cross entropy) and the distillation penalization:
\begin{equation}
\begin{split}
    L_t(\mathbf{x_t}, \mathbf{y_t}; \mathbf{\Theta_t}, \mathbf{\Theta_{t-1}}) =&\ \text{CE}_\mathbf{\Theta_t}(\mathbf{x_t}, \mathbf{y_t})\ + \\ +&\ \lambda\ \text{KL}[p_\mathbf{\Theta_t}(\mathbf{x_t}) || p_\mathbf{\Theta_{t-1}}(\mathbf{x_t})],
\end{split}
\end{equation}
where $\text{KL}[p\ ||\ q]$ is the KL-divergence between $p$ and $q$.\\
The softmax temperature $T$ used in the final layer of the previous model can be tuned to control the prediction confidence. 

\paragraph{Gradient Episodic Memory}
Gradient Episodic Memory (GEM) \cite{lopez-paz2017} is a CL strategy which mitigates forgetting by projecting the gradient update along a direction which does not interfere with previously acquired knowledge. The average gradient for each step is computed using a small buffer of samples collected from previous steps. The steps gradients are used to define a set of constraints which cannot be violated by the update direction. Finding the optimal direction requires solving a quadratic programming problem:
\begin{align}
    & \min_z \frac{1}{2}\ \| g - z \|_2^2 \\
    & \text{subject to} \quad Gz \geq \gamma,
\end{align}
where $G$ is a matrix in which each column is the gradient computed step-wise on the memory samples (one column per previous step). The vector $g$ is the gradient update and $\gamma$ is the margin within which the constraints must be respected. GEM can work also in \scenario{online} scenarios \cite{lopez-paz2017}.

\paragraph{Averaged Gradient Episodic Memory}
Average-GEM (A-GEM) \cite{chaudhry2019} is a more efficient version of GEM where the quadratic program solution is approximated over a random sample from the memory, thus removing the need to compute a constraint for each previous step. In this way the constraints may be broken for some step, resulting in worse performance. \\
The projected gradient $\hat{g}$ is computed by
\begin{equation}
    \hat{g} = g - \frac{g^T g_{ref}}{g_{ref}^T g_{ref}} g_{ref},
\end{equation}
where $g_{ref}$ is the gradient computed on the random memory sample and $g$ is the proposed gradient update. \\
A-GEM has also been tested in \scenario{online} scenarios \cite{chaudhry2019}.

\paragraph{Replay}
Replay strategies store a subset of input patterns from previous steps in a separate replay memory. To support the generality of our experiments, in this paper we selected patterns at random from the training set, without using any specific selection procedure. The memory accepts up to $K$ patterns per class. During training, each minibatch is augmented with $P$ patterns per previous class, selected at random from the replay memory. We keep $P$ to very small values in order to make the approach reasonable in a real CL environment.\\

The CL strategies presented above belong to different families and have different characteristics. In order to choose which strategy is more suitable to a specific application, one has to understand their advantages and limitations. Regularization strategies like EWC, LwF and MAS are very efficient since they operate without keeping a memory of previous pattern. However, their performance usually deteriorates when observing a large number of steps \cite{lesort2020b}. In contrast, Replay remains effective even on long streams of steps \cite{hayes2018}, even if the storage of previous patterns may be unfeasible for applications with strict memory bounds. Finally, GEM and A-GEM share the storage problems of replay, since they keep a buffer of previous samples. However, their performance is usually superior to the regularization strategies in class-incremental settings \cite{lopez-paz2017}.

\section{Continual Learning with Recurrent Models} \label{sec:relatedrnn}
The literature on continual learning focuses on feedforward and convolutional models, with experiments in Computer Vision and, to a lesser extent, Reinforcement Learning applications. Recently, there has been a growing interest towards continual learning applications with sequential data. In this section, we provide the first comprehensive review of the literature on recurrent neural networks in continual learning settings. In addition, we describe the main datasets and benchmarks for sequential data processing used in Continual Learning.

\subsection{Survey of Continual Learning in Recurrent Models}
\begin{table}[t]
    \centering
    \begin{tabular}{cc|cccc}
        \toprule
         & Paper    & \shortstack{Deep\\RNN} & \shortstack{Application} & \shortstack{CL\\scenario} & \shortstack{Large \\ Comparison} \\ \midrule
        \multirow{6}{*}{\rotatebox{90}{seminal}}    
         & \cite{ring1997}      & $\times$   & - & - & $\times$ \\ 
         & \cite{french1997a}   & $\times$   & - & \scenario{SIT} & $\times$ \\
         & \cite{ans2002}       & $\times$   & - & \scenario{SIT} & $\times$ \\
         & \cite{ans2004}       & $\times$   & - & \scenario{SIT} & $\times$ \\
         & \cite{coop2013}      & $\times$   & - & \scenario{SIT} & $\times$ \\
         \midrule
         \multirow{6}{*}{\rotatebox{90}{\shortstack{NLP}}}    
         & \cite{asghar2019}    & \checkmark & domain adaptation  & \scenario{SIT} & $\times$ \\
         & \cite{li2020b}       & \checkmark & instruction learning, NMT   & \scenario{SIT}  &   $\times$  \\
         & \cite{wolf2018}      & \checkmark & language modeling & - & $\times$  \\
         & \cite{kruszewski2020a} & $\times$ & language modeling & \scenario{Online} & $\times$ \\
         & \cite{madasu2020}    & \checkmark & sentiment analysis & - & $\times$ \\
         & \cite{thompson2019a} & \checkmark & NMT  & - & $\times$ \\
         \midrule
        \multirow{5}{*}{\rotatebox{90}{bio-inspired}}    
         & \cite{cui2016}       & $\times$   & - & \scenario{Online} & $\times$ \\
         & \cite{ororbia2020}   & $\times$   & - & \scenario{Online, SIT} & $\times$ \\
         & \cite{parisi2018}    & $\times$   & vision   & - & $\times$ \\
         & \cite{kobayashi2019} & $\times$   & motor control  & \scenario{MT} & $\times$ \\
         & \cite{ororbia2019}   & $\times$   & -    & \scenario{Online} & $\times$  \\
         \midrule
        \multirow{6}{*}{\rotatebox{90}{deep learning}}    
         & \cite{xue2019}        & \checkmark & ASR    & - & $\times$ \\
         & \cite{sodhani2019}   & \checkmark & -            & - & $\times$  \\
         & \cite{schak2019}     & \checkmark & synthetic    & - & $\times$  \\
         & \cite{cossu2020}     & \checkmark & generic      & \scenario{SIT} & $\times$ \\
         & \cite{duncker2020}   & \checkmark & neuroscience & - & $\times$ \\
         & \cite{ehret2020}     & \checkmark & -            & \scenario{MT} & \checkmark \\
         & this work            & \checkmark & -            & \scenario{SIT} & \checkmark \\
         \bottomrule
    \end{tabular}
    \caption{Overview of the literature based on our categorization. \textit{Deep RNN} refers to the use of a learning model attributable to the family of deep recurrent networks (e.g. LSTM). Application-agnostic papers have a dash in the \textit{application} column. A dash in the \textit{CL} scenario indicates that the paper does not provide its clear indication. \textit{Large comparison} refers to experiments with at least 3 CL baselines on two different application domains.}
    \label{tab:sotarnn}
\end{table}

We propose a new classification in which each paper is assigned to one of $4$ macro-areas: seminal work, natural language processing applications, bio-inspired and alternative recurrent models, and deep learning models. \\
Table \ref{tab:sotarnn} provides a fine-grained description of our review. For each paper we highlight $5$ different properties: the group to which the paper belongs to, the use of popular deep learning architectures, the application domain, the type of CL scenarios and if the paper provides a large experimental comparison. Table \ref{tab:sotarnn} shows that we are the only one to provide a large scale evaluation of recurrent models in \scenario{SIT+NC} scenario. The details of our categorization are discussed in the following.

\paragraph{Seminal Work}
Historically, interest in CL and Sequential Data Processing traces back to \cite{ring1997} and \cite{french1997a}. The CHILD model \cite{ring1997} represents one of the first attempts to deal with sequential data in reinforcement learning environments like mazes. Although CHILD lacks an explicit recurrent structure, the author discussed about the possibility to include a more powerful memory component into the model (i.e. a recurrent neural network) to address tasks with longer temporal dependencies. However, the author also recognizes the difficulties in this process, due to the challenge of learning very long sequences. Moreover, the fixed structure of a RNN compared with a growing model like CHILD highlighted the need for approaches based on model expansion.\\
In the same period, French introduced the pseudo-recurrent connectionist network \cite{french1997a, french1997}, a model which makes use of pseudo replay \cite{robins1995} (replay based on random, but fixed, input patterns), but did not address sequential data processing tasks. Later on, the pseudo recurrent network together with pseudo-replay inspired the Reverberating Simple Recurrent Network (RSRN) \cite{ans2002, ans2004}. This is a dual model composed by two auto-associative recurrent networks which exchange information by means of pseudo patterns. In the \textit{awake state}, the performance network learns a new pattern through backpropagation. The storage network generates a random pseudo pattern and presents it to the performance network, interleaved with the real one. Once the loss falls below a certain threshold, the system enters the \textit{sleep state}, in which the performance network generates a pseudo pattern and the storage network learns from it. In this way, pseudo patterns produced in the awake state carry information about previous sequences to the performance network, while pseudo patterns produced in the sleep state carry information about the recently acquired knowledge to the storage network. The authors showed the presence of forgetting on toy sequences and the beneficial effect of pseudo patterns in mitigating it. However, from their experiments it is not possible to assess the effectiveness of RSRN on more realistic benchmarks. \\
From the very beginning, sparsity has been a recurring theme in Continual Learning \cite{frenchCatastrophicForgettingConnectionist1999, french1991}. The Fixed Expansion Layer \cite{coop2012} introduced the use of a large, sparse layer to disentangle the model activations. Subsequently, the Fixed Expansion Layer has been applied to recurrent models \cite{coop2013}. However, in order to build the sparse layer in an optimal way, the model requires to solve a quadratic optimization problem (feature-sign search algorithm) which can be problematic in real world problems (as we discuss in Section \ref{sec:results}). 

\paragraph{Natural Language Processing}
Natural Language Processing is becoming one of the main test-beds for continual and online settings in sequential data processing \cite{biesialska2020a}. Most proposals in this area used modern recurrent architectures and focus on specific problems and strategies. Moreover, they rarely compare against existing CL techniques. It is therefore challenging to draw general conclusions on recurrent networks in CL from this kind of experiments. Examples of applications are online learning of language models where new words are added incrementally \cite{li2020b, wolf2018, kruszewski2020a}, continual learning in neural machine translation on multiple languages \cite{thompson2019a} and sentiment analysis on multiple domains \cite{madasu2020}.\\
The use of attention mechanisms \cite{bahdanauNeuralMachineTranslation2014}, now widespread in NLP, may provide approaches which are widely applicable, since there is no assumptions on the type of information to attend to. As an example, the Progressive Memory Banks \cite{asghar2019} augments a recurrent neural network with an external memory. The memory grows in time to accommodate for incoming information, while at the same time freezing previous memory cells successfully prevents forgetting. In addition, the author showed that finetuning old cells, instead of freezing them, increases forward transfer on future data. Their experiments are executed on incremental domain adaptation tasks, where the distribution of words shifts as new domains are introduced.

\paragraph{Bio-inspired and Alternative Recurrent Models}
There are a number of contributions that propose customized architectures to address continual learning problems and mitigate catastrophic forgetting. Spiking models for CL, such as the Hierarchical Temporal Memory \cite{cui2016} and the Spiking Neural Coding Network \cite{ororbia2020}, are naturally designed to tackle sequential data processing tasks. More importantly, they adopt learning algorithms which allow to control more efficiently the stability-plasticity trade-off. While promising, these approaches are still missing a thorough empirical validation on real world CL benchmarks. \\
Echo State Networks \cite{lukoseviciusReservoirComputingApproaches2009} are RNNs whose recurrent component, called \textit{reservoir}, is not trained. Therefore, they are appealing for CL since the reservoir cannot suffer from forgetting. One of the drawbacks may be the fact that the static connections must be able to learn different tasks without adapting their values. The work in \cite{kobayashi2019} tries to overcome the problem by employing a fractal reservoir combined with an external task vector (based on the task label) representing the current task. Different reservoir chunks process different patterns based on their associated task vector. While the final performance on different motor commands for reinforcement learning environments validated the approach, the requirement of multi task scenarios limits its applicability. \\
The Backpropagation Through Time (BPTT) is the most used algorithm to train recurrent networks. The Parallel Temporal Neural Coding Network \cite{ororbia2019} introduced a new learning algorithm which is less susceptible to forgetting than BPTT and other variants in sequential benchmarks like MNIST and language modelling. \\
Temporal information may also arrive from videos. This is particularly important since it allows to exploit the vast literature on CL and Computer Vision. However, it is also possible to develop specific solutions, as it has been done with recurrent Self Organizing Maps (SOM)  \cite{parisi2018}. The authors incorporate temporal information into the recurrent SOM and perform object detection from short videos with CORe50 dataset \cite{lomonaco2017}.\\

\paragraph{Deep Learning Models}
Recently, there have been a number of papers that studied CL applications in sequential domains using recurrent architectures widely used in the deep learning literature, such as Elman RNNs \cite{elmanFindingStructureTime1990} and LSTMs \cite{hochreiterLongShortTermMemory1997}. The advantage of this generic approach is that it can be easily adapted to specialized models to solve any sequential problem. As expected, vanilla recurrent models such as Elman RNNs and LSTMs suffer from catastrophic forgetting in CL scenarios \cite{sodhani2019,schak2019}. \\
The combination of existing CL strategies, like Gradient Episodic Memory \cite{lopez-paz2017} and Net2Net \cite{chen2016}, with RNNs has already showed promising results \cite{sodhani2019}. Contrary to vanilla LSTM networks, their model was able to mitigate forgetting in three simple benchmarks. This important result supports the need for an extensive evaluation of RNNs and CL strategies not specifically tailored to sequential data processing problems. \\
Recurrent networks are also inclined to be combined with architectural strategies, since most of them are model agnostic. The idea behind Progressive networks \cite{rusu2016} has been applied to recurrent models \cite{cossu2020} and also improved by removing the need for task labels at test time with a separate set of LSTM autoencoders, able to recognize the distribution from which the pattern is coming from. The resulting model is multi-headed but it is able to automatically select the appropriate head at test time in class-incremental scenarios. Hypernetworks (previously used for CL in \cite{oswaldContinualLearningHypernetworks2019}) are able to mitigate forgetting when combined with RNNs \cite{ehret2020}: this work was the first to provide an extensive comparison of traditional CL techniques in several sequential domains. Differently from this paper, they use multi-task scenarios with a multi-head (see Section \ref{sec:multihead} for a comparison between single-head and multi-head models). Preserving the space spanned by the connections from being corrupted by the weight update appears to be beneficial also to CL \cite{duncker2020}. Finance \cite{philps2019} and Automatic Speech Recognition \cite{xue2019} applications have been explored as candidate application domains for online and continual learning strategies. 

\subsection{Sequential Data Processing Datasets for Continual Learning} \label{sec:relateddataset}

\begin{table}[t]
  \centering
  \begin{tabular}{lcr}
    \toprule
    Dataset & Application & Scenario \\ 
    \midrule
    Copy Task \cite{sodhani2019, ehret2020} & synthetic & \scenario{MT+NI} \\ 
    Delay/Memory Pro/Anti \cite{duncker2020} & synthetic, neuroscience & \scenario{MT+NI} \\
    Seq. Stroke MNIST \cite{sodhani2019, ehret2020} & stroke classification & \scenario{SIT+(NI/NC)} \\
    Quick, Draw! $\dagger$ & stroke classification & \scenario{SIT+NC} \\
    MNIST-like \cite{cossu2020} \cite{coop2013} $\dagger$ & object classification & \scenario{SIT+(NI/NC)} \\
    CORe50 \cite{parisi2018} & object recognition & \scenario{SIT+(NI/NC)} \\ \hline
    MNLI \cite{asghar2019}& domain adaptation & \scenario{SIT+NI} \\
    MDSD \cite{madasu2020} & sentiment analysis & \scenario{SIT+NI} \\
    WMT17 \cite{bojar2017} & NMT & \scenario{MT+NC} \\
    OpenSubtitles18 \cite{lison2018} & NMT & \scenario{MT+NC} \\ 
    WIPO COPPA-V2 \cite{mjd2016coppa} \cite{thompson2019a} & NMT & \scenario{MT+NC} \\ 
    CALM \cite{kruszewski2020a} & language modeling & \scenario{Online} \\
    WikiText-2 \cite{wolf2018} & language modeling & \scenario{SIT+NI/NC} \\ \hline
    Audioset \cite{cossu2020, ehret2020} & sound classification & \scenario{SIT+NC} \\
    LibriSpeech, Switchboard \cite{xue2019} & speech recognition & \scenario{(SIT/MT)+NC}  \\
    Synthetic Speech Commands $\dagger$ & sound classification & \scenario{SIT+NC} \\ \hline
    Acrobot \cite{kobayashi2019} & reinforcement learning & \scenario{MT+NI} \\
    \bottomrule
  \end{tabular}
  \caption{Datasets used in continual learning for sequential data processing. The \textit{scenario} column indicates in which scenario the dataset has been used (or could be used when the related paper does not specify this information). Datasets used on this paper are marked with $\dagger$. }
  \label{tab:sotadatasets}
\end{table}

Due to the different application domains and different research communities interested in continual learning for SDP domains, there are no standard benchmarks used to evaluate CL strategies. Existing benchmarks vary greatly in terms of complexity. Furthermore, different application domains use a slightly different language. In this section, we provide a review of the different datasets and continual learning scenarios, following the classification described in Table \ref{tab:scenarios}. We believe that this review can favor the cross-pollination between classic CL techniques and sequential domains and between different sequential domains. 

Table \ref{tab:sotadatasets} provides an overview of the different datasets in literature. For each dataset we highlight previous work that used the datasets, the application domain of the data, and the CL scenario according to Table \ref{tab:scenarios}. Clearly, most datasets in literature are used by few, or even just one, paper. This is due to the different research questions each paper tries to answer: since few works are spread over very different areas, it is natural to find different benchmarks and evaluation protocols. Unfortunately, these differences in the experimental setups make it impossible to compare different models in the literature or to deduce general and task-independent conclusions.

Different \textbf{synthetic benchmarks} have been adapted to continual scenarios. The \emph{Copy Task} \cite{gravesNeuralTuringMachines2014}, a benchmark used to test the short-term memory of recurrent models, incrementally increases the sequence lengths in the continual setting (\scenario{SIT+NI}). However, the data generating distribution remains constant, which means that the drift between the different steps is limited. \emph{Pixel-MNIST} is another popular benchmark for RNN models \cite{leSimpleWayInitialize2015} where MNIST digits are presented one pixel at a time, either with the original order or using a fixed permutation. Continual learning scenarios based on pixel-MNIST include new classes (\scenario{SIT+NC} in \cite{cossu2020} and this paper or \scenario{MT+NC} in \cite{ehretContinualLearningRecurrent2020}) or new permutations (\scenario{SIT+NI} in this work). \emph{Sequential Stroke MNIST} \cite{dejongIncrementalSequenceLearning2016} represents MNIST digits \cite{lecunGradientBasedLearningApplied1998} as a sequence of pen strokes, with pen displacements as features (plus pen up/down bit). The dataset is adapted to a continual data stream by increasing the sequence length (\scenario{SIT+NI}) or by creating new classes (\scenario{SIT+NC} in \cite{cossu2020} and this paper, or \scenario{MT+NC} in \cite{ehretContinualLearningRecurrent2020}). More realistic CL benchmarks for computer vision, like CORe50 \cite{lomonaco2017}, can be used also in sequential contexts to leverage temporal correlated information from videos. 

In the \textbf{Natural Language Processing} domain a common scenario is the domain-incremental setting, where new instances from different topics are gradually introduced (\scenario{SIT+NI}). Examples of applications are natural language inference \cite{williamsBroadCoverageChallengeCorpus2018,asghar2019}, sentiment analysis \cite{madasu2020} and machine translation in a \scenario{MT+NC} scenario \cite{thompson2019a}. Alternatively, \cite{wolf2018} studies the problem of \scenario{online} learning a language model. In this work, EWC is used to keep the previous knowledge when the recurrent model is trained with a single sequence. CALM is a benchmark specifically designed for \scenario{online} CL in language modeling \cite{kruszewski2020a} which provides a realistic environment in which to test NLP models.

The most common CL scenario in the \textbf{audio signal processing} domain is the incremental classification (\scenario{MT/SIT+NC}), where new classes are gradually introduced \cite{cossu2020}. For example, \emph{AudioSet} \cite{gemmekeAudioSetOntology2017} is a dataset of annotated audio events. The raw audio signals are already preprocessed, generating sequences of $10$ timesteps with $128$ features each. Differently from Sequential Stroke MNIST or Copy Task, AudioSet is a real-world application. However, the small sequence length may conceal possible problems when working with RNNs. Other datasets in the literature refer to specific applications, like reinforcement learning \cite{kobayashi2019}, or neuroscience problems \cite{duncker2020}.

\section{Evaluating Continual Learning Strategies}
In our review, we showed that literature on CL strategies for SDP domains is still in its early stages. The use of heterogeneous benchmarks and of different jargon is detrimental to the comparison of the various approaches in literature. Most of the works focus on customized strategies, often ignoring popular techniques widely employed in other domains where CL is more mature, e.g. computer vision. Furthermore, benchmarks have large variations in terms of complexity. It is for these reasons that we believe of fundamental importance to put forward a solid and articulated evaluation of existing CL strategies on RNN architectures. To this end, we design an experimental setup which is easily reproducible and general enough to be used with all recurrent architectures.

In this section, we focus on two main points related to our evaluation protocol: the choice of the CL scenarios and the application-agnostic nature of the benchmarks.

\subsection{Defining the Continual Learning Scenario}
We rely on class-incremental classification (\scenario{SIT+NC}) of sequences as a standard and generic scenario to test novel CL techniques for recurrent models. Incremental classification is a challenging task - at least in the \scenario{SIT} scenario without explicit task labels - and it is directly applicable to many real-world problems, such as incremental audio classification.  \\
We deliberately avoid the use of Multitask scenarios (\scenario{MT, MIT}), in which the task label is explicitly provided for each pattern, both at training and test time. Although this assumption may be reasonable in very specific applications (e.g. machine translation), in the vast majority of the cases such information is not available. We also experimented with a popular benchmark in the \textit{Domain incremental} scenario (\scenario{SIT+NI}). This represents a simpler CL problem than class incremental. Still, it is compatible with our setup since it does not use knowledge about task labels.

\subsection{Benchmarks} \label{sec:benchmarks}
\begin{figure}[t]
    \centering
    \begin{subfigure}[t]{0.43\textwidth}
    \centering
    \includegraphics[width=\textwidth]{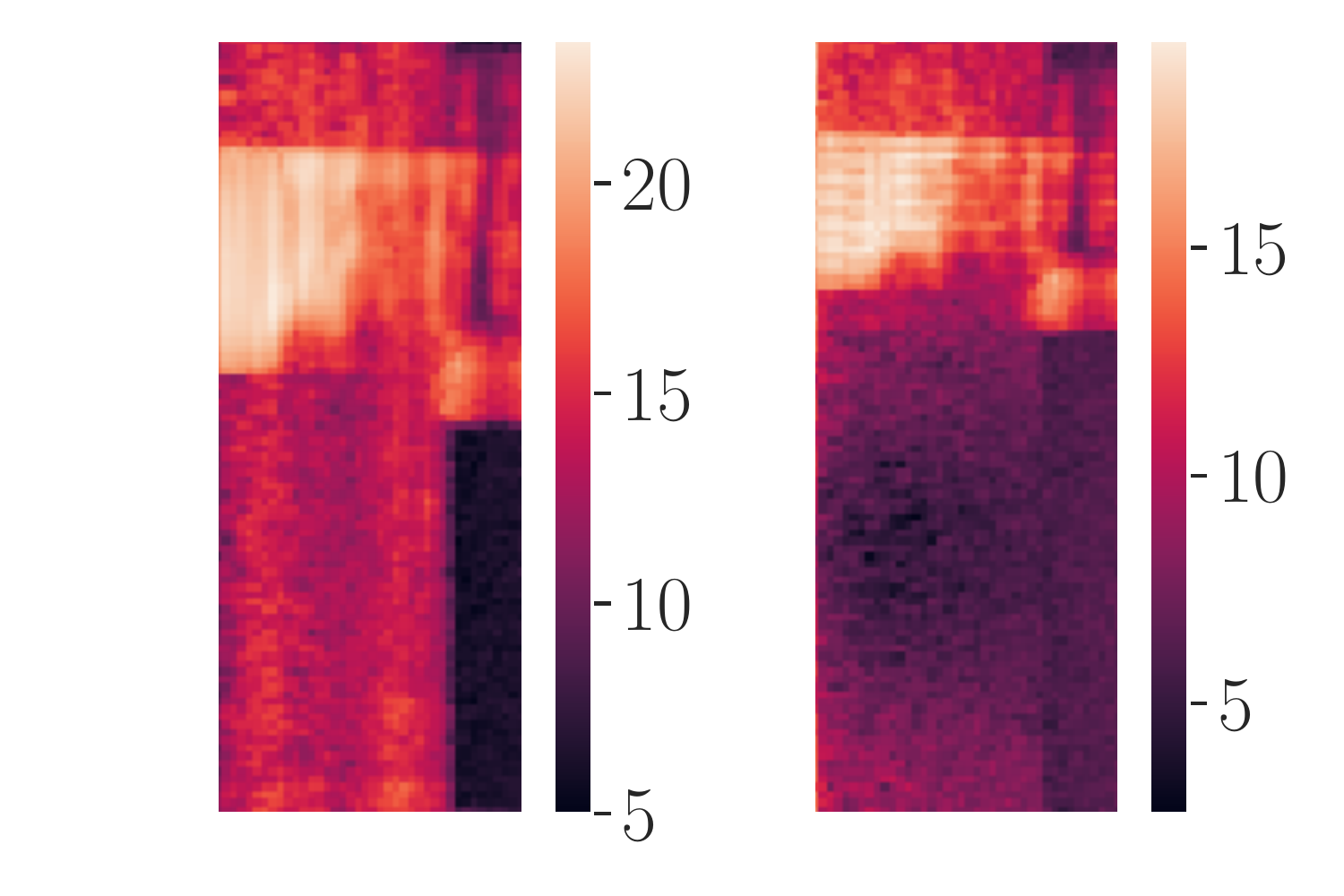}
    \caption{SSC}
    \label{fig:ssc-bench}
    \end{subfigure}
    \begin{subfigure}[t]{0.55\textwidth}
    \centering
    \includegraphics[width=\textwidth]{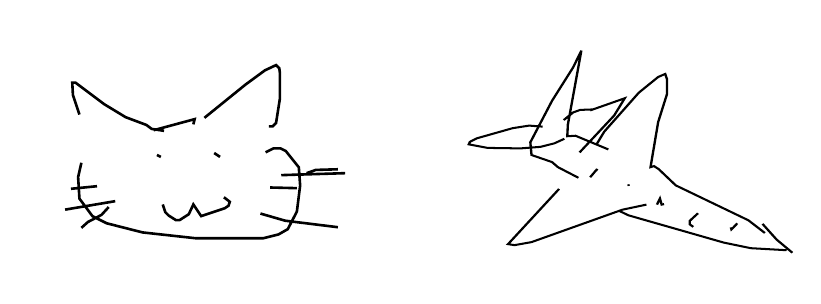}
    \caption{Quick, Draw!}
    \label{fig:quickdrawbench}
    \end{subfigure}
    \begin{subfigure}[t]{0.48\textwidth}
    \centering
    \includegraphics[width=\textwidth]{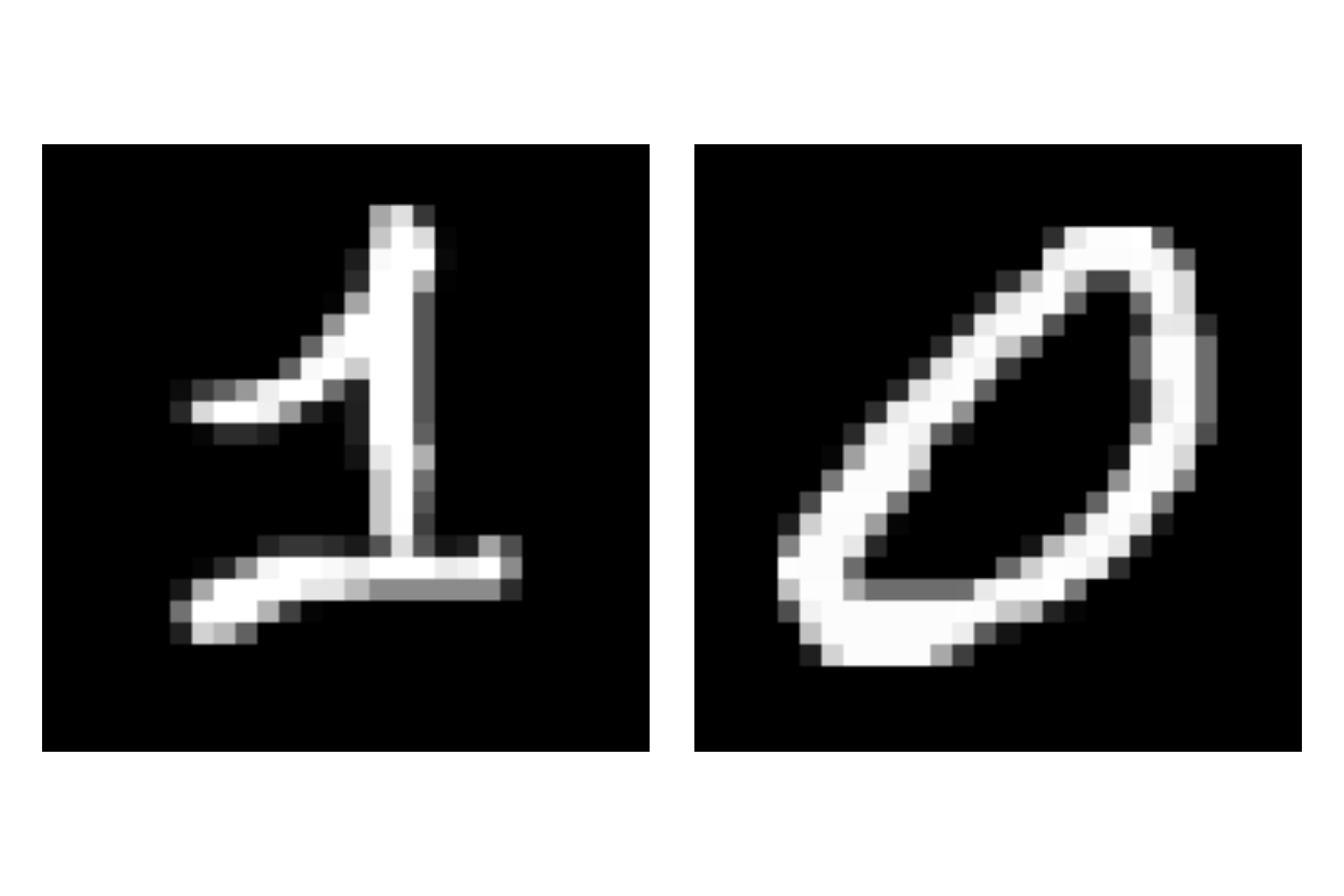}
    \caption{Row MNIST}
    \label{fig:rowmnist}
    \end{subfigure}
    \begin{subfigure}[t]{0.48\textwidth}
    \centering
    \includegraphics[width=\textwidth]{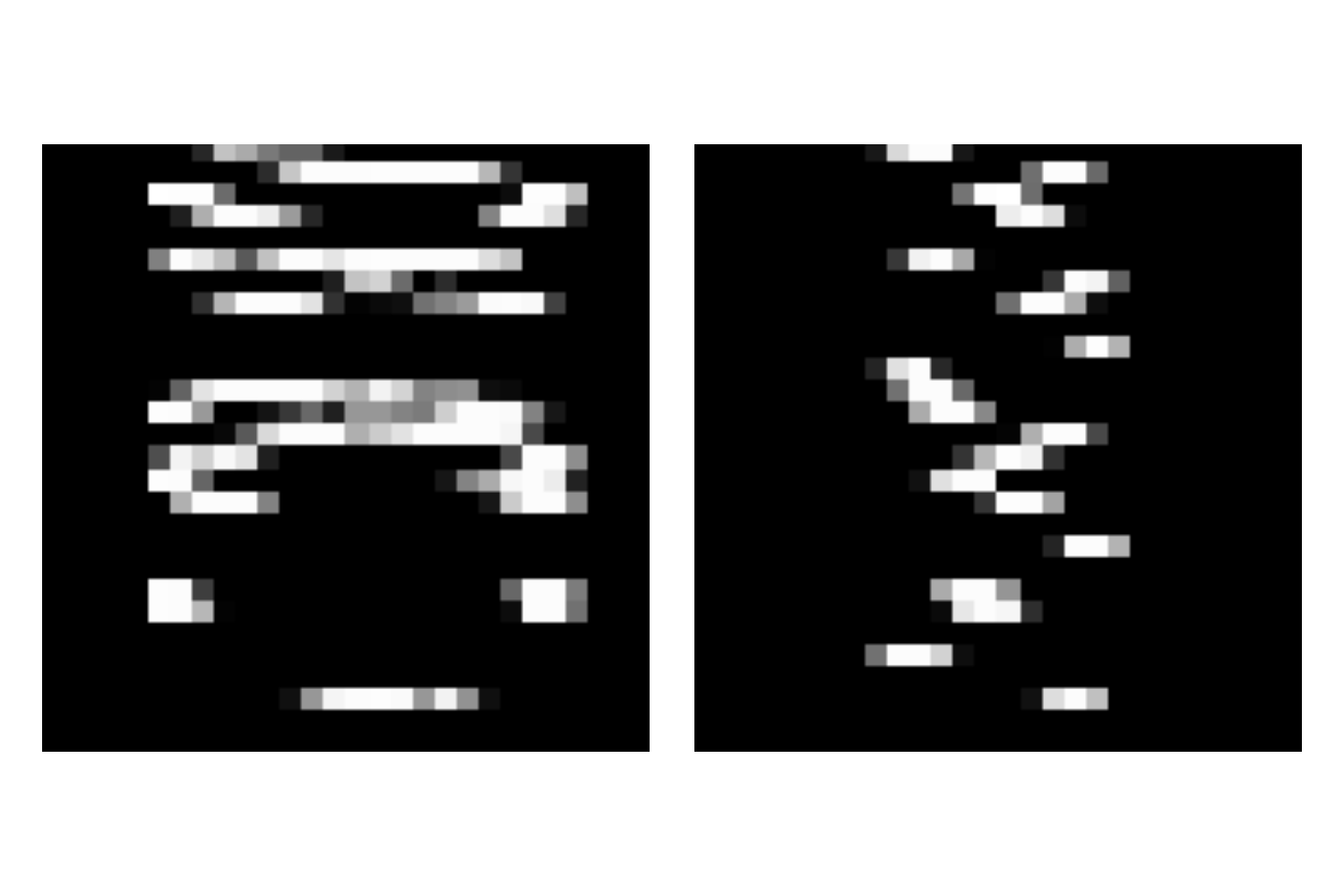}
    \caption{Permuted Row MNIST}
    \label{fig:permrowmnist}
    \end{subfigure}
    \caption{Graphical representation of two patterns for each benchmark used in the experiments. Fig. \ref{fig:ssc-bench} SSC uses $40$ Mel features (columns) for each of the $101$ steps (rows). Both plots use log values. Fig. \ref{fig:quickdrawbench} provides sketches of images (cat and airplane). Fig. \ref{fig:rowmnist} shows MNIST digits, which are provided to the model one row at a time in the Row MNIST version. Permuted MNIST (Fig. \ref{fig:permrowmnist}) permutes images row-wise. Best viewed in color.}
    \label{fig:benchmarks}
\end{figure}

\begin{figure}
    \centering
    \includegraphics[width=0.6\textwidth]{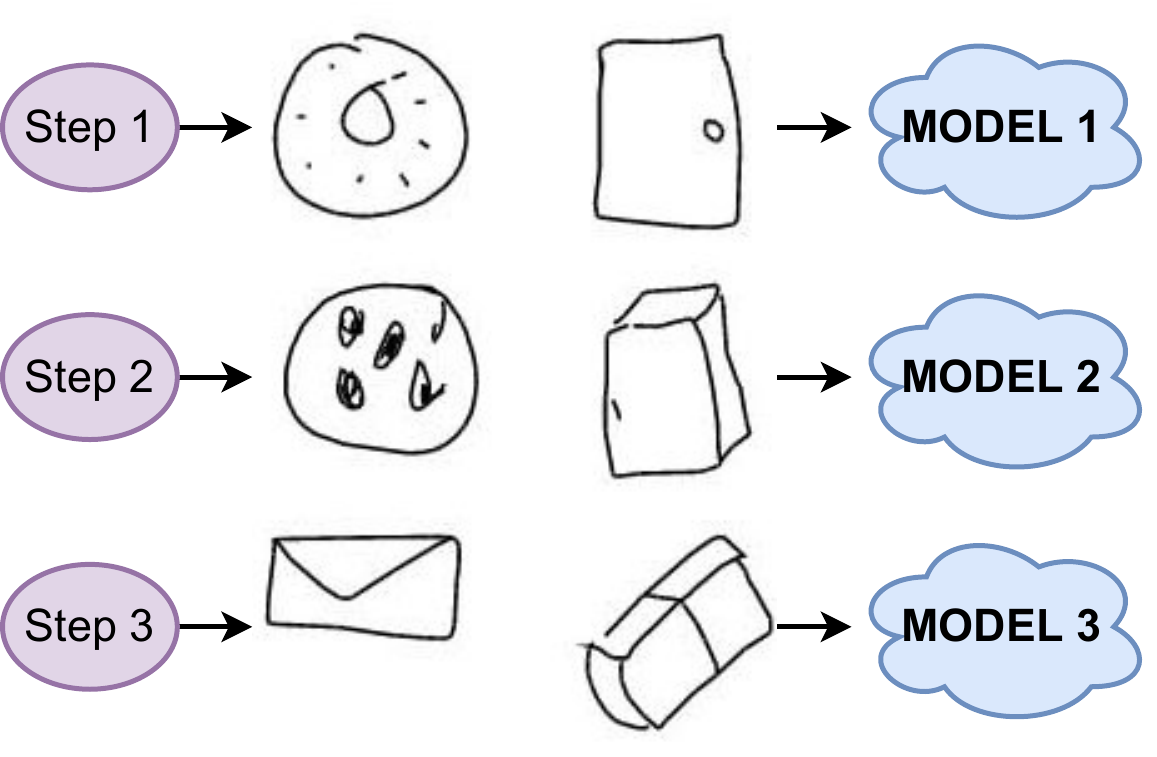}
    \caption{Class-incremental benchmark with $3$ steps starting from the patterns of Quick, Draw! dataset. Each step is composed by $2$ classes.}
    \label{fig:qd-bench}
\end{figure}

We believe there is a necessity for generic CL benchmarks that measure basic CL capabilities in sequential domains. This paper contributes to this goal by adapting the Synthetic Speech Commands dataset \cite{speechcommands} and Quick, Draw! dataset \cite{ha2018} to the incremental classification scenario. These are two tasks which are quite easy in an offline setting (especially the Synthetic Speech Commands) but become challenging in a continual setting, making them ideal as a benchmark. Moreover, they are suitable to evaluate recurrent models since they have a sequential nature: feedforward counterparts, although applicable, are more expensive in terms of adaptive parameters since there is no weight sharing and they do not incorporate any kind of temporal knowledge. In addition to the data above, our experimental protocol also includes the Permuted MNIST and Split MNIST datasets, since they are two of the most used benchmarks in the papers surveyed in our review. Table \ref{tab:bench} summarizes the main characteristics of the benchmarks used in the experiments, while Figure \ref{fig:benchmarks} provides a graphical representation of their patterns. Figure \ref{fig:qd-bench} highlights how our class-incremental benchmarks were created starting from the dataset classes.

\paragraph{Synthetic Speech Commands (SSC)} 
SSC is a dataset of audio recordings of spoken words \cite{speechcommands}. Each audio represents a single word, taken from a vocabulary of $30$ words, sampled at 16kHz. We preprocessed each sample to extract $40$ Mel coefficients using a sliding window length of $25$ ms, with $10$ ms stride. The resulting sequences have a fixed length of $101$ timesteps. We did not apply any further normalization. We create a Class Incremental scenario (\scenario{SIT+NC}) from SSC by learning two classes at a time, for a total number of $10$ steps. For model selection and assessment, we used a total of $16$ classes out of the $30$ available ones. \\ 
Additional details on the use of SSC in the experiments are reported in \ref{app:ssc}. 

\paragraph{Quick, Draw! (QD)}
QD is a dataset of hand-drawn sketches, grouped into $345$ classes \cite{ha2018}. Each drawing is a sequence of pen strokes, where each stroke has $3$ features: $x$ and $y$ displacements with respect to the previous stroke and a bit indicating if the pen has been lifted. We adopted a Class-incremental scenario (\scenario{SIT+NC}) by taking $2$ classes at a time for a total number of $10$ steps. For model selection and assessment, we used a total of $16$ classes out of the $345$ available ones. \ref{app:quickdraw} provides details about the classes used in the experiments. Differently from the previous datasets Quick, Draw! sequences have variable length. Since this is often the case in real-world applications, it is important to assess the performance of recurrent models in this configuration.

\paragraph{Permuted MNIST (PMNIST)} 
PMNIST is heavily used to train recurrent neural networks on long sequences \cite{leSimpleWayInitialize2015}. Each MNIST image is treated as a sequence of pixels and shuffled according to a fixed permutation. Each step of the data stream uses a different permutation to preprocess the images and uses all the $10$ MNIST classes. This allows to easily create an unlimited number of steps. This configuration corresponds to a Domain incremental scenario (\scenario{SIT+NI}). 

\paragraph{Split MNIST (SMNIST)} 
SMNIST is another popular continual adpatation of sequential MNIST. In this scenario, MNIST sequences are taken two classes at a time. First, all 0s and 1s are fed to the model. Then, all 2s and 3s. And so on and so forth up to the last pair (8 and 9). This scenario consists of only $5$ steps. The configuration corresponds to a Class Incremental benchmark (\scenario{SIT+NC}). 

\begin{table}[t]
    \centering
    \begin{tabular}{c|c|c|c|c}
    \toprule
    & SSC & QD & SMNIST & PMNIST \\
    \midrule
    CL Steps & 10 & 10 & 5 & 10 \\
    Classes per step & 2 & 2 & 2 & 2 \\
    Sequence length & 101 & variable (8-211) & 28/196 & 28/196 \\
    Total number of classes & 30 & 345 & 10 & 10 \\
    Input size & 40 & 3 & 28/4 & 28/4 \\
    \bottomrule
    \end{tabular}
    \caption{Benchmarks used in the experimental evaluation. QuickD, Draw has patterns with \textit{variable} sequence length, from a minimum of $8$ steps to a maximum of $211$ steps}. Experiments with SMNIST and PMNIST have been conducted with different sequence lengths by taking $28$ or $4$ pixels at a time, resulting in sequences of length $28$ and $196$, respectively.
    \label{tab:bench}
\end{table}

\section{An Experimental Protocol for CL in SDP Domains} \label{sec:protocol}
In the following section we describe the experimental settings that we propose to robustly assess CL strategies within SDP tasks and that are used in the experiments discussed in the following\footnote{We publish the code and configuration files needed to reproduce our experiments at this link \url{https://github.com/AndreaCossu/ContinualLearning_RecurrentNetworks}}.

\paragraph{Model Selection} 
Hyperparameter tuning is a critical phase when building solid machine learning based predictors. This process is made considerably more difficult in CL scenarios, 
since we cannot assume to have a separate validation set comprising the entire stream of data in advance. As a result, it is not possible to build the standard train-validation-split used in model selection. To solve this problem, we follow the model selection procedure for CL introduced in \cite{chaudhry2019}. We separate the data stream into validation and test steps. In particular, we hold out the first $3$ steps from our data stream and use them to perform model selection. After selecting the best hyperparameters, we use the remainder of the stream (the test stream, composed of $10$ unseen steps in our experiments), to perform model assessment. Since Split MNIST has a limited number of steps ($5$), we decided to perform model selection and model assessment on the same data stream for this particular benchmark. \ref{app:conf} reports the parameters used in the model selection procedure and the selected configurations for both feedforward and recurrent models.

\paragraph{Data Preprocessing}
Preprocessing a dataset in a CL setting suffers from the same limitations highlighted for the model selection phase. Computing global statistics such as the mean and variance require the entire stream, which is not available beforehand. As a result, we cannot operate any kind of normalization that needs access to a global statistic, since do not have access to the entire stream in advance. For MNIST data we simply divided the pixel value for $255$, which does not require any prior knowledge about the data stream (only about the limits of the data generating device).

\paragraph{Models} 
The reference recurrent model that we use in our experiments is the popular Long Short-Term Memory network (LSTM) \cite{hochreiterLongShortTermMemory1997} (PyTorch implementation \cite{paszkePyTorchImperativeStyle2019}). LSTMs are among the most used recurrent models \cite{chungEmpiricalEvaluationGated2014} and their performance is comparable to other alternatives, like Gated Recurrent Units \cite{choPropertiesNeuralMachine2014}. Therefore, we decided to use LSTM as general representatives of recurrent architectures.\\
To provide a means of comparison between feedforward and recurrent models, we also consider and experiment with Multi Layer Perceptrons (MLP) with ReLU activations. Notice that the feedforward models take as input the entire sequence, therefore, they can be viewed as recurrent networks which receive sequences of length $1$. This alternative view over these models is useful since we experimented with multiple sequence lengths. On PMNIST and SMNIST, we ran experiments with LSTM which takes as input $4$ pixels at a time (LSTM-4), $16$ pixels at a time (LSTM-16) and $28$ pixels at a time (ROW-LSTM).

\paragraph{Performance Metrics} 
In order to assess the performance of the continual learning strategies we chose to focus on catastrophic forgetting by measuring the average accuracy over all steps at the end of training on last step $T$. This metric, called ACC \cite{lopez-paz2017}, can been formalized as:
\begin{equation}
    ACC = \frac{1}{T} \sum_{t=1}^T R_{T,t},
\end{equation}
where $R_{T,t}$ is the accuracy on step $t$ after training on step $T$. In \ref{app:times} we also compare execution times in order to assess the computational efficiency of different strategies.
\paragraph{Strategies} \label{sec:strategies}
We compare the performances of our models in combination with $6$ CL strategies: EWC \cite{kirkpatrick2017}, MAS \cite{aljundi2018}, LwF \cite{li2016}, GEM \cite{lopez-paz2017}, A-GEM \cite{chaudhry2019} and Replay.\\
In addition to the continual learning strategies, we provide two baselines. \textit{Naive} trains the model on the sequence without applying any measure against forgetting. It is useful to provides a lower-bound to any reasonable CL strategy. \textit{Joint Training} trains the model by considering the concatenation of the full stream of sequential tasks in a single batch\footnote{Joint Training is often used as an upper bound performance for CL. However, in the presence of a strong positive forward and backward transfer this may not be true.}. Notice that this is not a CL strategy since it assumes access to the entire data stream. Joint Training can be used to highlight the effect of the continuous training on the model performance.

\section{Results} \label{sec:results}

\begin{table}[t]
\centering
\begin{tabular}{lccc}
    \toprule
     \textbf{PMNIST} & MLP  & ROW-LSTM & LSTM-4 \\
     \midrule
     EWC    & \meanstd{0.92}{0.02} & \meanstd{0.58}{0.09} & \meanstd{0.29}{0.05} \\ 
     LWF    & \meanstd{0.82}{0.03} & \meanstd{0.77}{0.05} & \meanstd{0.35}{0.04} \\
     MAS    & \meanstd{0.58}{0.04} & \meanstd{0.64}{0.07} & \meanstd{0.31}{0.04}  \\
     GEM    & \boldmeanstd{0.94}{0.01} & \meanstd{0.90}{0.01} & \meanstd{0.71}{0.02} \\
     A-GEM  & \meanstd{0.80}{0.03} & \meanstd{0.58}{0.08} & \meanstd{0.20}{0.03} \\
     REPLAY & \meanstd{0.92}{0.00} & \boldmeanstd{0.92}{0.01} & \boldmeanstd{0.82}{0.01}  \\
     \midrule
     NAIVE          & \meanstd{0.31}{0.04} & \meanstd{0.61}{0.06} & \meanstd{0.30}{0.03}  \\
     Joint Training & \meanstd{0.97}{0.00} & \meanstd{0.96}{0.00} & \meanstd{0.94}{0.01} \\
     \bottomrule
     & & &                        \\
     \toprule \textbf{SMNIST} & MLP  & ROW-LSTM & LSTM-4  \\
     \midrule
     EWC    & \meanstd{0.21}{0.01} & \meanstd{0.21}{0.02} & \meanstd{0.19}{0.00} \\
     LWF    & \meanstd{0.70}{0.02} & \meanstd{0.31}{0.07} & \meanstd{0.26}{0.06} \\
     GEM    & \boldmeanstd{0.91}{0.01} & \boldmeanstd{0.93}{0.03} & \boldmeanstd{0.66}{0.04} \\
     A-GEM  & \meanstd{0.20}{0.00} & \meanstd{0.20}{0.00} & \meanstd{0.13}{0.02} \\
     MAS    & \meanstd{0.20}{0.00} & \meanstd{0.20}{0.00} & \meanstd{0.19}{0.00} \\
     REPLAY & \meanstd{0.82}{0.02} & \meanstd{0.89}{0.01} & \meanstd{0.61}{0.06} \\
     \midrule
     NAIVE          & \meanstd{0.20}{0.00} & \meanstd{0.20}{0.00} & \meanstd{0.19}{0.00}  \\
     Joint Training & \meanstd{0.95}{0.00} & \meanstd{0.97}{0.00} & \meanstd{0.95}{0.01} \\
     \bottomrule
\end{tabular}
\caption{Mean ACC and standard deviation over $5$ runs on PMNIST and SMNIST benchmarks.}
\label{tab:resultsmnist}
\end{table}

\begin{table}[t]
\centering
\begin{tabular}{cccc}
\toprule
\textbf{SSC}& MLP  & LSTM     \\
\midrule
 EWC & \meanstd{0.10}{0.00} & \meanstd{0.10}{0.00} \\
 LWF & \meanstd{0.05}{0.00} & \meanstd{0.12}{0.01} \\
 MAS & \meanstd{0.10}{0.00} & \meanstd{0.10}{0.00} \\
 GEM & \meanstd{0.55}{0.00} & \meanstd{0.53}{0.01} \\
 A-GEM & \meanstd{0.05}{0.00} & \meanstd{0.09}{0.01} \\
 REPLAY & \boldmeanstd{0.81}{0.03} & \boldmeanstd{0.73}{0.04} \\ \midrule
 NAIVE & \meanstd{0.10}{0.00} & \meanstd{0.10}{0.00} \\
 Joint Training & \meanstd{0.93}{0.00} & \meanstd{0.89}{0.02} \\   
 \bottomrule
 & & & \\
 \toprule
\textbf{QD} && LSTM \\
\midrule
EWC && \meanstd{0.12}{0.02} \\
LWF && \meanstd{0.12}{0.01} \\
MAS &&  \meanstd{0.10}{0.00} \\
GEM && \meanstd{0.47}{0.03} \\
A-GEM && \meanstd{0.10}{0.00} \\
REPLAY && \boldmeanstd{0.49}{0.02} \\ \midrule
NAIVE && \meanstd{0.10}{0.00} \\
Joint Training && \meanstd{0.96}{0.00} \\
\bottomrule
\end{tabular}
\caption{Mean ACC and standard deviation over $5$ runs on Synthetic Speech Commands and Quick, Draw! benchmarks.}
\label{tab:resultssdp}
\end{table}

\begin{figure}[t]
    \centering
    \begin{subfigure}[t]{0.52\textwidth}
    \centering
    \includegraphics[width=\textwidth]{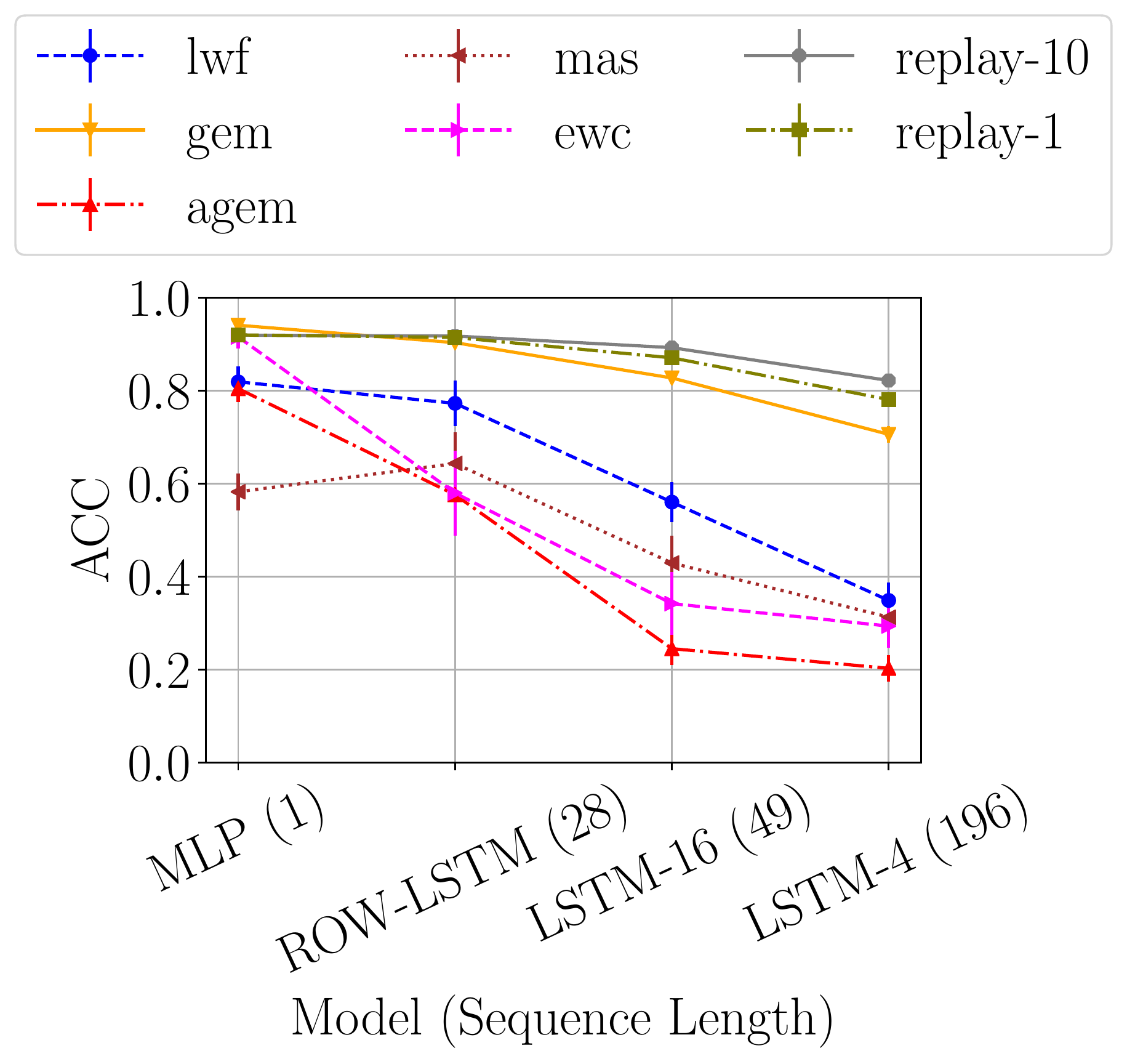}
    \caption{Permuted MNIST}
    \label{fig:seqlengthpmnist}
    \end{subfigure}
    \begin{subfigure}[t]{0.46\textwidth}
    \centering
    \includegraphics[width=\textwidth]{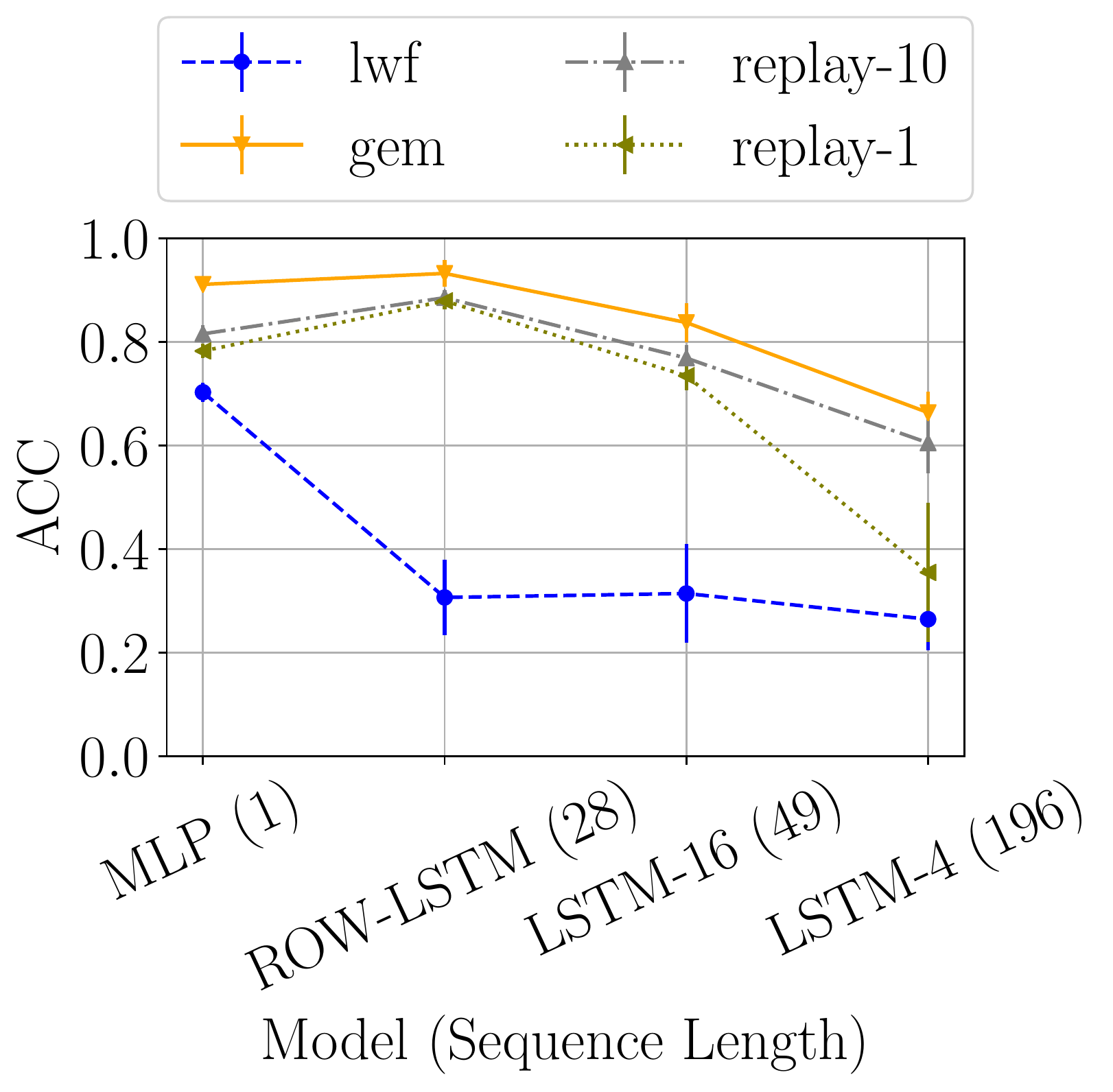}
    \caption{Split MNIST}
    \label{fig:seqlengthsmnist}
    \end{subfigure}
    \caption{Average ACC on all steps for different sequence lengths and different CL strategies. Sequence length causes a decrease in performances among all strategies. Best viewed in color.}
    \label{fig:seqlength}
\end{figure}

\begin{figure}[t]
    \centering
    \begin{subfigure}[t]{0.4\textwidth}
    \centering
    \includegraphics[width=\textwidth]{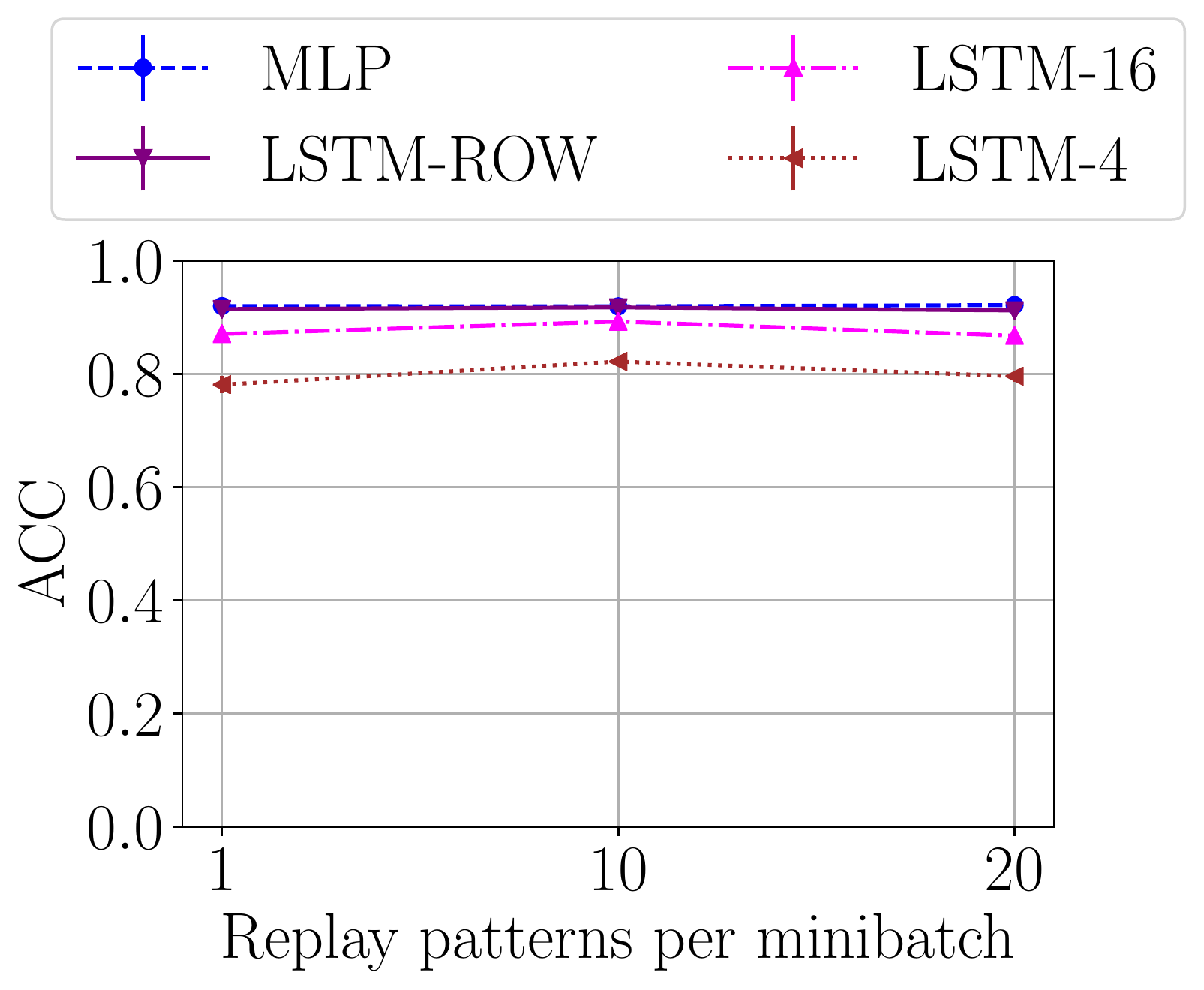}
    \caption{Permuted MNIST}
    \label{fig:replaypmnist}
    \end{subfigure}
    \begin{subfigure}[t]{0.4\textwidth}
    \centering
    \includegraphics[width=\textwidth]{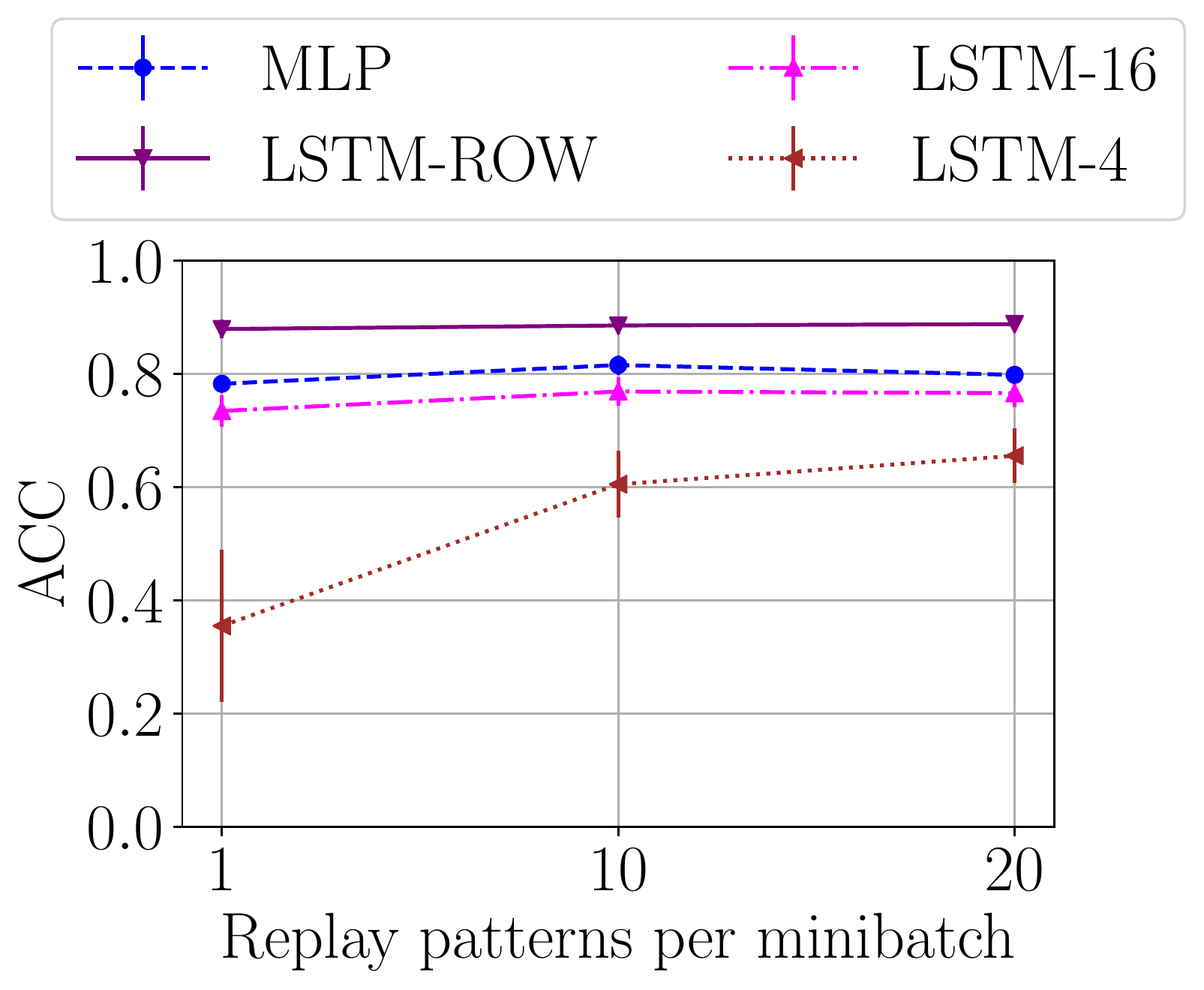}
    \caption{Split MNIST}
    \label{fig:replaysmnist}
    \end{subfigure}
    \begin{subfigure}[t]{0.4\textwidth}
    \centering
    \includegraphics[width=\textwidth]{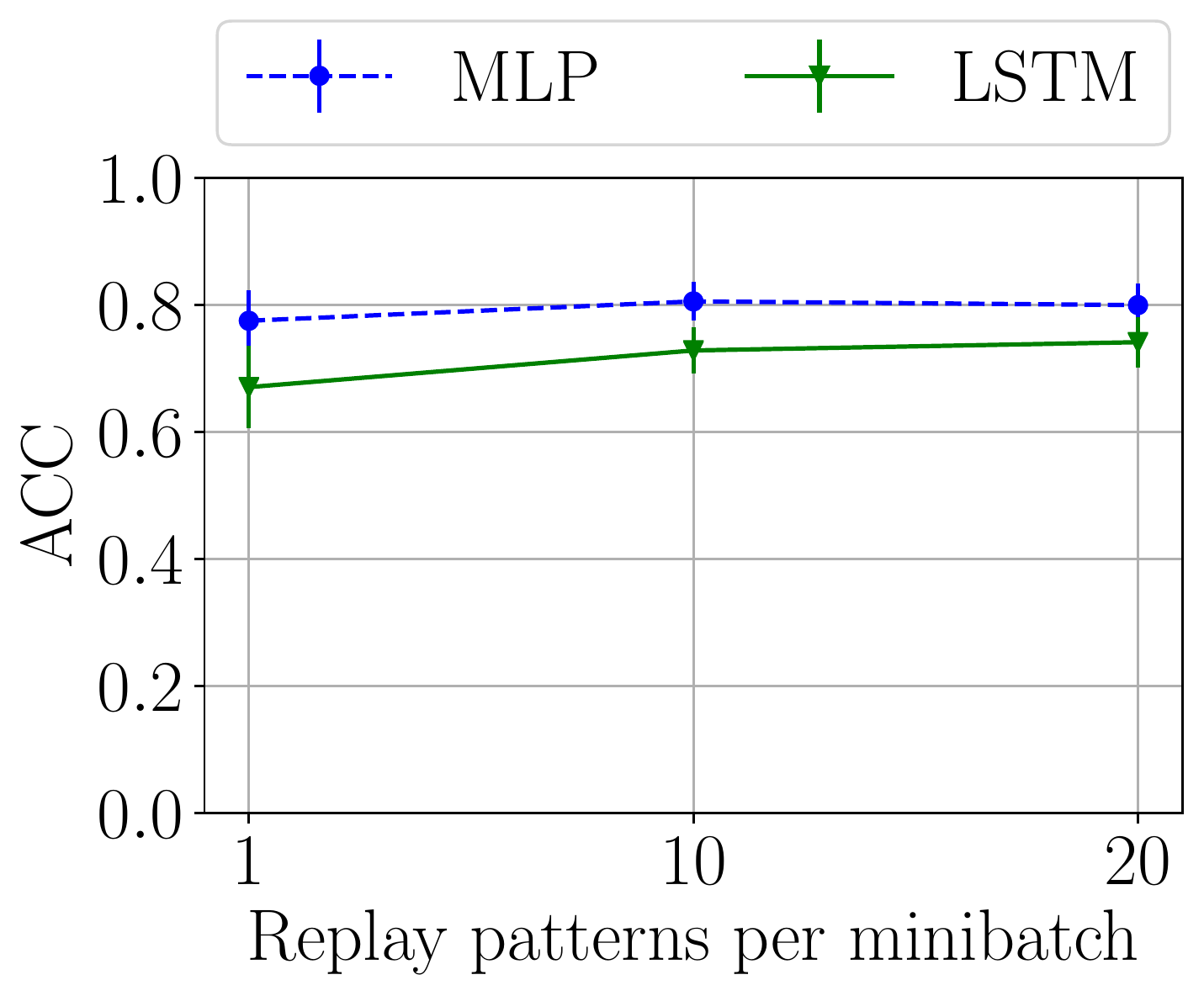}
    \caption{SSC}
    \label{fig:replayssc}
    \end{subfigure}
    \begin{subfigure}[t]{0.4\textwidth}
    \centering
    \includegraphics[width=\textwidth]{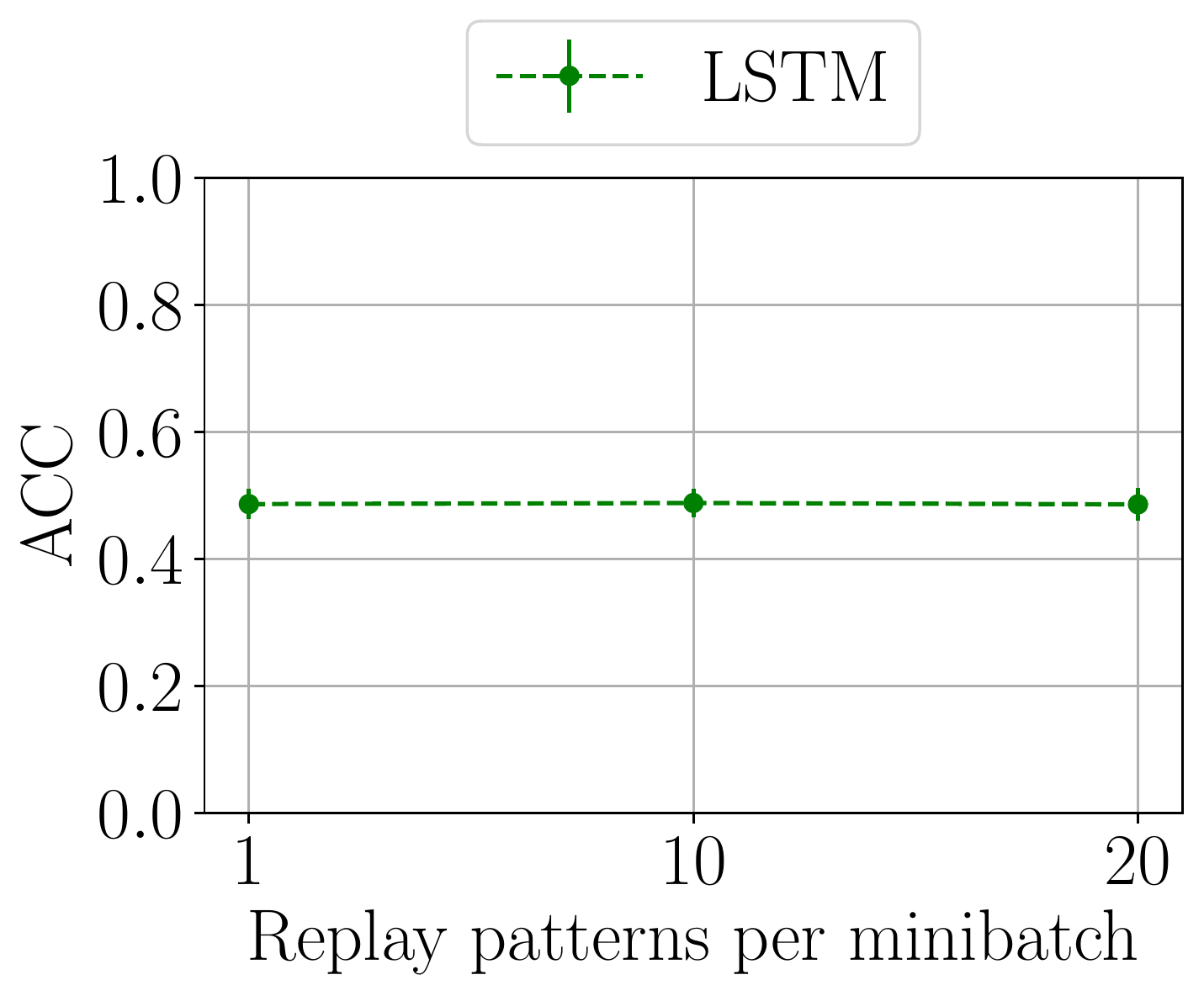}
    \caption{QD}
    \label{fig:replayqd}
    \end{subfigure}
    \caption{Average ACC for different replay patterns per minibatch. The longer the input sequence, the more the patterns needed to recover the performance. Quick, Draw! performance does not change significantly with respect to replay patterns in the minibatch. Best viewed in color. }
    \label{fig:replay}
\end{figure}

We ran a set of experiments aimed at verifying 1) whether general CL strategies are able to mitigate forgetting in recurrent models and 2) the effect of sequence length on the catastrophic forgetting. Table \ref{tab:resultsmnist} reports the mean ACC and its standard deviation (over $5$ runs) computed on PMNIST and SMNIST. Table \ref{tab:resultssdp} reports the same results computed on SSC and QD.

\subsection{Catastrophic Forgetting with CL strategies}
In Class-incremental scenarios, importance-based regularization strategies are subjected to a complete forgetting, as also showed by \cite{lesort2020b} for feedforward models. We confirmed their results also for recurrent models.The effectiveness of replay in \scenario{SIT} scenario is also confirmed by our experiments.

GEM emerges as one of the best performing approaches not only for SMNIST and PMNIST, but also for the more complex SSC and QD. However, its computational cost remains very large (as showed in \ref{app:times}), due to the quadratic optimization requested to find a projecting direction. Unfortunately, its more efficient version A-GEM results in complete forgetting. This is caused by the fact that A-GEM is not constrained by each previous steps, but instead, it computes an approximation of the constraints by sampling patterns randomly from the memory. While this may be sufficient in a \scenario{MT} scenario, it does not work appropriately in a \scenario{SIT} scenario.

LwF has been the most difficult strategy to use, requiring a careful tuning of hyperparameters which questions its applicability to real-world environment. While it works on SMNIST, its performance rapidly decreases on more complex benchmarks like SSC and QD, where it suffers from complete forgetting.

\subsection{Sequence Length affects Catastrophic Forgetting}
Our experiments on PMNIST and SMNIST also allow to draw some conclusions on the effect of sequence length on forgetting. Fig. \ref{fig:seqlength} shows mean ACC results for different sequence lengths. The decreasing trends highlights a clear phenomenon: \textit{long sequences cause more forgetting in recurrent networks}. Regularization strategies suffer the most when increasing sequence length.

Figure \ref{fig:replay} offers some insights also on replay: although more robust than the other strategies in terms of final performance, replay needs more patterns to recover from forgetting as sequences become longer. 

The results on PMNIST with MAS may seem to contradict the sequence-length effect, since ROW-LSTM performs surprisingly better than MLP. However, the performance has to be measured relatively to the Naive performance, in which no CL strategy is applied. When compared to this value, the ROW-LSTM does not perform any better than MLP. Instead it achieves a worse accuracy with respect to the Naive, as expected. The MAS-Naive ratio is $1.87$ for MLP, but only $1.05$ for MAS. 

\subsection{Comparison between Multi-Head and Single-Head Models} \label{sec:multihead}

\begin{figure}[t]
    \centering
    \begin{subfigure}[t]{0.49\textwidth}
    \centering
    \includegraphics[width=\textwidth]{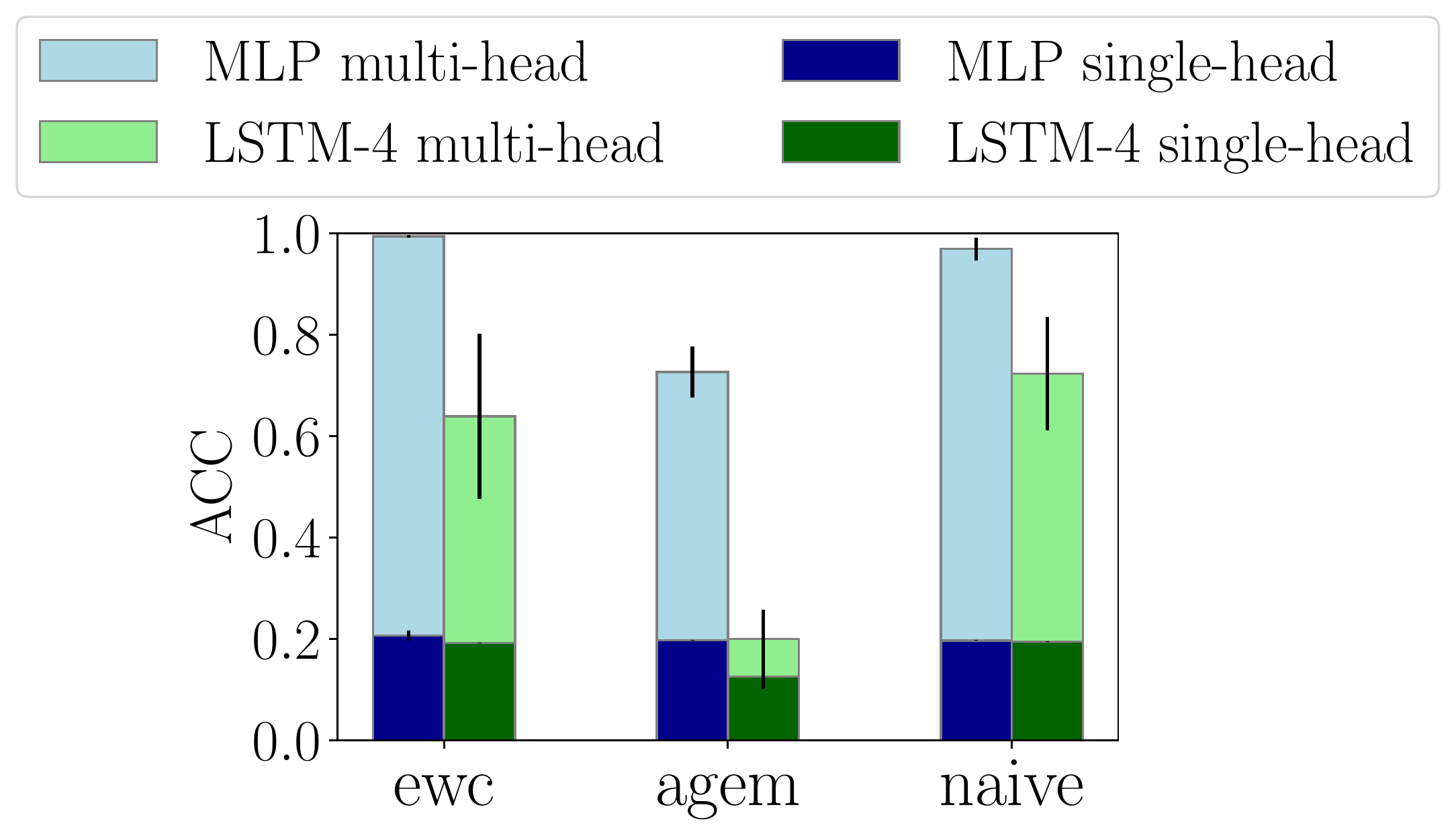}
    \caption{Split MNIST}
    \label{fig:smnistmultihead}
    \end{subfigure}
    \begin{subfigure}[t]{0.46\textwidth}
    \centering
    \includegraphics[width=\textwidth]{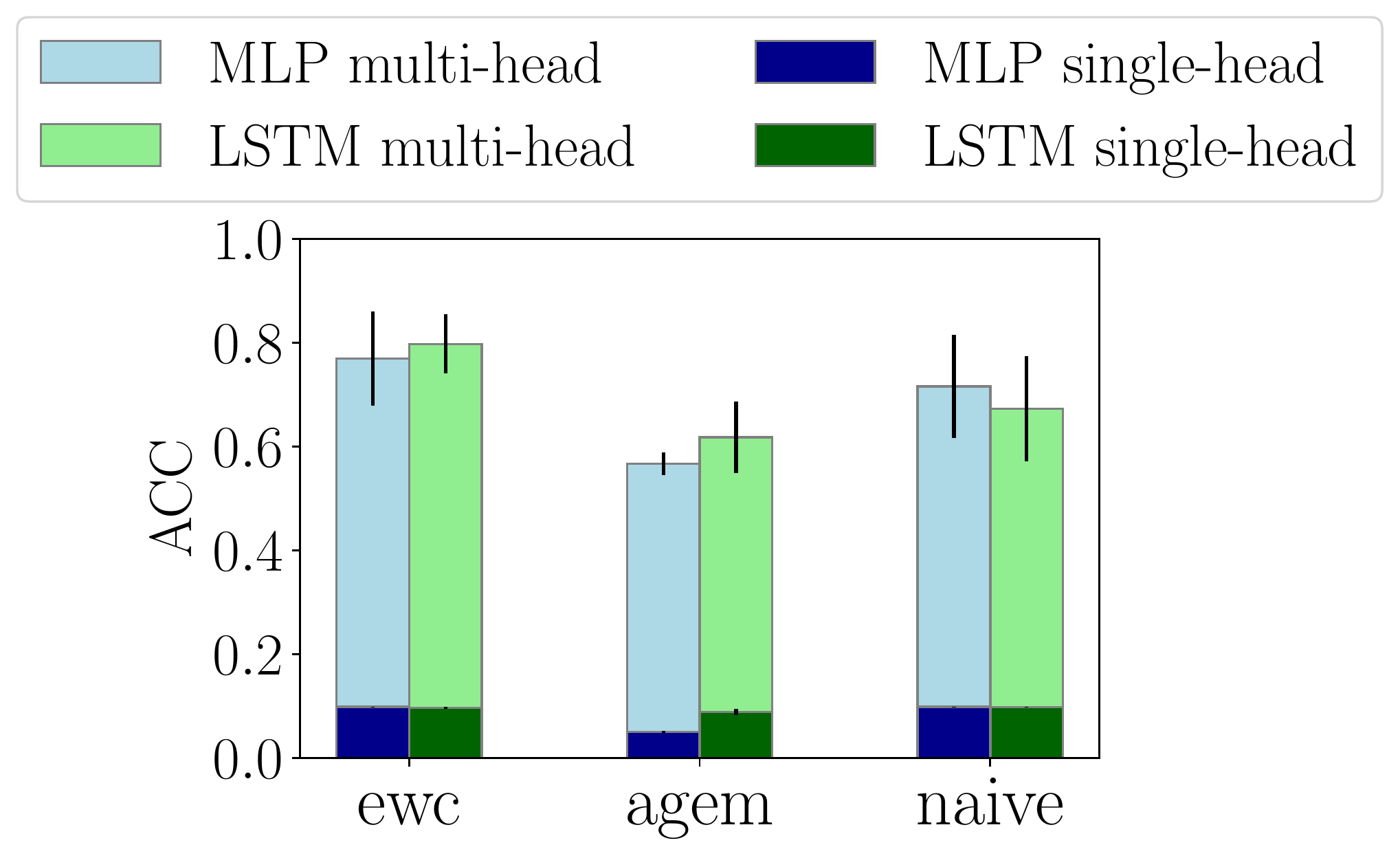}
    \caption{Synthetic Speech Commands}
    \label{fig:sscmultihead}
    \end{subfigure}
    \begin{subfigure}[t]{0.48\textwidth}
    \centering
    \includegraphics[width=\textwidth]{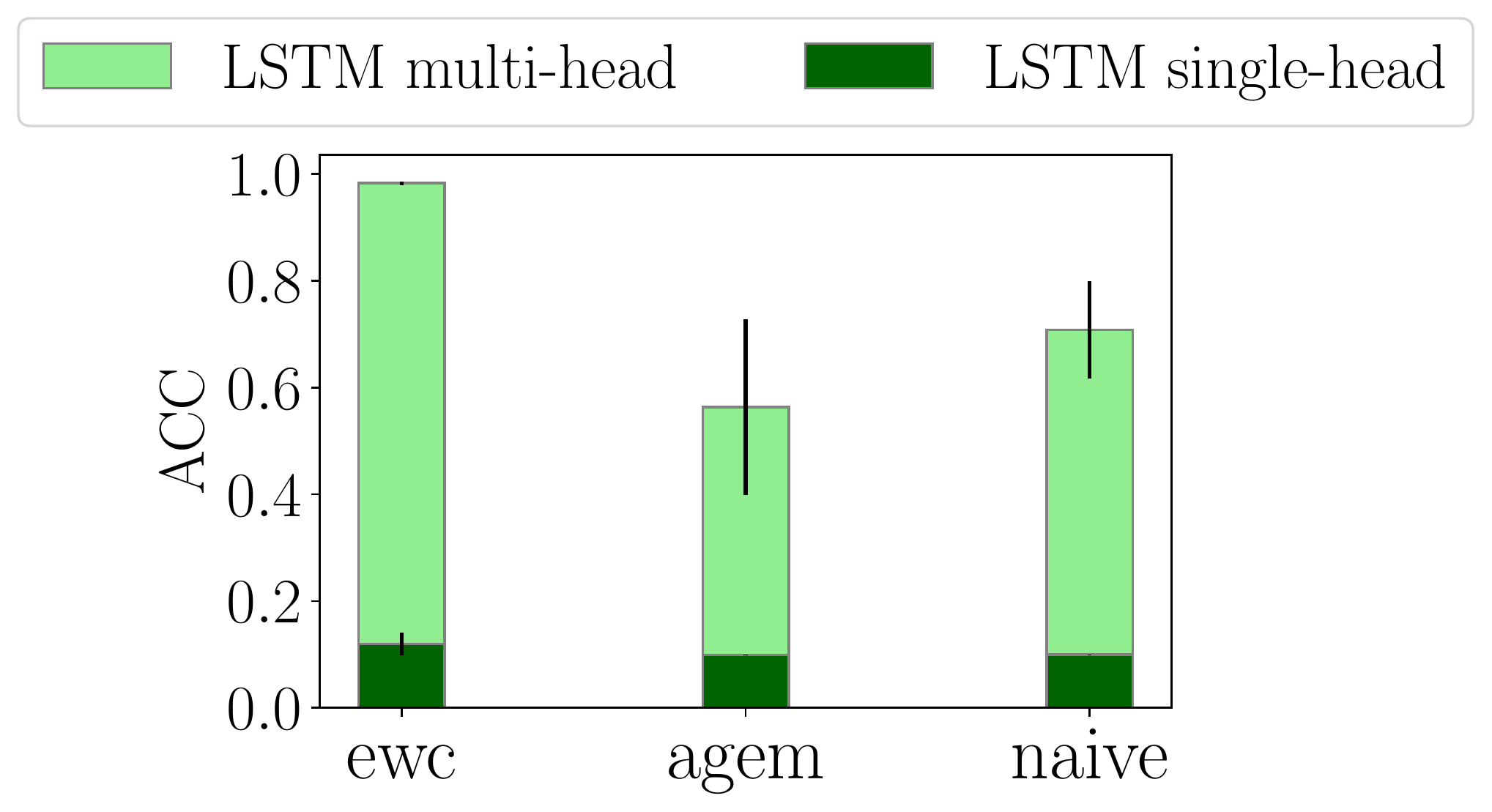}
    \caption{Quick, Draw!}
    \label{fig:qdmultihead}
    \end{subfigure}
    \caption{Bar plot comparing multi-head and single-head performances. Best viewed in color.}
    \label{fig:multiheadresults}
\end{figure}

Although in this work we focused on Class Incremental and Domain Incremental scenarios, we also provide a comparison with Task-Incremental scenarios (MT + NC). Figure \ref{fig:multiheadresults} shows results for the same experimental configuration in Section \ref{sec:results}, but with a multi-head model instead of a single-head. We did not test multi-head models in PMNIST since this benchmark always uses the same output units for all tasks. \ref{app:additional} reports Table \ref{tab:multiheadresults} with detailed results.

There is a clear difference between single and multi-head performances. A multi-head setting easily prevents CF: even a simple Naive procedure is able to retain most of the knowledge in both class-incremental scenarios.
The influence of sequence length in single-headed models is not reflected in the multi-head case. In fact, the LSTM performs better than MLP on SSC: this could be related to the fact that, given the already large mitigation of forgetting with multi-head, the sequential nature of the problem favours a recurrent network.

We concluded that \scenario{MT} scenarios with multi-head models should be used only when there is the need to achieve high-level performances in environments where tasks are easily distinguishable. However, task label information should never be used to derive general conclusions about the behavior and performance of CL algorithms.

\section{Research Directions}
The problem of learning continuously on sequential data processing tasks is far from being solved. Fortunately, the recent interest in the topic is fostering the discovery of new solutions and a better understanding of the phenomena at play. We believe that our work could play an important role in sketching promising research directions. 

\paragraph{Leveraging existing CL strategies} 
From our experiments, it is clear that some families of CL strategies are better suited to the Class-incremental scenario than others (e.g. GEM mitigates forgetting across all benchmarks while EWC performs poorly in class-incremental settings). Instead of designing entirely new approaches, it could be better to start building from the existing ones. Replay offers a simple and effective solution, but storing previous patterns may not be possible in all environments. It would be interesting to extend LwF with knowledge distillation algorithms specifically designed for recurrent networks. Depending on the effectiveness of the distillation, results could drastically change. \\
Even if architectural strategies have not been part of our study, they can be an important part of the research. In particular, they may be a valid option when all other approaches struggle. In fact, task-specific components or freezing of previous parameters can remove forgetting by design. The usually large computational cost can be mitigated by pruning. Combining pruning techniques for RNNs with dynamic architectures may open a new topic of research in CL. 

\paragraph{Improving recurrent models}
The role of sequence length in continuous learning processes can be investigated under different points of view. From the point of view of dynamical systems, the learning trajectory followed by a recurrent model can be studied in terms of different regimes (chaotic, edge-of-chaos, attractors) \cite{ceni2020}. Longer sequences would produce longer trajectories which could make the Continual Learning problem more difficult. Inspecting and interpreting these trajectories could in turn foster the design of new regularization strategies to constrain the learning path. Instead, from an experimental point of view, the optimization problem that has to be solved by recurrent networks is highly dependent on the sequence length. For example, the Backpropagation Through Time has to propagate information from the end of the sequence to the very beginning in order to compute gradients. Some of the obstacles encountered by many strategies in our experimental analysis may be overcome by alternative learning algorithms, for example Truncated Backpropagation.

\paragraph{Design of realistic scenarios}
The complexity of the CL environments is another important topic to address in future researches. Providing target labels to the system each time a new pattern arrives is often too expensive, even unrealistic. Instead, sequences are able to work with sparse targets \cite{graves2006}. Sequential data processing may lead to more autonomous systems in which the required level of supervision is drastically reduced. \\
Our proposed benchmarks could contribute to realize such scenario through a stream of realistic patterns, resembling what an agent would experience in the real world. To build this setting, it could be required to concatenate multiple commands from SSC or multiple drawings from QD, possibly interleaved with noisy patterns. This \scenario{online} scenario would free the model from the necessity to learn over multiple passes over the data. Instead, the environment itself would provide new instances and new classes (\scenario{SIT+NIC}) which can be experienced one (or few) at a time. With this achievement, CL models could seamlessly be integrated in many real world applications.

\section{Conclusion} \label{sec:future}
In this work we focused on Continual Learning in Recurrent Neural Network models. First, we reviewed the literature and proposed a taxonomy of contributions. We also analysed and organized the benchmarks and datasets used in CL for sequential data processing.\\
The number of existing benchmarks compared to the relatively small number of contributions revealed that most papers proposed their own model and tested it on datasets which are almost never reused in subsequent works. Therefore, we discussed the opportunity to build standard benchmarks and, in addition, we proposed two new CL benchmarks based on existing datasets: \textit{Synthetic Speech Commands} (SSC) and \textit{Quick, Draw!} (QD). Both datasets are specifically tailored to Sequential Data Processing and are general enough to be model agnostic. \\
From the literature review, we also found out that all available works focus on new CL strategies with customized architectures, often for a specialized domain. The different experimental setups prevent a comparison of multiple approaches and there is little information about how recurrent networks behave with general CL strategies. \\
To fill this gap, we ran an extensive experimental analysis with single-head models focused on Class-Incremental scenarios. We compared traditional LSTMs against feedforward networks on $6$ popular CL strategies. Our main finding shows increasing forgetting as sequences are made artificially longer without modifying their content. Even replay strategies, which are the most robust among the considered methods, are subjected to this phenomenon. This holds also for Domain-incremental scenarios with Permuted MNIST. \\
The effect of sequence length on the final performance may be questioned by results presented in \cite{ehret2020}: the authors found the \textit{working memory} (the amount of features needed to address a task) to be the main driving factor in the performance of recurrent models. In our work, we showed that, \textit{with fixed working memory}, sequence length plays a role. These two conclusions are not directly in contrast with each other, because the experiments of \cite{ehret2020} have been conducted on either a \scenario{MT+NC} scenarios with multi-head models or on simpler scenarios like Domain-incremental. This difficulty in comparing results confirms the need for our large-scale evaluation.\\
Finally, to justify our choice of Class-incremental scenarios, we ran additional experiments with multi-head models in Task-incremental scenarios. We verified that a multi-head approach greatly simplifies the problem in both feedforward and recurrent architectures. However, this comes at the cost of a strong assumption on task label knowledge.\\
Concluding, we entered into a discussion of interesting research paths inspired by our findings on the topic of Continual Learning in sequential domains. We believe that addressing such challenges will make Continual Learning more effective in realistic environments, where sequential data processing is the most natural way to acquire and process new information across a lifetime. 

\section*{Acknowledgments} We thank the ContinualAI association and all of its members for the fruitful discussions.

\section*{Funding} 
This work has been partially supported by the European Community H2020 programme under project TEACHING (grant n. 871385).

\bibliographystyle{plain}
\bibliography{CL, Others, bib_ac}

\begin{thebibliography}{100}

\bibitem{ahmad2016}
Subutai Ahmad and Jeff Hawkins.
\newblock How do neurons operate on sparse distributed representations? {{A}}
  mathematical theory of sparsity, neurons and active dendrites.
\newblock {\em arXiv}, pages 1--23, 2016.

\bibitem{ahn2019}
Hongjoon Ahn, Sungmin Cha, Donggyu Lee, and Taesup Moon.
\newblock Uncertainty-based {{Continual Learning}} with {{Adaptive
  Regularization}}.
\newblock In {\em {{NeurIPS}}}, pages 4392--4402, 2019.

\bibitem{aljundi2018}
Rahaf Aljundi, Francesca Babiloni, Mohamed Elhoseiny, Marcus Rohrbach, and
  Tinne Tuytelaars.
\newblock Memory {{Aware Synapses}}: {{Learning}} what (not) to forget.
\newblock In {\em The {{European Conference}} on {{Computer Vision}}
  ({{ECCV}})}, 2018.

\bibitem{aljundi2019b}
Rahaf Aljundi, Eugene Belilovsky, Tinne Tuytelaars, Laurent Charlin, Massimo
  Caccia, Min Lin, and Lucas {Page-Caccia}.
\newblock Online {{Continual Learning}} with {{Maximal Interfered Retrieval}}.
\newblock In {\em NeurIPS}, pages 11849--11860. {Curran Associates, Inc.},
  2019.

\bibitem{aljundi2019d}
Rahaf Aljundi, Klaas Kelchtermans, and Tinne Tuytelaars.
\newblock Task-{{Free Continual Learning}}.
\newblock In {\em The {{IEEE Conference}} on {{Computer Vision}} and {{Pattern
  Recognition}} ({{CVPR}})}, 2019.

\bibitem{aljundi2019c}
Rahaf Aljundi, Marcus Rohrbach, and Tinne Tuytelaars.
\newblock Selfless {{Sequential Learning}}.
\newblock In {\em {{ICLR}}}, 2019.

\bibitem{haresn2016}
Giuseppe Amato, Davide Bacciu, Stefano Chessa, Mauro Dragone, Claudio
  Gallicchio, Claudio Gennaro, Hector Lozano, Alessio Micheli, Gregory M.~P.
  O'Hare, Arantxa Renteria, and Claudio Vairo.
\newblock A benchmark dataset for human activity recognition and ambient
  assisted living.
\newblock In Helena Lindgren, Juan~F. De~Paz, Paulo Novais, Antonio
  Fern{\'a}ndez-Caballero, Hyun Yoe, Andres Jim{\'e}nez~Ram{\'i}rez, and
  Gabriel Villarrubia, editors, {\em Ambient Intelligence- Software and
  Applications -- 7th International Symposium on Ambient Intelligence (ISAmI
  2016)}, pages 1--9. Springer International Publishing, 2016.

\bibitem{ans2004}
Bernard Ans, St{\'e}phane Rousset, Robert~M. French, and Serban Musca.
\newblock Self-refreshing memory in artificial neural networks: Learning
  temporal sequences without catastrophic forgetting.
\newblock {\em Connection Science}, 16(2):71--99, 2004.

\bibitem{ans2002}
Bernard Ans, Stephane Rousset, Robert~M. French, and Serban~C. Musca.
\newblock Preventing {{Catastrophic Interference}} in {{MultipleSequence
  Learning Using Coupled Reverberating Elman Networks}}.
\newblock In {\em Proceedings of the 24th {{Annual Conference}} of the
  {{Cognitive Science Society}}}, 2002.

\bibitem{asghar2019}
Nabiha Asghar, Lili Mou, Kira~A Selby, Kevin~D Pantasdo, Pascal Poupart, and
  Xin Jiang.
\newblock Progressive {{Memory Banks}} for {{Incremental Domain Adaptation}}.
\newblock In {\em ICLR}, 2019.

\bibitem{bahdanauNeuralMachineTranslation2014}
Dzmitry Bahdanau, Kyunghyun Cho, and Yoshua Bengio.
\newblock Neural {{Machine Translation}} by {{Jointly Learning}} to {{Align}}
  and {{Translate}}.
\newblock {\em ICLR}, 2015.

\bibitem{beaulieu2020}
Shawn Beaulieu, Lapo Frati, Thomas Miconi, Joel Lehman, Kenneth~O. Stanley,
  Jeff Clune, and Nick Cheney.
\newblock Learning to {{Continually Learn}}.
\newblock In {\em {{ECAI}}}, 2020.

\bibitem{biesialska2020a}
Magdalena Biesialska, Katarzyna Biesialska, and Marta~R. {Costa-juss{\`a}}.
\newblock Continual {{Lifelong Learning}} in {{Natural Language Processing}}:
  {{A Survey}}.
\newblock In {\em Proceedings of the 28th {{International Conference}} on
  {{Computational Linguistics}}}, pages 6523--6541, {Barcelona, Spain
  (Online)}, 2020. {International Committee on Computational Linguistics}.

\bibitem{bojar2017}
Ond{\v r}ej Bojar, Rajen Chatterjee, Christian Federmann, Yvette Graham, Barry
  Haddow, Shujian Huang, Matthias Huck, Philipp Koehn, Qun Liu, Varvara
  Logacheva, Christof Monz, Matteo Negri, Matt Post, Raphael Rubino, Lucia
  Specia, and Marco Turchi.
\newblock Findings of the 2017 {{Conference}} on {{Machine Translation}}
  ({{WMT17}}).
\newblock In {\em Proceedings of the {{Second Conference}} on {{Machine
  Translation}}}, pages 169--214, {Copenhagen, Denmark}, 2017. {Association for
  Computational Linguistics}.

\bibitem{speechcommands}
Johannes Buchner.
\newblock Synthetic {{Speech Commands}}: {{A}} public dataset for single-word
  speech recognition.
\newblock {\em Kaggle Dataset}, 2017.

\bibitem{caccia2020}
Massimo Caccia, Pau Rodriguez, Oleksiy Ostapenko, Fabrice Normandin, Min Lin,
  Lucas Caccia, Issam Laradji, Irina Rish, Alexande Lacoste, David Vazquez, and
  Laurent Charlin.
\newblock Online {{Fast Adaptation}} and {{Knowledge Accumulation}}: A {{New
  Approach}} to {{Continual Learning}}.
\newblock {\em arXiv}, 2020.

\bibitem{carta2021}
Antonio Carta, Andrea Cossu, Federico Errica, and Davide Bacciu.
\newblock Catastrophic {{Forgetting}} in {{Deep Graph Networks}}: An
  {{Introductory Benchmark}} for {{Graph Classification}}.
\newblock {\em The 2021 Web Conference (WWW) Workshop on Graph Benchmarks
  Learning (GLB)}, 2021.

\bibitem{ceni2020}
Andrea Ceni, Peter Ashwin, and Lorenzo Livi.
\newblock Interpreting {{Recurrent Neural Networks Behaviour}} via {{Excitable
  Network Attractors}}.
\newblock {\em Cognitive Computation}, 12(2):330--356, 2020.

\bibitem{chaudhry2018}
Arslan Chaudhry, Puneet~K. Dokania, Thalaiyasingam Ajanthan, and Philip H.~S.
  Torr.
\newblock Riemannian {{Walk}} for {{Incremental Learning}}: {{Understanding
  Forgetting}} and {{Intransigence}}.
\newblock In {\em Proceedings of the {{European Conference}} on {{Computer
  Vision}} ({{ECCV}})}, pages 532--547, 2018.

\bibitem{chaudhry2019}
Arslan Chaudhry, Marc'Aurelio Ranzato, Marcus Rohrbach, and Mohamed Elhoseiny.
\newblock Efficient {{Lifelong Learning}} with {{A}}-{{GEM}}.
\newblock In {\em {{ICLR}}}, 2019.

\bibitem{chenRecurrentNeuralNetwork2019}
Dechao Chen, Shuai Li, and Liefa Liao.
\newblock A recurrent neural network applied to optimal motion control of
  mobile robots with physical constraints.
\newblock {\em Applied Soft Computing}, 85:105880, December 2019.

\bibitem{chen2016}
Tianqi Chen, Ian Goodfellow, and Jonathon Shlens.
\newblock {{Net2Net}}: {{Accelerating Learning}} via {{Knowledge Transfer}}.
\newblock In {\em {{ICLR}}}, 2016.

\bibitem{choPropertiesNeuralMachine2014}
Kyunghyun Cho, Bart {van Merri{\"e}nboer}, Dzmitry Bahdanau, and Yoshua Bengio.
\newblock On the {{Properties}} of {{Neural Machine Translation}}:
  {{Encoder}}\textendash{{Decoder Approaches}}.
\newblock In {\em Proceedings of {{SSST}}-8, {{Eighth Workshop}} on {{Syntax}},
  {{Semantics}} and {{Structure}} in {{Statistical Translation}}}, pages
  103--111, {Doha, Qatar}, 2014. {Association for Computational Linguistics}.

\bibitem{chungEmpiricalEvaluationGated2014}
Junyoung Chung, Caglar Gulcehre, Kyunghyun Cho, and Yoshua Bengio.
\newblock {Empirical evaluation of gated recurrent neural networks on sequence
  modeling}.
\newblock In {\em {NIPS Workshop on Deep Learning}}, 2014.

\bibitem{coop2012}
Robert Coop and Itamar Arel.
\newblock Mitigation of catastrophic interference in neural networks using a
  fixed expansion layer.
\newblock In {\em 2012 {{IEEE}} 55th {{International Midwest Symposium}} on
  {{Circuits}} and {{Systems}} ({{MWSCAS}})}, pages 726--729. {IEEE}, 2012.

\bibitem{coop2013}
Robert Coop and Itamar Arel.
\newblock Mitigation of catastrophic forgetting in recurrent neural networks
  using a {{Fixed Expansion Layer}}.
\newblock In {\em The 2013 {{International Joint Conference}} on {{Neural
  Networks}} ({{IJCNN}})}, pages 1--7, {Dallas, TX, USA}, 2013. {IEEE}.

\bibitem{cossu2020}
Andrea Cossu, Antonio Carta, and Davide Bacciu.
\newblock Continual {{Learning}} with {{Gated Incremental Memories}} for
  sequential data processing.
\newblock In {\em Proceedings of the 2020 {{International Joint Conference}} on
  {{Neural Networks}} ({{IJCNN}} 2020)}, 2020.

\bibitem{cui2016}
Yuwei Cui, Subutai Ahmad, and Jeff Hawkins.
\newblock Continuous {{Online Sequence Learning}} with an {{Unsupervised Neural
  Network Model}}.
\newblock {\em Neural Computation}, 28(11):2474--2504, 2016.

\bibitem{dejongIncrementalSequenceLearning2016}
Edwin~D. {de Jong}.
\newblock Incremental {{Sequence Learning}}.
\newblock {\em arXiv: 1611.03068 [cs]}, 2016.

\bibitem{delange2019}
Matthias De~Lange, Rahaf Aljundi, Marc Masana, Sarah Parisot, Xu~Jia, Ales
  Leonardis, Gregory Slabaugh, and Tinne Tuytelaars.
\newblock A continual learning survey: {{Defying}} forgetting in classification
  tasks.
\newblock {\em arXiv}, 2019.

\bibitem{ditzler2015}
Gregory Ditzler, Manuel Roveri, Cesare Alippi, and Robi Polikar.
\newblock Learning in {{Nonstationary Environments}}: {{A Survey}}.
\newblock {\em IEEE Computational Intelligence Magazine}, 10(4):12--25, 2015.

\bibitem{duncker2020}
Lea Duncker, Laura~N Driscoll, Krishna~V Shenoy, Maneesh Sahani, and David
  Sussillo.
\newblock Organizing recurrent network dynamics by task-computation to enable
  continual learning.
\newblock In {\em NeurIPS}, volume~33, 2020.

\bibitem{ebrahimi2020}
Sayna Ebrahimi, Mohamed Elhoseiny, Trevor Darrell, and Marcus Rohrbach.
\newblock Uncertainty-guided {{Continual Learning}} with {{Bayesian Neural
  Networks}}.
\newblock In {\em {{ICLR}}}, 2020.

\bibitem{ehret2020}
Benjamin Ehret, Christian Henning, Maria~R Cervera, Alexander Meulemans,
  Johannes {von Oswald}, and Benjamin~F Grewe.
\newblock Continual {{Learning}} in {{Recurrent Neural Networks}}.
\newblock {\em arXiv}, 2020.

\bibitem{ehretContinualLearningRecurrent2020}
Benjamin Ehret, Christian Henning, Maria~R. Cervera, Alexander Meulemans,
  Johannes {von Oswald}, and Benjamin~F. Grewe.
\newblock Continual {{Learning}} in {{Recurrent Neural Networks}} with
  {{Hypernetworks}}.
\newblock {\em arXiv:2006.12109 [cs, stat]}, 2020.

\bibitem{elmanFindingStructureTime1990}
Jeffrey~L. Elman.
\newblock Finding {{Structure}} in {{Time}}.
\newblock {\em Cognitive Science}, 14(2):179--211, 1990.

\bibitem{farquhar2018}
Sebastian Farquhar and Yarin Gal.
\newblock A {{Unifying Bayesian View}} of {{Continual Learning}}.
\newblock In {\em {{NeurIPS Bayesian Deep Learning Workshop}}}, 2018.

\bibitem{farquhar2019}
Sebastian Farquhar and Yarin Gal.
\newblock Towards {{Robust Evaluations}} of {{Continual Learning}}.
\newblock In {\em Privacy in {{Machine Learning}} and {{Artificial
  Intelligence}} Workshop, {{ICML}}}, 2019.

\bibitem{finn2019}
Chelsea Finn, Aravind Rajeswaran, Sham Kakade, and Sergey Levine.
\newblock Online {{Meta}}-{{Learning}}.
\newblock In {\em {{ICML}}}, 2019.

\bibitem{french1991}
Robert French.
\newblock Using {{Semi}}-{{Distributed Representations}} to {{Overcome
  Catastrophic Forgetting}} in {{Connectionist Networks}}.
\newblock In {\em In {{Proceedings}} of the 13th {{Annual Cognitive Science
  Society Conference}}}, pages 173--178. {Erlbaum}, 1991.

\bibitem{french1997}
Robert French.
\newblock Pseudo-recurrent {{Connectionist Networks}}: {{An Approach}} to the
  '{{Sensitivity}}-{{Stability}}' {{Dilemma}}.
\newblock {\em Connection Science}, 9(4):353--380, 1997.

\bibitem{french1997a}
Robert French.
\newblock Using {{Pseudo}}-{{Recurrent Connectionist Networks}} to {{Solve}}
  the {{Problem}} of {{Sequential Learning}}.
\newblock In {\em Proceedings of the 19th {{Annual Cognitive Science Society
  Conference}}}, 1997.

\bibitem{frenchCatastrophicForgettingConnectionist1999}
Robert French.
\newblock Catastrophic forgetting in connectionist networks.
\newblock {\em Trends in Cognitive Sciences}, 3(4):128--135, 1999.

\bibitem{frenchUsingSemiDistributedRepresentations1991}
Robert~M. French.
\newblock Using {{Semi}}-{{Distributed Representations}} to {{Overcome
  Catastrophic Forgetting}} in {{Connectionist Networks}}.
\newblock In {\em In {{Proceedings}} of the 13th {{Annual Cognitive Science
  Society Conference}}}, pages 173--178. {Erlbaum}, 1991.

\bibitem{gamaSurveyConceptDrift2014}
Jo{\~a}o Gama, Indr{\.e} {\v Z}liobait{\.e}, Albert Bifet, Mykola Pechenizkiy,
  and Abdelhamid Bouchachia.
\newblock A survey on concept drift adaptation.
\newblock {\em ACM Computing Surveys (CSUR)}, 46(4):44:1--44:37, 2014.

\bibitem{gemmekeAudioSetOntology2017}
J.~F. Gemmeke, D.~P.~W. Ellis, D.~Freedman, A.~Jansen, W.~Lawrence, R.~C.
  Moore, M.~Plakal, and M.~Ritter.
\newblock Audio {{Set}}: {{An}} ontology and human-labeled dataset for audio
  events.
\newblock In {\em 2017 {{IEEE International Conference}} on {{Acoustics}},
  {{Speech}} and {{Signal Processing}} ({{ICASSP}})}, pages 776--780, 2017.

\bibitem{golkar2019}
Siavash Golkar, Michael Kagan, and Kyunghyun Cho.
\newblock Continual {{Learning}} via {{Neural Pruning}}.
\newblock {\em arXiv}, 2019.

\bibitem{gravesSequenceTransductionRecurrent2012}
Alex Graves.
\newblock Sequence {Transduction} with {Recurrent} {Neural} {Networks}.
\newblock {\em arXiv}, 2012.

\bibitem{graves2006}
Alex Graves, Santiago Fern{\'a}ndez, Faustino Gomez, and J{\"u}rgen
  Schmidhuber.
\newblock Connectionist temporal classification: Labelling unsegmented sequence
  data with recurrent neural networks.
\newblock In {\em ICML}, {{ICML}} '06, pages 369--376, {New York, NY, USA},
  2006. {Association for Computing Machinery}.

\bibitem{gravesNeuralTuringMachines2014}
Alex Graves, Greg Wayne, and Ivo Danihelka.
\newblock Neural {{Turing Machines}}.
\newblock {\em arXiv:1410.5401 [cs]}, 2014.

\bibitem{grossberg1980}
Stephen Grossberg.
\newblock How does a brain build a cognitive code?
\newblock {\em Psychological Review}, 87(1):1--51, 1980.

\bibitem{ha2018}
David Ha and Douglas Eck.
\newblock A {{Neural Representation}} of {{Sketch Drawings}}.
\newblock In {\em {{ICLR}}}, 2018.

\bibitem{harriesExtractingHiddenContext1998}
Michael~Bonnell Harries, Claude Sammut, and Kim Horn.
\newblock Extracting {{Hidden Context}}.
\newblock {\em Machine Learning}, 32(2):101--126, 1998.

\bibitem{harrison2019}
James Harrison, Apoorva Sharma, Chelsea Finn, and Marco Pavone.
\newblock Continuous meta-learning without tasks.
\newblock {\em arXiv}, 2019.

\bibitem{hasanContinuousLearningFramework2015}
Mahmudul Hasan and Amit~K. {Roy-Chowdhury}.
\newblock A {{Continuous Learning Framework}} for {{Activity Recognition Using
  Deep Hybrid Feature Models}}.
\newblock {\em IEEE Transactions on Multimedia}, 17(11):1909--1922, 2015.

\bibitem{hayes2018}
Tyler~L Hayes, Nathan~D Cahill, and Christopher Kanan.
\newblock Memory {{Efficient Experience Replay}} for {{Streaming Learning}}.
\newblock {\em IEEE International Conference on Robotics and Automation
  (ICRA)}, 2018.

\bibitem{he2019}
Xu~He, Jakub Sygnowski, Alexandre Galashov, Andrei~A Rusu, Yee~Whye Teh, and
  Razvan Pascanu.
\newblock {\em Task {{Agnostic Continual Learning}} via {{Meta Learning}}}.
\newblock arxiv, 2019.

\bibitem{hintonDistillingKnowledgeNeural2015}
Geoffrey Hinton, Oriol Vinyals, and Jeffrey Dean.
\newblock Distilling the {{Knowledge}} in a {{Neural Network}}.
\newblock In {\em {{NIPS Deep Learning}} and {{Representation Learning
  Workshop}}}, 2015.

\bibitem{hochreiterLongShortTermMemory1997}
Sepp Hochreiter and Jurgen Schmidhuber.
\newblock Long {{Short}}-{{Term Memory}}.
\newblock {\em Neural Computation}, 9:1735--1780, 1997.

\bibitem{hospedales2020}
Timothy Hospedales, Antreas Antoniou, Paul Micaelli, and Amos Storkey.
\newblock Meta-{{Learning}} in {{Neural Networks}}: {{A Survey}}.
\newblock {\em arXiv:2004.05439 [cs, stat]}, 2020.

\bibitem{hung2019}
Steven C~Y Hung, Cheng-Hao Tu, Cheng-En Wu, Chien-Hung Chen, Yi-Ming Chan, and
  Chu-Song Chen.
\newblock Compacting, {{Picking}} and {{Growing}} for {{Unforgetting Continual
  Learning}}.
\newblock In {\em {{NeurIPS}}}, pages 13669--13679, 2019.

\bibitem{javed2019}
Khurram Javed and Martha White.
\newblock Meta-{{Learning Representations}} for {{Continual Learning}}.
\newblock In {\em {{NeurIPS}}}, 2019.

\bibitem{mjd2016coppa}
Marcin {Junczys-Dowmunt}, Bruno Pouliquen, and Christophe Mazenc.
\newblock {{COPPA V2}}.0: {{Corpus}} of parallel patent applications.
  {{Building}} large parallel corpora with {{GNU}} make.
\newblock In {\em Proceedings of the 4th Workshop on Challenges in the
  Management of Large Corpora, Portoro\v{z}, Slovenia, May 23-28, 2016}, 2016.

\bibitem{kirkpatrick2017}
James Kirkpatrick, Razvan Pascanu, Neil Rabinowitz, Joel Veness, Guillaume
  Desjardins, Andrei~A Rusu, Kieran Milan, John Quan, Tiago Ramalho, Agnieszka
  {Grabska-Barwinska}, Demis Hassabis, Claudia Clopath, Dharshan Kumaran, and
  Raia Hadsell.
\newblock Overcoming catastrophic forgetting in neural networks.
\newblock {\em PNAS}, 114(13):3521--3526, 2017.

\bibitem{kobayashi2019}
Taisuke Kobayashi and Toshiki Sugino.
\newblock Continual {{Learning Exploiting Structure}} of {{Fractal Reservoir
  Computing}}.
\newblock In Igor~V Tetko, V{\v e}ra K{\r{u}}rkov{\'a}, Pavel Karpov, and
  Fabian Theis, editors, {\em Artificial {{Neural Networks}} and {{Machine
  Learning}} \textendash{} {{ICANN}} 2019: {{Workshop}} and {{Special
  Sessions}}}, volume 11731, pages 35--47, {Cham}, 2019. {Springer
  International Publishing}.

\bibitem{kruszewski2020a}
Germ{\'a}n Kruszewski, Ionut-Teodor Sorodoc, and Tomas Mikolov.
\newblock Evaluating {{Online Continual Learning}} with {{CALM}}.
\newblock {\em arXiv}, 2020.

\bibitem{kurle2020}
Richard Kurle, Botond Cseke, Alexej Klushyn, Patrick van~der Smagt, and Stephan
  G{\"u}nnemann.
\newblock Continual {{Learning}} with {{Bayesian Neural Networks}} for
  {{Non}}-{{Stationary Data}}.
\newblock In {\em ICLR}, 2020.

\bibitem{kusupatiFastGRNNFastAccurate2019}
Aditya Kusupati, Manish Singh, Kush Bhatia, Ashish Kumar, Prateek Jain, and
  Manik Varma.
\newblock {FastGRNN}: {A} {Fast}, {Accurate}, {Stable} and {Tiny} {Kilobyte}
  {Sized} {Gated} {Recurrent} {Neural} {Network}.
\newblock {\em arXiv:1901.02358 [cs, stat]}, January 2019.

\bibitem{leSimpleWayInitialize2015}
Quoc~V Le, Navdeep Jaitly, and Geoffrey~E Hinton.
\newblock A simple way to initialize recurrent networks of rectified linear
  units.
\newblock {\em arXiv preprint arXiv:1504.00941}, 2015.

\bibitem{lecunGradientBasedLearningApplied1998}
Yann LeCun, L{\'e}on Bottou, Yoshua Bengio, and Patrick Haffner.
\newblock Gradient-{{Based Learning Applied}} to {{Document Recognition}}.
\newblock {\em Proceedings of the IEEE}, 86(11):2278--2324, 1998.

\bibitem{lesort2020}
Timoth{\'e}e Lesort, Vincenzo Lomonaco, Andrei Stoian, Davide Maltoni, David
  Filliat, and Natalia {D{\'i}az-Rodr{\'i}guez}.
\newblock Continual learning for robotics: {{Definition}}, framework, learning
  strategies, opportunities and challenges.
\newblock {\em Information Fusion}, 58:52--68, 2020.

\bibitem{lesort2020b}
Timoth{\'e}e Lesort, Andrei Stoian, and David Filliat.
\newblock Regularization {{Shortcomings}} for {{Continual Learning}}.
\newblock {\em arXiv}, 2020.

\bibitem{li2019}
HongLin Li, Payam Barnaghi, Shirin Enshaeifar, and Frieder Ganz.
\newblock Continual {{Learning Using Bayesian Neural Networks}}.
\newblock {\em arXiv}, 2019.

\bibitem{li2020b}
Yuanpeng Li, Liang Zhao, Kenneth Church, and Mohamed Elhoseiny.
\newblock Compositional {{Language Continual Learning}}.
\newblock In {\em ICLR}, 2020.

\bibitem{li2016}
Zhizhong Li and Derek Hoiem.
\newblock Learning without {{Forgetting}}.
\newblock In {\em European {{Conference}} on {{Computer Vision}}}, Springer,
  pages 614--629, 2016.

\bibitem{lison2018}
Pierre Lison, J{\"o}rg Tiedemann, and Milen Kouylekov.
\newblock {{OpenSubtitles2018}}: {{Statistical Rescoring}} of {{Sentence
  Alignments}} in {{Large}}, {{Noisy Parallel Corpora}}.
\newblock In {\em Proceedings of the 11th {{International Conference}} on
  {{Language Resources}} and {{Evaluation}} ({{LREC}} 2018)}, pages 1742--1748.
  {European Language Resources Association (ELRA)}, 2018.

\bibitem{lomonaco2017}
Vincenzo Lomonaco and Davide Maltoni.
\newblock {{CORe50}}: A {{New Dataset}} and {{Benchmark}} for {{Continuous
  Object Recognition}}.
\newblock In Sergey Levine, Vincent Vanhoucke, and Ken Goldberg, editors, {\em
  Proceedings of the 1st {{Annual Conference}} on {{Robot Learning}}},
  volume~78 of {\em Proceedings of {{Machine Learning Research}}}, pages
  17--26. {PMLR}, 2017.

\bibitem{lomonaco2021}
Vincenzo Lomonaco, Lorenzo Pellegrini, Andrea Cossu, Antonio Carta, Gabriele
  Graffieti, Tyler~L. Hayes, Matthias De~Lange, Marc Masana, Jary Pomponi, Gido
  {van de Ven}, Martin Mundt, Qi~She, Keiland Cooper, Jeremy Forest, Eden
  Belouadah, Simone Calderara, German~I. Parisi, Fabio Cuzzolin, Andreas
  Tolias, Simone Scardapane, Luca Antiga, Subutai Amhad, Adrian Popescu,
  Christopher Kanan, Joost {van de Weijer}, Tinne Tuytelaars, Davide Bacciu,
  and Davide Maltoni.
\newblock Avalanche: An {{End}}-to-{{End Library}} for {{Continual Learning}}.
\newblock In {\em {{CLVision Workshop}} at {{CVPR}}}, 2021.

\bibitem{lopez-paz2017}
David {Lopez-Paz} and Marc'Aurelio Ranzato.
\newblock Gradient {{Episodic Memory}} for {{Continual Learning}}.
\newblock In {\em {{NIPS}}}, 2017.

\bibitem{lukoseviciusReservoirComputingApproaches2009}
Mantas Luko{\v s}evi{\v c}ius and Herbert Jaeger.
\newblock Reservoir computing approaches to recurrent neural network training.
\newblock {\em Computer Science Review}, 3(3):127--149, 2009.

\bibitem{madasu2020}
Avinash Madasu and Vijjini~Anvesh Rao.
\newblock Sequential {{Domain Adaptation}} through {{Elastic Weight
  Consolidation}} for {{Sentiment Analysis}}.
\newblock {\em arXiv}, 2020.

\bibitem{maltoni2019}
Davide Maltoni and Vincenzo Lomonaco.
\newblock Continuous {{Learning}} in {{Single}}-{{Incremental}}-{{Task
  Scenarios}}.
\newblock {\em Neural Networks}, 116:56--73, 2019.

\bibitem{mcclellandIntegrationNewInformation2020}
James~L. McClelland, Bruce~L. McNaughton, and Andrew~K. Lampinen.
\newblock Integration of new information in memory: New insights from a
  complementary learning systems perspective.
\newblock {\em Philosophical Transactions of the Royal Society B: Biological
  Sciences}, 375(1799):20190637, 2020.

\bibitem{mccloskeyCatastrophicInterferenceConnectionist1989}
Michael McCloskey and Neal~J. Cohen.
\newblock Catastrophic {{Interference}} in {{Connectionist Networks}}: {{The
  Sequential Learning Problem}}.
\newblock In Gordon~H. Bower, editor, {\em Psychology of {{Learning}} and
  {{Motivation}}}, volume~24, pages 109--165. {Academic Press}, 1989.

\bibitem{mehta2020}
Nikhil Mehta, Kevin~J Liang, and Lawrence Carin.
\newblock Bayesian {{Nonparametric Weight Factorization}} for {{Continual
  Learning}}.
\newblock {\em arXiv}, pages 1--17, 2020.

\bibitem{nguyen2018}
Cuong~V Nguyen, Yingzhen Li, Thang~D Bui, and Richard~E Turner.
\newblock Variational {{Continual Learning}}.
\newblock In {\em {{ICLR}}}, 2018.

\bibitem{nguyen2019}
Hung Nguyen, Xuejian Wang, and Leman Akoglu.
\newblock Continual {{Rare}}-{{Class Recognition}} with {{Emerging Novel
  Subclasses}}.
\newblock In {\em {{ECML}}}, 2019.

\bibitem{ororbia2020}
Alexander Ororbia.
\newblock Spiking {{Neural Predictive Coding}} for {{Continual Learning}} from
  {{Data Streams}}.
\newblock {\em arXiv}, 2020.

\bibitem{ororbia2019}
Alexander Ororbia, Ankur Mali, C~Lee Giles, and Daniel Kifer.
\newblock Continual {{Learning}} of {{Recurrent Neural Networks}} by {{Locally
  Aligning Distributed Representations}}.
\newblock {\em arXiv}, 2019.

\bibitem{ororbia2019a}
Alexander Ororbia, Ankur Mali, Daniel Kifer, and C~Lee Giles.
\newblock Lifelong {{Neural Predictive Coding}}: {{Sparsity Yields Less
  Forgetting}} when {{Learning Cumulatively}}.
\newblock {\em arXiv}, pages 1--11, 2019.

\bibitem{parisi2019}
German~I Parisi, Ronald Kemker, Jose~L Part, Christopher Kanan, and Stefan
  Wermter.
\newblock Continual lifelong learning with neural networks: {{A}} review.
\newblock {\em Neural Networks}, 113:54--71, 2019.

\bibitem{parisi2018}
German~I Parisi, Jun Tani, Cornelius Weber, and Stefan Wermter.
\newblock Lifelong {{Learning}} of {{Spatiotemporal Representations With
  Dual}}-{{Memory Recurrent Self}}-{{Organization}}.
\newblock {\em Frontiers in Neurorobotics}, 12, 2018.

\bibitem{paszkePyTorchImperativeStyle2019}
Adam Paszke, Sam Gross, Francisco Massa, Adam Lerer, James Bradbury, Gregory
  Chanan, Trevor Killeen, Zeming Lin, Natalia Gimelshein, Luca Antiga, Alban
  Desmaison, Andreas K{\"o}pf, Edward Yang, Zach DeVito, Martin Raison, Alykhan
  Tejani, Sasank Chilamkurthy, Benoit Steiner, Lu~Fang, Junjie Bai, and Soumith
  Chintala.
\newblock {{PyTorch}}: {{An Imperative Style}}, {{High}}-{{Performance Deep
  Learning Library}}.
\newblock {\em NeurIPS}, 2019.

\bibitem{philps2019}
Daniel Philps, Artur d'Avila Garcez, and Tillman Weyde.
\newblock Making {{Good}} on {{LSTMs}}' {{Unfulfilled Promise}}.
\newblock {\em arXiv}, 2019.

\bibitem{ring1997}
Mark~B Ring.
\newblock {{CHILD}}: {{A First Step Towards Continual Learning}}.
\newblock {\em Machine Learning}, 28(1):77--104, 1997.

\bibitem{robins1995}
Anthony Robins.
\newblock Catastrophic {{Forgetting}}; {{Catastrophic Interference}};
  {{Stability}}; {{Plasticity}}; {{Rehearsal}}.
\newblock {\em Connection Science}, 7(2):123--146, 1995.

\bibitem{rolnick2019}
David Rolnick, Arun Ahuja, Jonathan Schwarz, Timothy~P Lillicrap, and Greg
  Wayne.
\newblock Experience {{Replay}} for {{Continual Learning}}.
\newblock In {\em {{NeurIPS}}}, pages 350--360, 2019.

\bibitem{rusu2016}
Andrei~A Rusu, Neil~C Rabinowitz, Guillaume Desjardins, Hubert Soyer, James
  Kirkpatrick, Koray Kavukcuoglu, Razvan Pascanu, and Raia Hadsell.
\newblock Progressive {{Neural Networks}}.
\newblock {\em arXiv}, 2016.

\bibitem{schaferRecurrentNeuralNetworks2006}
Anton~Maximilian Sch{\"a}fer and Hans~Georg Zimmermann.
\newblock Recurrent {{Neural Networks Are Universal Approximators}}.
\newblock In Stefanos~D. Kollias, Andreas Stafylopatis, W{\l}odzis{\l}aw Duch,
  and Erkki Oja, editors, {\em Artificial {{Neural Networks}} \textendash{}
  {{ICANN}} 2006}, Lecture {{Notes}} in {{Computer Science}}, pages 632--640.
  {Springer Berlin Heidelberg}, 2006.

\bibitem{schak2019}
Monika Schak and Alexander Gepperth.
\newblock A {{Study}} on {{Catastrophic Forgetting}} in {{Deep LSTM Networks}}.
\newblock In Igor~V Tetko, V{\v e}ra K{\r{u}}rkov{\'a}, Pavel Karpov, and
  Fabian Theis, editors, {\em Artificial {{Neural Networks}} and {{Machine
  Learning}} \textendash{} {{ICANN}} 2019: {{Deep Learning}}}, Lecture
  {{Notes}} in {{Computer Science}}, pages 714--728. {Springer International
  Publishing}, {Cham}, 2019.

\bibitem{schlimmerIncrementalLearningNoisy1986}
Jeffrey~C. Schlimmer and Richard~H. Granger.
\newblock Incremental learning from noisy data.
\newblock {\em Machine Learning}, 1(3):317--354, 1986.

\bibitem{schwarz2018}
Jonathan Schwarz, Wojciech Czarnecki, Jelena Luketina, Agnieszka
  {Grabska-Barwinska}, Yee~Whye Teh, Razvan Pascanu, and Raia Hadsell.
\newblock Progress \& {{Compress}}: {{A}} scalable framework for continual
  learning.
\newblock In {\em ICML}, pages 4528--4537, 2018.

\bibitem{sodhani2019}
Shagun Sodhani, Sarath Chandar, and Yoshua Bengio.
\newblock Toward {{Training Recurrent Neural Networks}} for {{Lifelong
  Learning}}.
\newblock {\em Neural Computation}, 32(1):1--35, 2019.

\bibitem{sokar2020}
Ghada Sokar, Decebal~Constantin Mocanu, and Mykola Pechenizkiy.
\newblock {{SpaceNet}}: {{Make Free Space For Continual Learning}}.
\newblock {\em arXiv}, 2020.

\bibitem{sun2020}
Fan-Keng Sun, Cheng-Hao Ho, and Hung-Yi Lee.
\newblock {{LAMOL}}: {{LAnguage MOdeling}} for {{Lifelong Language Learning}}.
\newblock In {\em {{ICLR}}}, 2020.

\bibitem{tang2020}
Binh Tang and David~S. Matteson.
\newblock Graph-{{Based Continual Learning}}.
\newblock In {\em ICLR}, 2020.

\bibitem{thompson2019a}
Brian Thompson, Jeremy Gwinnup, Huda Khayrallah, Kevin Duh, and Philipp Koehn.
\newblock Overcoming {{Catastrophic Forgetting During Domain Adaptation}} of
  {{Neural Machine Translation}}.
\newblock In {\em Proceedings of the 2019 {{Conference}} of the {{North
  American Chapter}} of the {{Association}} for {{Computational Linguistics}}:
  {{Human Language Technologies}}, {{Volume}} 1 ({{Long}} and {{Short
  Papers}})}, pages 2062--2068, {Minneapolis, Minnesota}, 2019. {Association
  for Computational Linguistics}.

\bibitem{toneva2019}
Mariya Toneva, Alessandro Sordoni, Remi~Tachet {des Combes}, Adam Trischler,
  Yoshua Bengio, and Geoffrey~J Gordon.
\newblock An {{Empirical Study}} of {{Example Forgetting}} during {{Deep Neural
  Network Learning}}.
\newblock In {\em ICLR}, 2019.

\bibitem{vandeven2020}
Gido~M. {van de Ven}, Hava~T. Siegelmann, and Andreas~S. Tolias.
\newblock Brain-inspired replay for continual learning with artificial neural
  networks.
\newblock {\em Nature Communications}, 11, 2020.

\bibitem{vandeven2018}
Gido~M. {van de Ven} and Andreas~S. Tolias.
\newblock Generative replay with feedback connections as a general strategy for
  continual learning.
\newblock {\em arXiv}, 2018.

\bibitem{vandeven2018a}
Gido~M {van de Ven} and Andreas~S Tolias.
\newblock Three scenarios for continual learning.
\newblock In {\em Continual {{Learning Workshop NeurIPS}}}, 2018.

\bibitem{vaswaniAttentionAllYou2017}
Ashish Vaswani, Noam Shazeer, Niki Parmar, Jakob Uszkoreit, Llion Jones,
  Aidan~N Gomez, {\L}ukasz Kaiser, and Illia Polosukhin.
\newblock Attention is {{All}} you {{Need}}.
\newblock In I.~Guyon, U.~V. Luxburg, S.~Bengio, H.~Wallach, R.~Fergus,
  S.~Vishwanathan, and R.~Garnett, editors, {\em NIPS}, pages 5998--6008.
  {Curran Associates, Inc.}, 2017.

\bibitem{oswaldContinualLearningHypernetworks2019}
Johannes von Oswald, Christian Henning, Jo\&\#xE3, O~Sacramento, and
  Benjamin~F. Grewe.
\newblock Continual learning with hypernetworks.
\newblock In {\em ICLR}, 2019.

\bibitem{waibel1989}
A.~Waibel, T.~Hanazawa, G.~Hinton, K.~Shikano, and K.~J. Lang.
\newblock Phoneme recognition using time-delay neural networks.
\newblock {\em IEEE Transactions on Acoustics, Speech, and Signal Processing},
  37(3):328--339, 1989.

\bibitem{wang2019}
Zhepei Wang, Cem Subakan, Efthymios Tzinis, Paris Smaragdis, and Laurent
  Charlin.
\newblock Continual {{Learning}} of {{New Sound Classes}} using {{Generative
  Replay}}.
\newblock {\em arXiv}, 2019.

\bibitem{widmer1996}
Gerhard Widmer and Miroslav Kubat.
\newblock Learning in the presence of concept drift and hidden contexts.
\newblock {\em Machine Learning}, 23(1):69--101, 1996.

\bibitem{williamsBroadCoverageChallengeCorpus2018}
Adina Williams, Nikita Nangia, and Samuel Bowman.
\newblock A {{Broad}}-{{Coverage Challenge Corpus}} for {{Sentence
  Understanding}} through {{Inference}}.
\newblock In {\em Proceedings of the 2018 {{Conference}} of the {{North
  American Chapter}} of the {{Association}} for {{Computational Linguistics}}:
  {{Human Language}} {{Technologies}}, {{Volume}} 1 ({{Long Papers}})}, pages
  1112--1122, {New Orleans, Louisiana}, 2018. {Association for Computational
  Linguistics}.

\bibitem{wolf2018}
Thomas Wolf, Julien Chaumond, and Clement Delangue.
\newblock Continuous {{Learning}} in a {{Hierarchical Multiscale Neural
  Network}}.
\newblock In {\em {{ACL}}}, 2018.

\bibitem{xue2019}
Jiabin Xue, Jiqing Han, Tieran Zheng, Xiang Gao, and Jiaxing Guo.
\newblock A {{Multi}}-{{Task Learning Framework}} for {{Overcoming}} the
  {{Catastrophic Forgetting}} in {{Automatic Speech Recognition}}.
\newblock {\em arXiv}, 2019.

\bibitem{yoon2018}
Jaehong Yoon, Eunho Yang, Jeongtae Lee, and Sung~Ju Hwang.
\newblock Lifelong {{Learning With Dynamically Expandable Networks}}.
\newblock In {\em {{ICLR}}}, 2018.

\bibitem{youngRecentTrendsDeep2018}
T~Young, Devamanyu Hazarika, S~Poria, and E~Cambria.
\newblock Recent {Trends} in {Deep} {Learning} {Based} {Natural} {Language}
  {Processing}.
\newblock {\em IEEE Computational Intelligence Magazine}, 13:55--75, 2018.

\bibitem{zenke2017}
Friedemann Zenke, Ben Poole, and Surya Ganguli.
\newblock Continual {{Learning Through Synaptic Intelligence}}.
\newblock In {\em ICML}, pages 3987--3995, 2017.

\bibitem{zeno2018}
Chen Zeno, Itay Golan, Elad Hoffer, and Daniel Soudry.
\newblock Task {{Agnostic Continual Learning Using Online Variational Bayes}}.
\newblock In {\em {{NeurIPS Bayesian Deep Learning Workshop}}}, 2018.

\end{thebibliography}

\appendix

\section{Experiment configuration} \label{app:conf}
Here we report additional details on the experiments configuration.\\
Each step is divided into a training set and a test set. We used the default train-test split of Torchvision for MNIST-based datasets. For SSC we randomly generate a train-test split for each class, with $20\%$ of patterns for the test set. Quick, Draw! already provides separate splits for each class. We changed random seed on each run. \\
To make training stabler with RNNs, we used a fixed permutation on SMNIST which does not change across steps. This uniformly spreads the information along the sequence. \\
Following the original paper, A-GEM concatenates a one-hot task vector to the input patterns. Since we adopt class-incremental scenarios, we concatenate the vector only at training time. On SSC, we leveraged the Avalanche library \cite{lomonaco2021} for the GEM implementation. \\ 
We tested a wide range of hyperparameters for the grid search, as showed by Table \ref{tab:hyperparpmnist} for PMNIST, Table \ref{tab:hyperparsmnist} for SMNIST, Table \ref{tab:hyperparssc} for SSC and Table \ref{tab:hyperparqd} for QD. The tables use the following abbreviations: number of layers is \textit{nl}, number of hidden units is \textit{hs}, minibatch size is \textit{mbs}, optimizer is \textit{opt}, learning rate is \textit{lr}, patterns per step in GEM memory is \textit{pps}, sample size from A-GEM memory is \textit{ss}, softmax temperature in LwF is \textit{T}.
MLPs are with ReLU activations. LSTMs have 1 layer for MNIST-based benchmarks, 2 layers for SSC and Quick, Draw!. LSTM with QD uses $512$ hidden units. Multi-head experiments use the same configuration. \\
All models use gradient clipping at norm $5$. The number of epochs is set to guarantee convergence except for online methods (GEM, A-GEM) which use $2$ epochs and minibatches of $10$ patterns. The default optimizer is Adam with no regularization. \\

\begin{table}[t]
    \centering
    \begin{tabularx}{\textwidth}{c|Y|Y}
    \toprule
\multicolumn{1}{c}{\textbf{PMNIST}}  & \multicolumn{1}{c}{LSTM} & \multicolumn{1}{c}{MLP}  \\
\midrule
    EWC & $\lambda$=(0.1, \textbf{1}, 10, 100), hs=(\textbf{256}, 512)           & $\lambda$=(\textbf{0.1}, 1, 10, 100), nl=(1,\textbf{2}), hs=512   \\ \midrule
    MAS   & $\lambda$=(0.1, \textbf{1}, 10, 100), mbs=(\textbf{64},128), hs=(\textbf{256}, 512)         & $\lambda$=(0.1, 1, \textbf{10}), mbs=(32,\textbf{64},128), lr=(1e-2, \textbf{1e-3}), nl=1, hs=512       \\ \midrule
    GEM    & pps=(64, 128, \textbf{256}), $\gamma$=(0.5, \textbf{1}), lr=(\textbf{1e-2}, 1e-3), hs=256        & pps=(64, \textbf{128}, 256), $\gamma$=(\textbf{0.5}, 1), lr=(\textbf{1e-1}, 1e-2, 1e-3), nl=(1,\textbf{2}), hs=512      \\ \midrule
    A-GEM   & patterns per step i memory=(64, \textbf{128}, 256), ss=(256, \textbf{512}), lr=(\textbf{1e-2}, 1e-3), hs=256       & pps=(\textbf{64}, 128, 256), ss=(\textbf{256}, 512), lr=(\textbf{1e-1}, 1e-2, 1e-3), nl=(\textbf{1},2), hs=5120        \\ \midrule
    LwF     & $\alpha$=(0.1, \textbf{1}), hs=256, T=(0.5, 1, \textbf{1.5}, 2), opt=(sgd, \textbf{adam}), lr=(\textbf{1e-3}, 1e-4)  &  $\alpha$=(0.1, \textbf{1}), nl=(1,\textbf{2}), hs=512, T=(0.5, 1, 1.5, \textbf{2}), opt=(\textbf{sgd}, adam), mbs=(\textbf{64}, 128), lr=(\textbf{1e-2}, 1e-3)      \\ \midrule
    Replay  & lr=(\textbf{1e-3}, 1e-4), hs=(\textbf{256}, 512)      & lr=(\textbf{1e-3}, 1e-4), hs=(128,\textbf{256}), nl=1          \\ \midrule
    Naive &  hs=(\textbf{256}, 512), lr=(\textbf{1e-3}, 1e-4)          & hs=(256,\textbf{512}), nl=1         \\ \midrule
    Joint Training &  lr=(1e-2, \textbf{1e-3}), mbs=128, hs=512   & hs=(256, \textbf{512}), nl=1         \\ \bottomrule
    \end{tabularx}
    \caption{Hyperparameter selection on PMNIST. Grid search has been performed for all the combinations in parenthesis. Bold notation indicates best hyperparameter value. See the text in \ref{app:conf} for explanation of the abbreviations used in this table.}
    \label{tab:hyperparpmnist}
\end{table}

\begin{table}[t]
    \centering
    \begin{tabularx}{\textwidth}{c|Y|Y}
    \toprule
\multicolumn{1}{c}{\textbf{SMNIST}}  & \multicolumn{1}{c}{LSTM} & \multicolumn{1}{c}{MLP}  \\
\midrule
    EWC & $\lambda$=(\textbf{0.1}, 1, 10, 100, 1000), hs=(128, \textbf{256}) & $\lambda$=(0.1, 1, 10, \textbf{100}, 1000), nl=(1,\textbf{2}) hs=128    \\ \midrule
    MAS & $\lambda$=(0.1, 1, \textbf{10}, 100, 1000), hs=(\textbf{128}, 256) & $\lambda$=(0.1, 1, \textbf{10}, 100), lr=(1e-2, \textbf{1e-3}), nl=1, hs=128 \\ \midrule
    GEM & pps=(64, 128, \textbf{256}), $\gamma$=(\textbf{0.5}, 1), lr=(\textbf{1e-2}, 1e-3), hs=128 & pps=(64, 128, \textbf{256}), $\gamma$=(\textbf{0.5}, 1), lr=(1e-1, \textbf{1e-2}, 1e-3), hs=128, nl=1 \\ \midrule
    A-GEM & pps=(\textbf{128}, 256, 512), ss=(256, \textbf{512}), lr=(\textbf{1e-2}, 1e-3), hs=128 & pps=(128, \textbf{256}, 512), ss=(256, \textbf{512}), lr=(\textbf{1e-1}, 1e-2, 1e-3), hs=512, nl=1 \\ \midrule
    LwF & $\alpha$=([0, 1/2, 2*(2/3), 3*(3/4), 4*(4/5)], \textbf{1}), hs=(256,\textbf{512}), T=(\textbf{0.5}, 1, 1.5, 2), opt=(sgd, \textbf{adam}), lr=(1e-3, \textbf{1e-4}), mbs=(\textbf{64},128) & $\alpha$=(\textbf{[0, 1/2, 2*(2/3), 3*(3/4), 4*(4/5)]}, 1), nl=(\textbf{1},2), hs=256, T=(0.5, \textbf{1}, 1.5, 2), opt=(\textbf{sgd}, adam), mbs=(64, \textbf{128}), lr=(1e-2, \textbf{1e-3}) \\ \midrule
    Replay & lr=(1e-3, \textbf{1e-4}), hs=(128, \textbf{256}) & hs=128, nl=(\textbf{1},2) \\ \midrule
    Naive & hs=128, mbs=(\textbf{32},64), lr=(\textbf{1e-3}, 1e-4) & hs=(\textbf{128},256), nl=1 \\ \midrule
    Joint Training & hs=512, mbs=128, lr=(\textbf{1e-3}, 1e-4) & hs=128, nl=(1,\textbf{2}) \\ \bottomrule
    \end{tabularx}
    \caption{Hyperparameter selection on SMNIST. Grid search has been performed for all the combinations in parenthesis. Bold notation indicates best hyperparameter value. See the text in \ref{app:conf} for explanation of the abbreviations used in this table.}
    \label{tab:hyperparsmnist}
\end{table}

\begin{table}[t]
    \centering
    \begin{tabularx}{\textwidth}{c|Y|Y}
    \toprule
\multicolumn{1}{c}{\textbf{SSC}}  & \multicolumn{1}{c}{LSTM} & \multicolumn{1}{c}{MLP}  \\
\midrule
    EWC & $\lambda$=(0.1, 1, 10, 100, \textbf{1000}), hs=512 & $\lambda$=(0.1, \textbf{1}, 10, 100, 1000), lr=(\textbf{1e-2}, 1e-3) nl=1, hs=1024 \\ \midrule
    MAS & $\lambda$=(\textbf{0.1}, 1, 10, 100), lr=(1e-3, \textbf{1e-4}), hs=512, mbs=(32, \textbf{64}, 128) & $\lambda$=(\textbf{0.1}, 1, 10, 100), mbs=(32,64,\textbf{128}), lr=(\textbf{1e-2}, 1e-3), nl=1, hs=1024 \\ \midrule
    GEM & pps=(256, \textbf{512}), $\gamma$=(0, \textbf{0.5}, 1) & pps=(\textbf{256}, 512), $\gamma$=(0, 0.5, \textbf{1})  \\ \midrule
    A-GEM & pps=(64, 128, \textbf{256}), ss=(256, \textbf{512}), lr=(1e-1, \textbf{1e-2}, 1e-3), hs=512 & pps=(64, \textbf{128}, 256), ss=(\textbf{256}, 512), lr=(1e-1, \textbf{1e-2}, 1e-3), nl=1, hs=1024 \\ \midrule
    LwF & $\alpha$=(0.1, \textbf{1}), hs=(512, \textbf{1024}), T=(0.5, \textbf{1}, 1.5, 2), opt=(sgd, \textbf{adam}), lr=(1e-3, \textbf{1e-4}), mbs=(64,\textbf{128}) & $\alpha$=(0.1, \textbf{1}), nl=(\textbf{1},2), hs=1024, T=(0.5, 1, \textbf{1.5}, 2), opt=(sgd, \textbf{adam}), mbs=(\textbf{64}, 128), lr=(\textbf{1e-2}, 1e-3) \\ \midrule
    Replay & hs=(\textbf{512}, 1024) & lr=(1e-2, \textbf{1e-3}), hs=1024, nl=(\textbf{1},2) \\ \midrule
    Naive & hs=(\textbf{512}, 1024) & hs=1024, nl=(1,\textbf{2},3), lr=(1e-2, \textbf{1e-3}) \\ \midrule
    Joint Training & lr=(0.01, \textbf{0.001}), hs=(\textbf{512}, 1024), layers=2, mbs=128 & hs=1024, nl=(\textbf{1},2), weight decay=(0, \textbf{1e-4}), lr=1e-3, mbs=128 \\ \bottomrule    
    \end{tabularx}
    \caption{Hyperparameter selection on SSC. Grid search has been performed for all the combinations in parenthesis. Bold notation indicates best hyperparameter value. See the text in \ref{app:conf} for explanation of the abbreviations used in this table.}
    \label{tab:hyperparssc}
\end{table}

\begin{table}[t]
    \centering
    \begin{tabularx}{\textwidth}{c|Y}
    \toprule
\multicolumn{1}{c}{\textbf{QD}}  & \multicolumn{1}{c}{LSTM}  \\
\midrule
    EWC & $\lambda$=(1, 10, \textbf{100}) \\ \midrule
    MAS & $\lambda$=(1, 10, \textbf{100}) \\ \midrule
    GEM & pps=(32, \textbf{64}), $\gamma$=(0, \textbf{0.5}), lr=(1e-2, \textbf{1e-3}) \\ \midrule
    A-GEM & pps=(\textbf{64}, 128, 256), ss=(256, \textbf{512}), lr=(1e-1, \textbf{1e-2}) \\ \midrule
    LwF & $\alpha$=(0.1, \textbf{1}), hs=512, T=(0.5, \textbf{1}, 1.5, 2), opt=(sgd, \textbf{adam}), lr=1e-4, mbs=(64,\textbf{128}) \\ \midrule
    Replay & hs=512, lr=(1e-3, \textbf{1e-4}) \\ \midrule
    Naive & hs=512, layers=(1,\textbf{2}), directions=(\textbf{1}, 2) \\ \midrule
    Joint Training & hs=(256, \textbf{512}), layers=(1,\textbf{2}), bidirectional=(true, \textbf{false}), mbs=512, lr=1e-4 \\ \bottomrule
    \end{tabularx}
    \caption{Hyperparameter selection on QD. Grid search has been performed for all the combinations in parenthesis. Bold notation indicates best hyperparameter value. See the text in \ref{app:conf} for explanation of the abbreviations used in this table.}
    \label{tab:hyperparqd}
\end{table}

\subsection{Computational Complexity} \label{app:times}

We monitored execution times for recurrent architectures with different CL strategies. We ran experiments on a single V100 GPU, with $3$ threads at a time on a Intel\textsuperscript{\textregistered} Xeon\textsuperscript{\textregistered} Gold 6140M CPU with 2.30GHz frequency. \\
Our results clearly show that the cost of GEM is quite large and may be prohibitive in realistic applications. \\

\begin{figure}[t]
    \centering
    \begin{subfigure}[t]{0.4\textwidth}
    \centering
    \includegraphics[width=\textwidth]{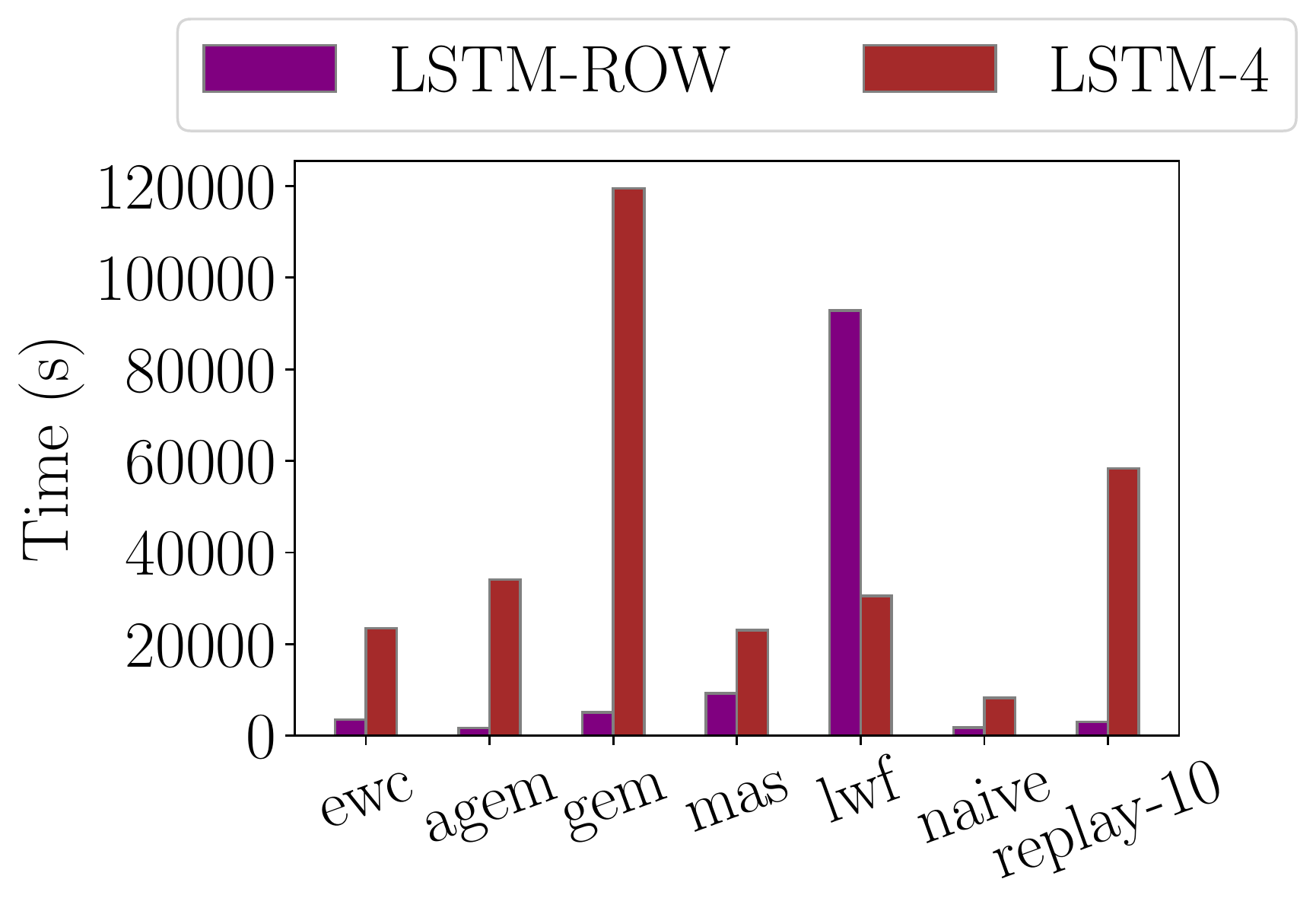}
    \caption{Permuted MNIST}
    \label{fig:timespmnist}
    \end{subfigure}
    \begin{subfigure}[t]{0.4\textwidth}
    \centering
    \includegraphics[width=\textwidth]{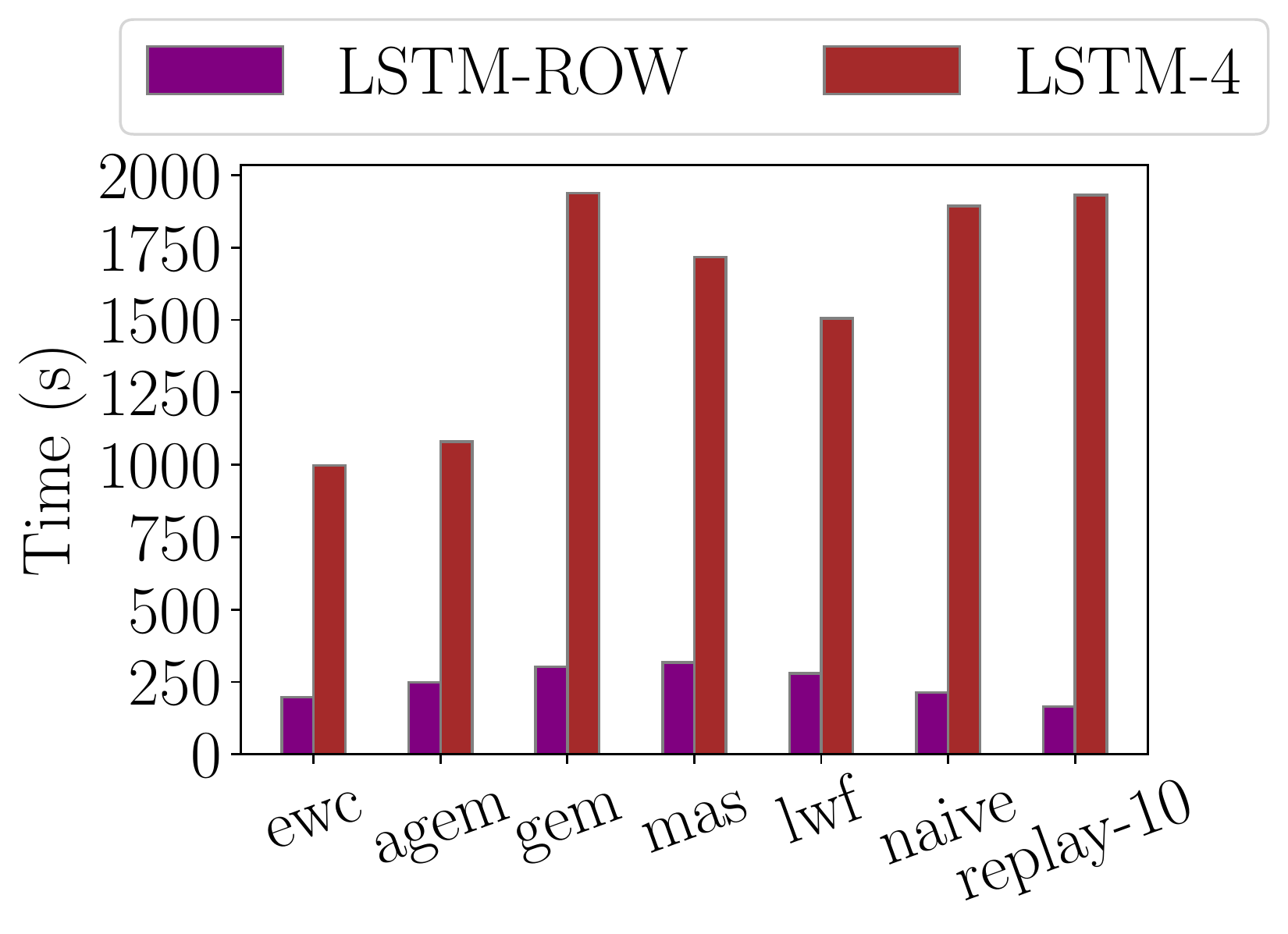}
    \caption{Split MNIST}
    \label{fig:timesmnist}
    \end{subfigure}
    \begin{subfigure}[t]{0.4\textwidth}
    \centering
    \includegraphics[width=\textwidth]{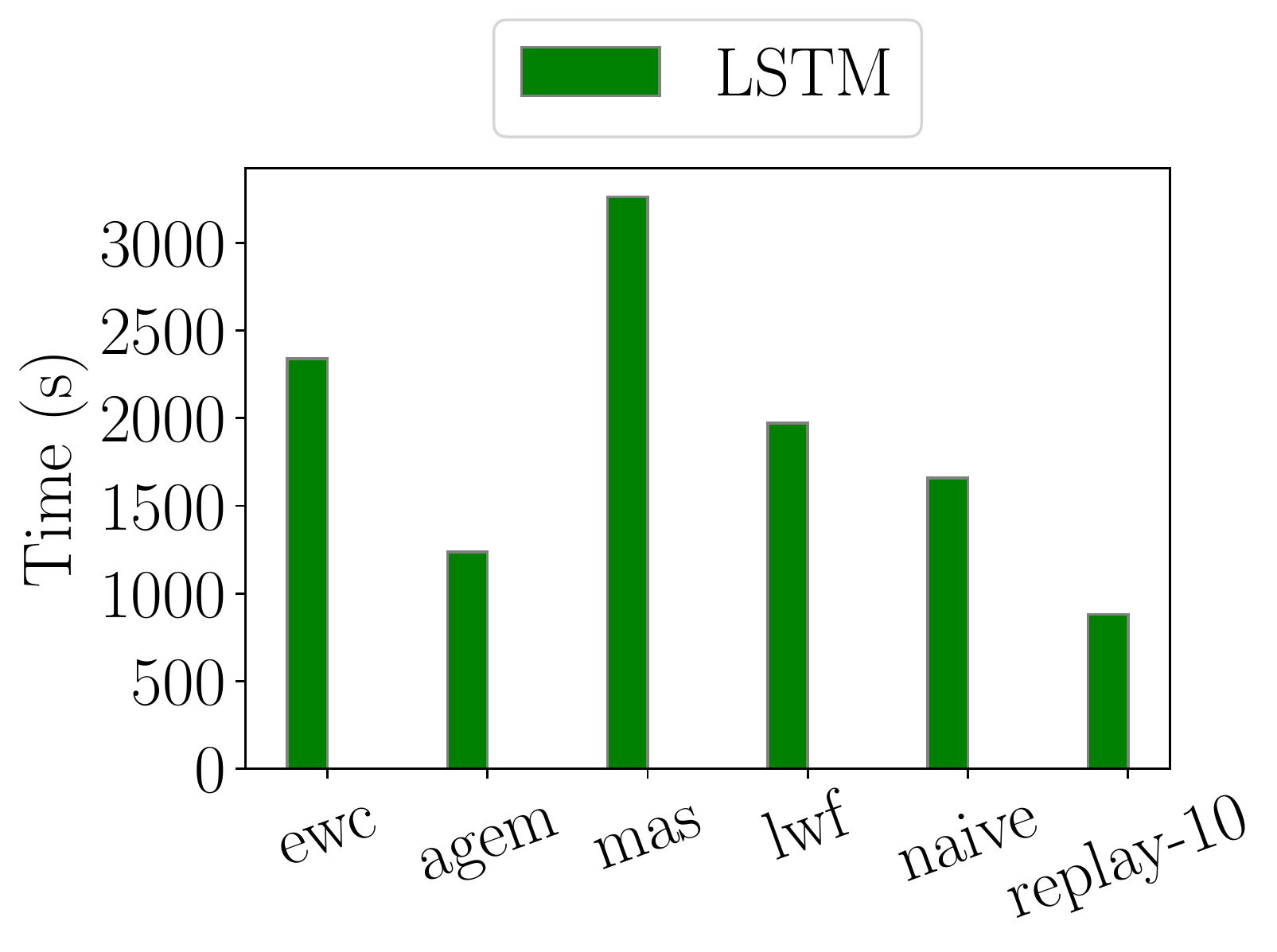}
    \caption{SSC}
    \label{fig:timessc}
    \end{subfigure}
    \begin{subfigure}[t]{0.4\textwidth}
    \centering
    \includegraphics[width=\textwidth]{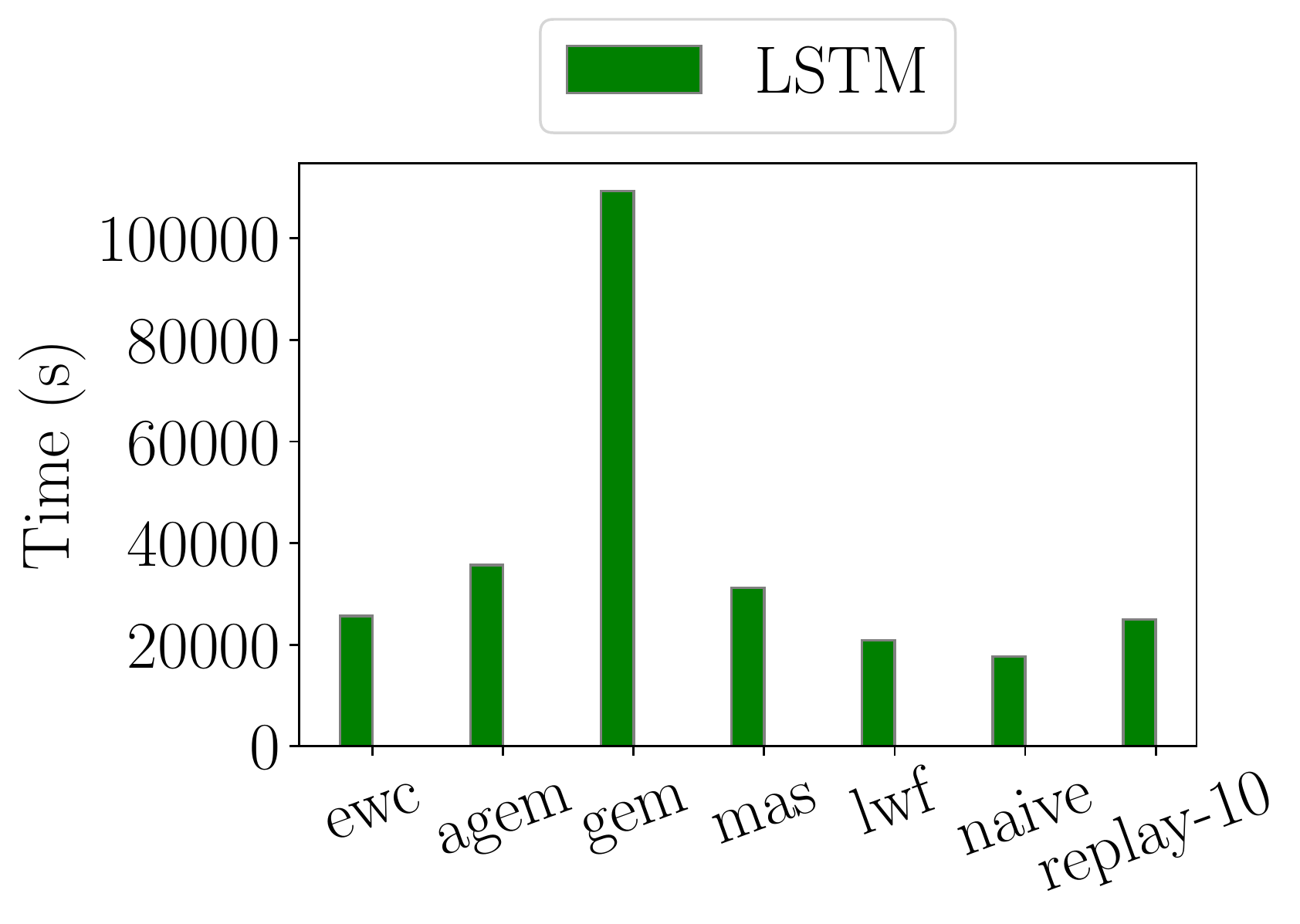}
    \caption{QD}
    \label{fig:timeqd}
    \end{subfigure}
    \caption{Average execution time of recurrent models for different CL strategies. On SSC, we did not include the GEM strategy since it has been executed with a different framework. Best viewed in color.}
 \label{fig:times}
 \end{figure}
 
\section{Benchmarks Description}
\subsection{Synthetic Speech Commands} \label{app:ssc}
We introduced the Synthetic Speech Commands (SSC) \cite{speechcommands} benchmark\footnote{The Synthetic Speec Commands Dataset is available at \url{https://www.kaggle.com/jbuchner/synthetic-speech-commands-dataset}} and adapted it for a class-incremental setting. We took $2$ classes at a time and we created sequences of $3$ steps for model selection and sequences of $10$ steps for model assessment. We preprocessed each audio with a MelSpectrogram with $10$ ms hop length, $25$ ms window size and $40$ MFCCs features. \\
Table \ref{tab:sscdetails} following, we report statistics for all the $30$ classes in the dataset. \\

\begin{table}[t]
    \centering
    \begin{tabular}{ccc||c||cccc}
    \toprule
    Class name & patterns & use &  & Class name & patterns & use  \\ 
    \midrule
    bed        & 1356     & S1  &  & off        & 2244     & A3   \\
    bird       & 1346     & S1  &  & one        & 1276     & A4   \\
    cat        & 1378     & S2  &  & on         & 2228     & A4   \\
    dog        & 1474     & S2  &  & right      & 1276     & A5   \\
    down       & 1188     & S3  &  & seven      & 1411     & A5   \\
    eight      & 1113     & S3  &  & sheila     & 1463     & A6   \\
    five       & 1092     & -   &  & six        & 1485     & A6   \\
    four       & 2400     & -   &  & stop       & 1485     & A7   \\
    go         & 960      & -   &  & three      & 1188     & A7   \\
    happy      & 1481     & -   &  & tree       & 1188     & A8   \\
    house      & 2382     & A1  &  & two        & 902      & A8   \\
    left       & 1485     & A1  &  & up         & 1187     & A9   \\
    marvel     & 1253     & A2  &  & wow        & 957      & A9   \\
    nine       & 1144     & A2  &  & yes        & 1244     & A10  \\
    no         & 957      & A3  &  & zero       & 1306     & A10  \\
    \bottomrule
    \end{tabular}
    \caption{Details of SSC dataset classes: name of the class, total number of patterns, how it has been used. S for model selection, A for model assessment, the following number indicates which step uses the class. Dash indicates that the class has not been used in our experiments.}
    \label{tab:sscdetails}
\end{table}

\begin{table}[b]
    \centering
    \begin{tabular}{cc||c||cc}
    \toprule
    Class name & use &  & Class name    & use  \\
    \midrule
    hot dog    & S1  &  & octopus       & A4   \\
    palm tree  & S1  &  & cloud         & A5   \\
    moon       & S2  &  & bicycle       & A5   \\
    envelope   & S2  &  & swan          & A6   \\
    dumbbell   & S3  &  & picture frame & A6   \\
    microwave  & S3  &  & shorts        & A7   \\
    drill      & A1  &  & flying saucer & A7   \\
    telephone  & A1  &  & basketball    & A8   \\
    airplane   & A2  &  & harp          & A8   \\
    dishwasher & A2  &  & beard         & A9   \\
    chair      & A3  &  & binoculars    & A9   \\
    grass      & A3  &  & tiger         & A10  \\
    rhinoceros & A4  &  & book          & A10 \\
    \bottomrule
    \end{tabular}
    \caption{Details of Quick, Draw! dataset classes: name of the class, how it has been used. S for model selection, A for model assessment, the following number indicates which step uses the class.}
    \label{tab:qddetails}
\end{table}

\subsection{Quick, Draw!} \label{app:quickdraw}
Quick, Draw! dataset has been downloaded from Google Cloud Console, in the \textit{sketchrnn} folder\footnote{Quick, Draw! details are available at \url{https://github.com/googlecreativelab/quickdraw-dataset}}. All classes have $70,000$ patterns for training and $2,500$ for test. Table \ref{tab:qddetails} summarizes the classes used in the experiments. 

\section{Additional Results} \label{app:additional}

We provide a set of paired plots showing the relationship between the training accuracy at the end of each step and the final test accuracy at the end of training on all steps. These plots are useful to compare the relative performance of different CL strategies and different architectures. \\

Table \ref{tab:multiheadresults} shows the mean ACC and its standard deviation for different CL strategies with multi-head models in \scenario{MT+NC} scenarios. A-GEM concatenates the one-hot task vector to each train and test input pattern.\\

\begin{table}[t]
\centering
\begin{tabular}{cccc}
                        & &                    \\
\toprule                        
\textbf{SMNIST} & MLP  & LSTM-4  \\
\midrule
 EWC & \meanstd{0.99}{0.00} & \meanstd{0.64}{0.16} \\
 A-GEM & \meanstd{0.73}{0.05} & \meanstd{0.20}{0.06} \\
 NAIVE & \meanstd{0.97}{0.02} & \meanstd{0.72}{0.11} \\
 \bottomrule
& & & \\
\toprule
\textbf{SSC}& MLP  & LSTM     \\
\midrule
 EWC & \meanstd{0.77}{0.09} & \meanstd{0.80}{0.06} \\
 A-GEM & \meanstd{0.57}{0.02} & \meanstd{0.62}{0.07} \\
 NAIVE & \meanstd{0.72}{0.10} & \meanstd{0.67}{0.10} \\
 \bottomrule
 & & & \\
 \toprule
 \textbf{QD} && LSTM \\
 \midrule
 EWC && \meanstd{0.98}{0.00} \\
 A-GEM && \meanstd{0.56}{0.16} \\
 NAIVE && \meanstd{0.71}{0.09} \\
 \bottomrule
\end{tabular}
\caption{Multi-head experiments showing average ACC and standard deviation. Naive strategy already recovers large part of the performance.}\label{tab:multiheadresults}
\end{table}

\begin{figure}[t]
    \centering
\begin{subfigure}[t]{0.4\textwidth}\centering\includegraphics[width=\textwidth]{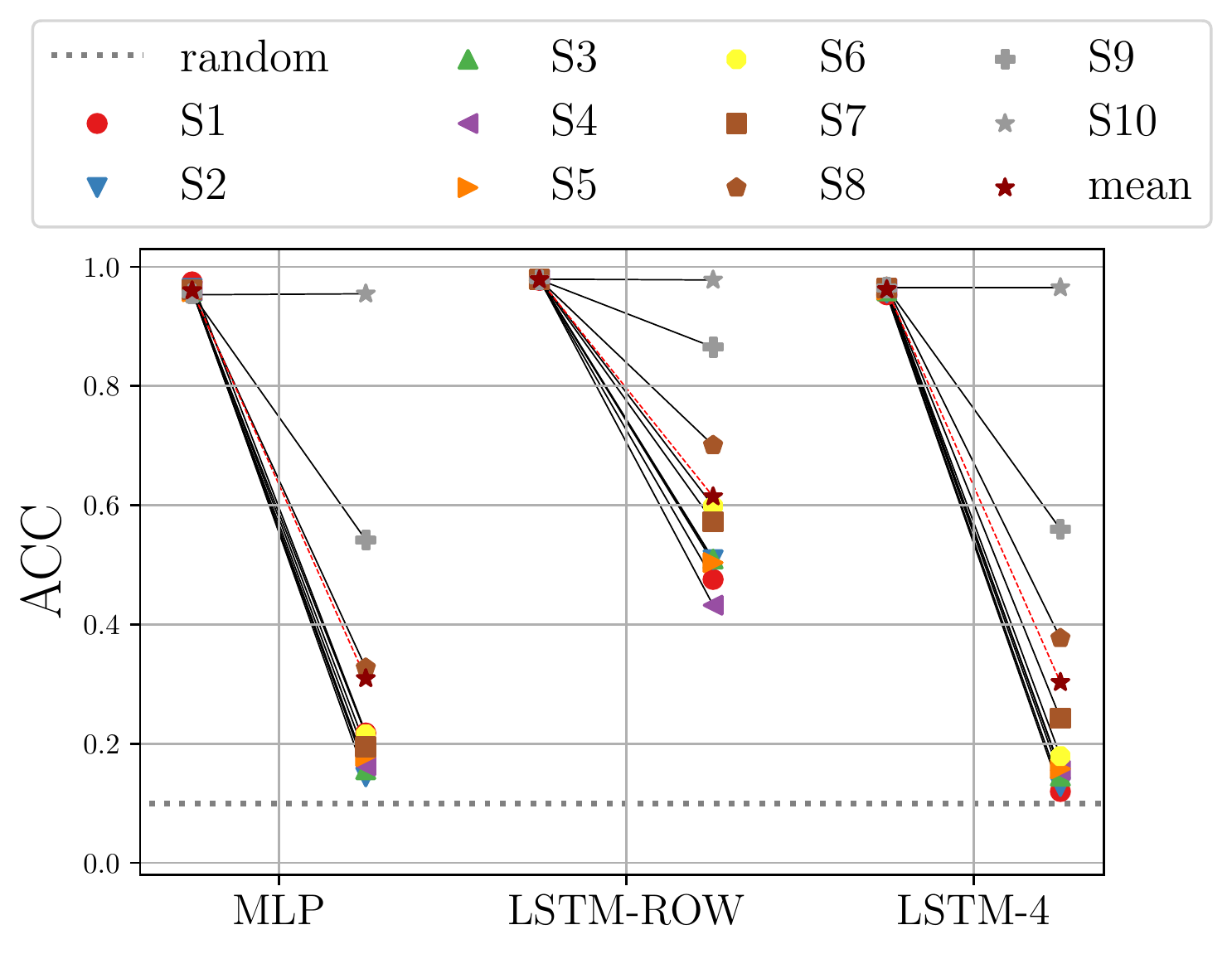}\caption{PMNIST + naive}\end{subfigure}
\begin{subfigure}[t]{0.4\textwidth}\centering\includegraphics[width=\textwidth]{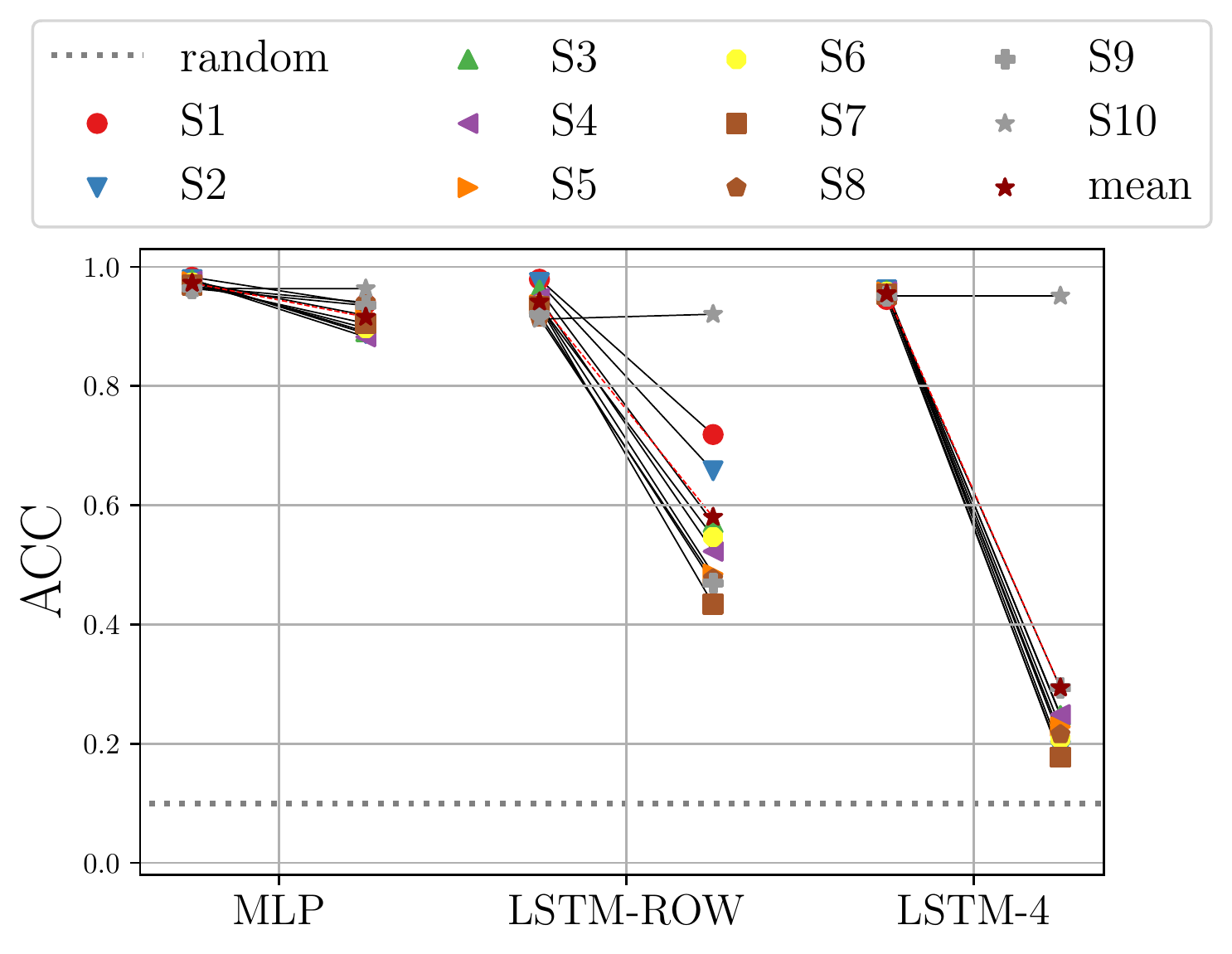}\caption{PMNIST + ewc}\end{subfigure}
\begin{subfigure}[t]{0.4\textwidth}\centering\includegraphics[width=\textwidth]{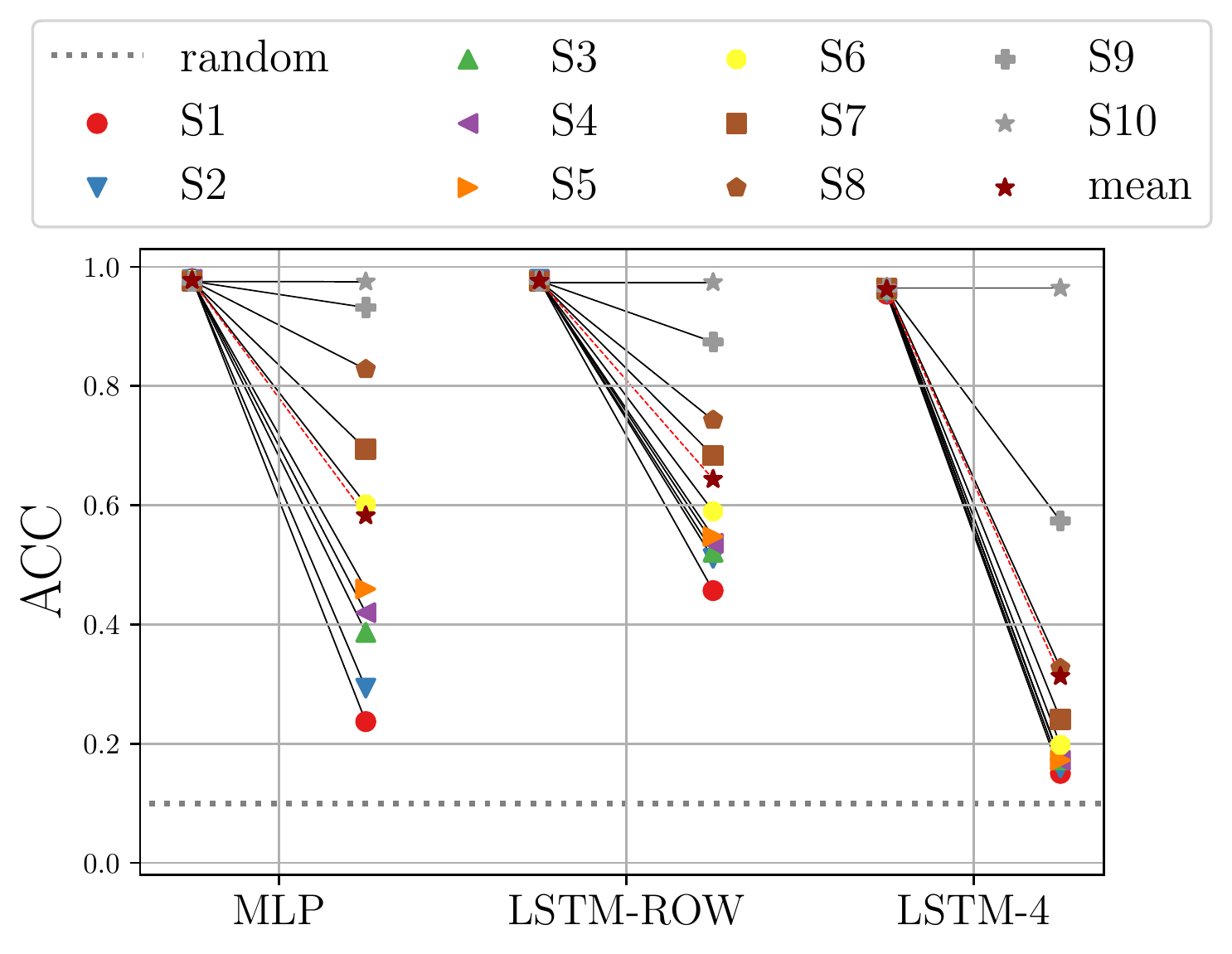}\caption{PMNIST + mas}\end{subfigure}
\begin{subfigure}[t]{0.4\textwidth}\centering\includegraphics[width=\textwidth]{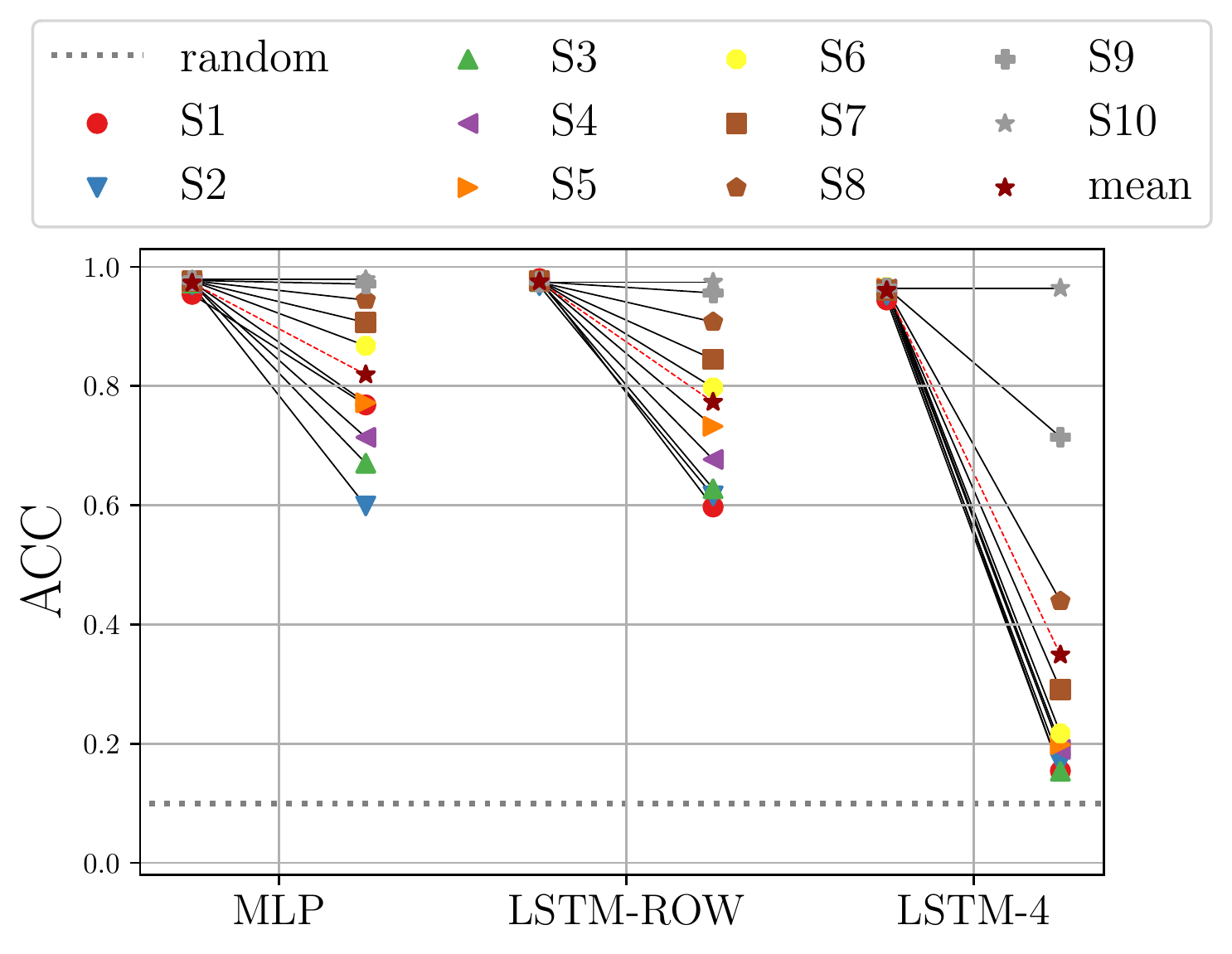}\caption{PMNIST + lwf}\end{subfigure}
\begin{subfigure}[t]{0.4\textwidth}\centering\includegraphics[width=\textwidth]{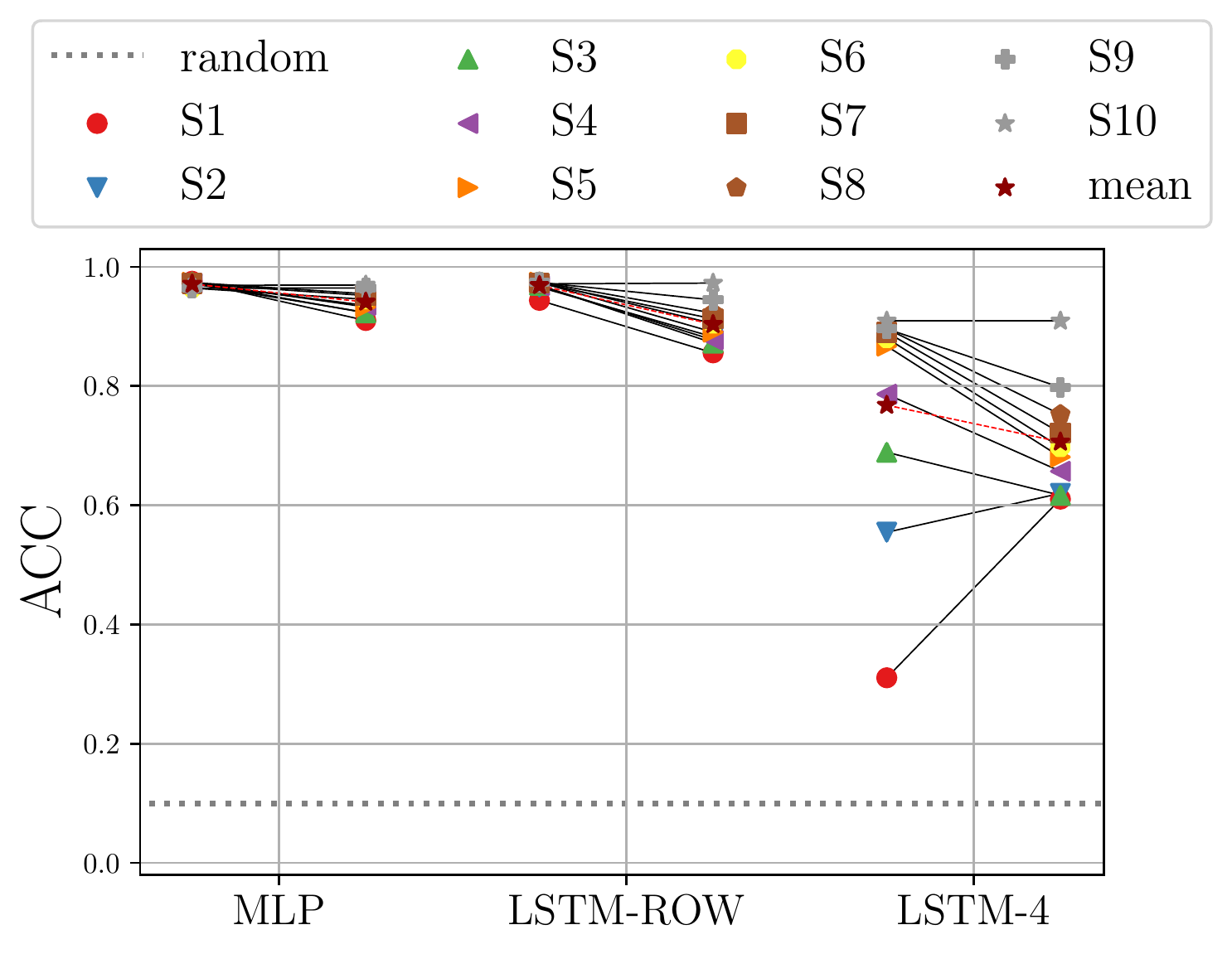}\caption{PMNIST + gem}\end{subfigure}
\begin{subfigure}[t]{0.4\textwidth}\centering\includegraphics[width=\textwidth]{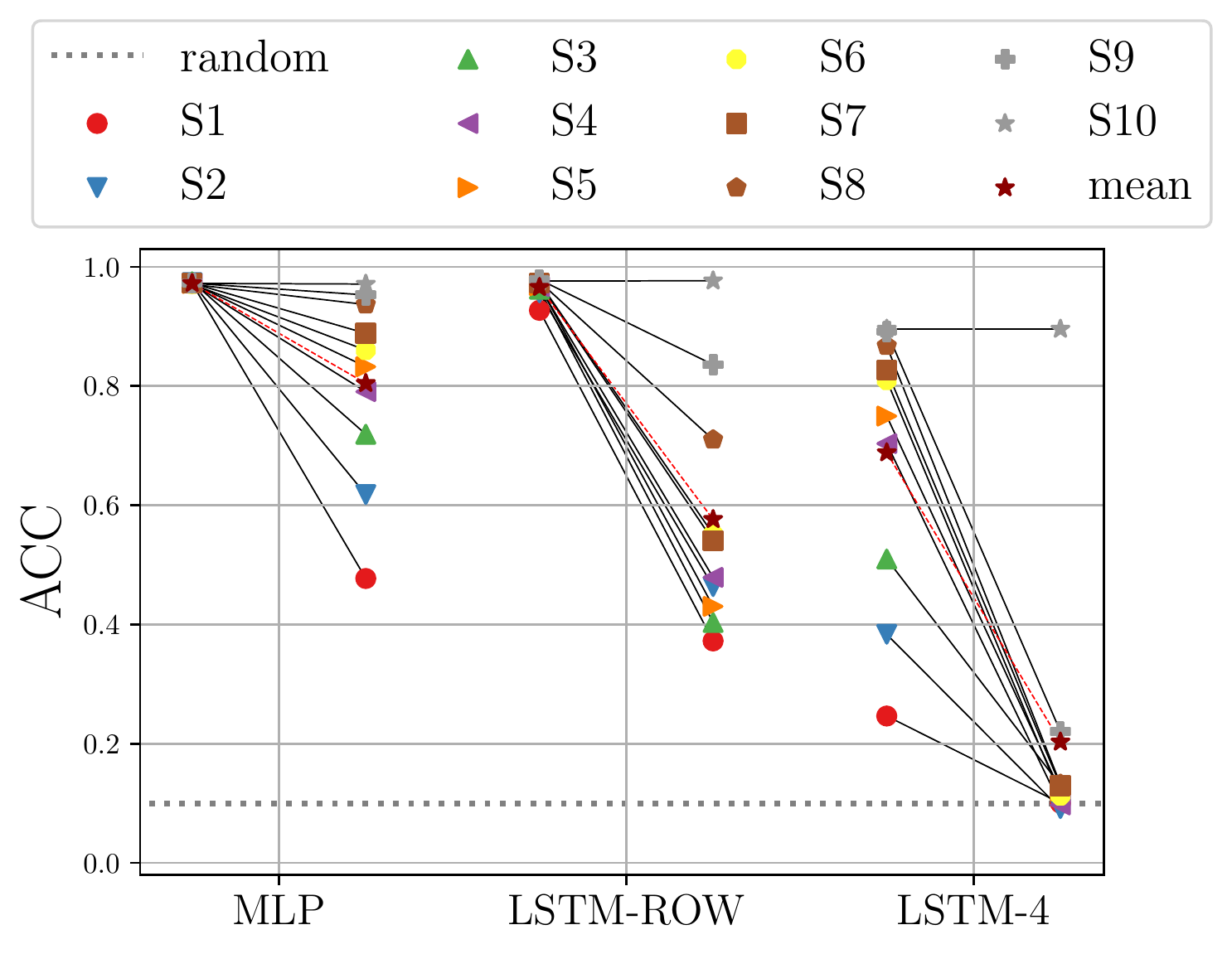}\caption{PMNIST + agem}\end{subfigure}
\begin{subfigure}[t]{0.4\textwidth}\centering\includegraphics[width=\textwidth]{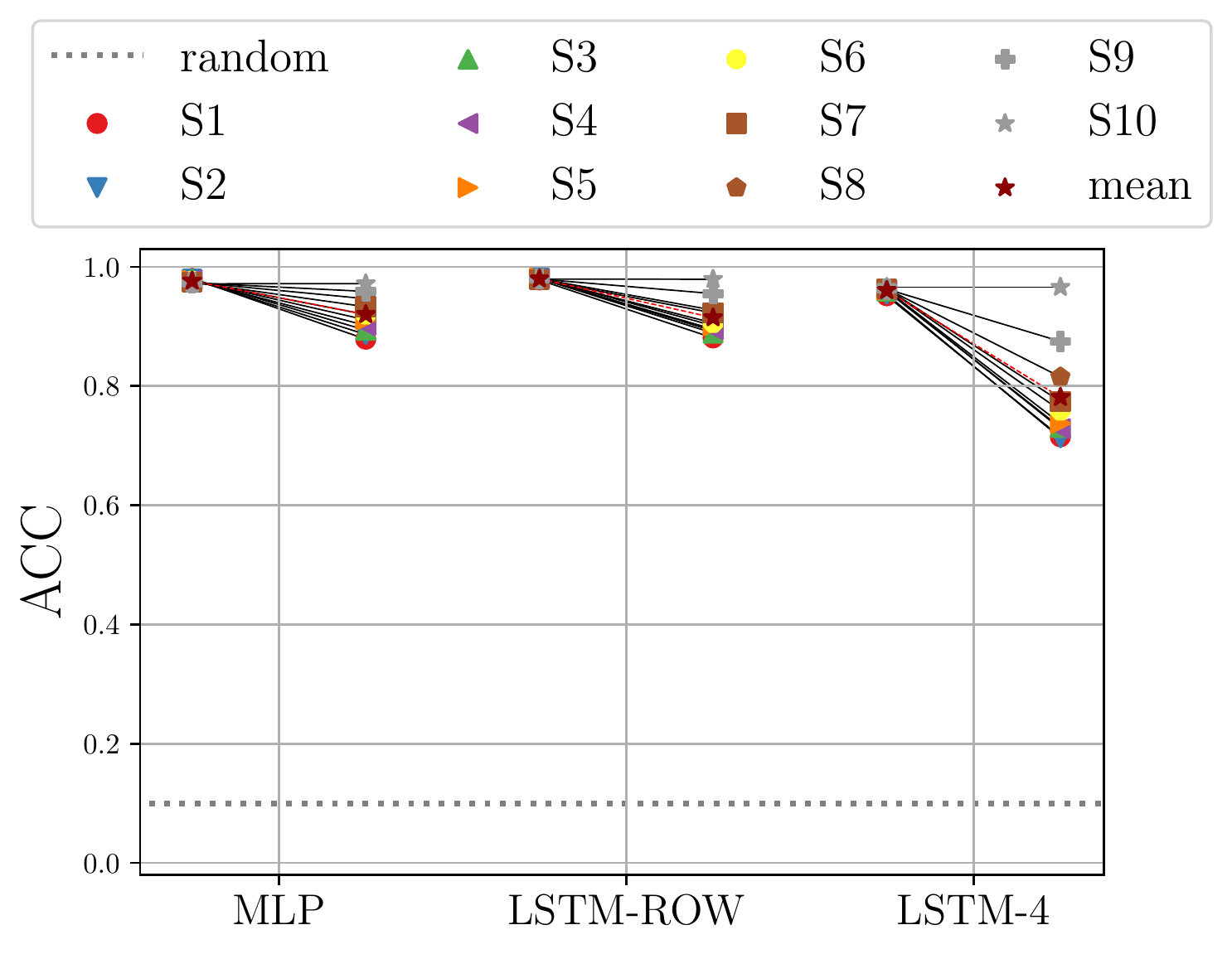}\caption{PMNIST + replay-1}\end{subfigure}
\begin{subfigure}[t]{0.4\textwidth}\centering\includegraphics[width=\textwidth]{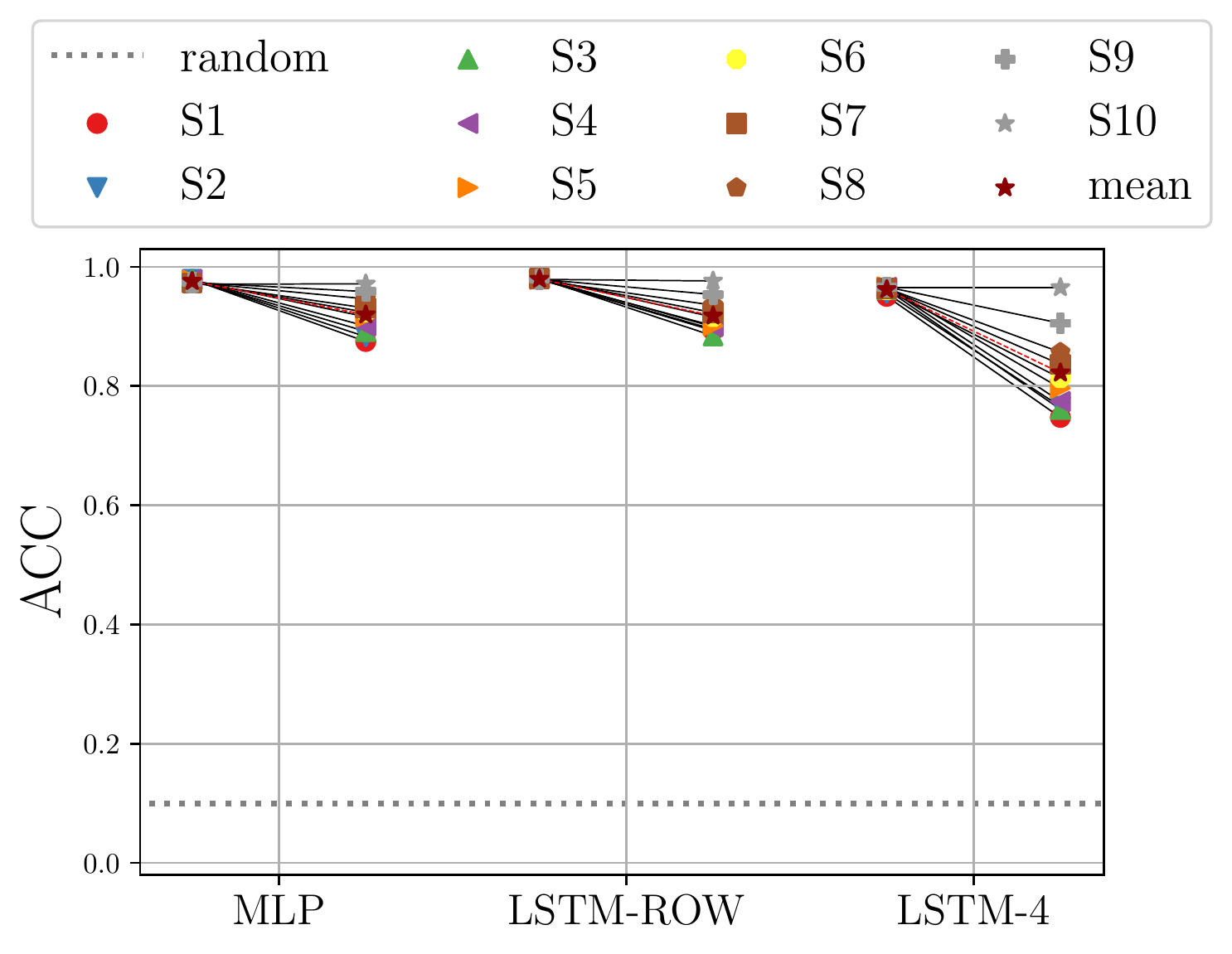}\caption{PMNIST + replay-10}\end{subfigure}
\end{figure}
\begin{figure}
\centering
\ContinuedFloat
\begin{subfigure}[t]{0.4\textwidth}\centering\includegraphics[width=\textwidth]{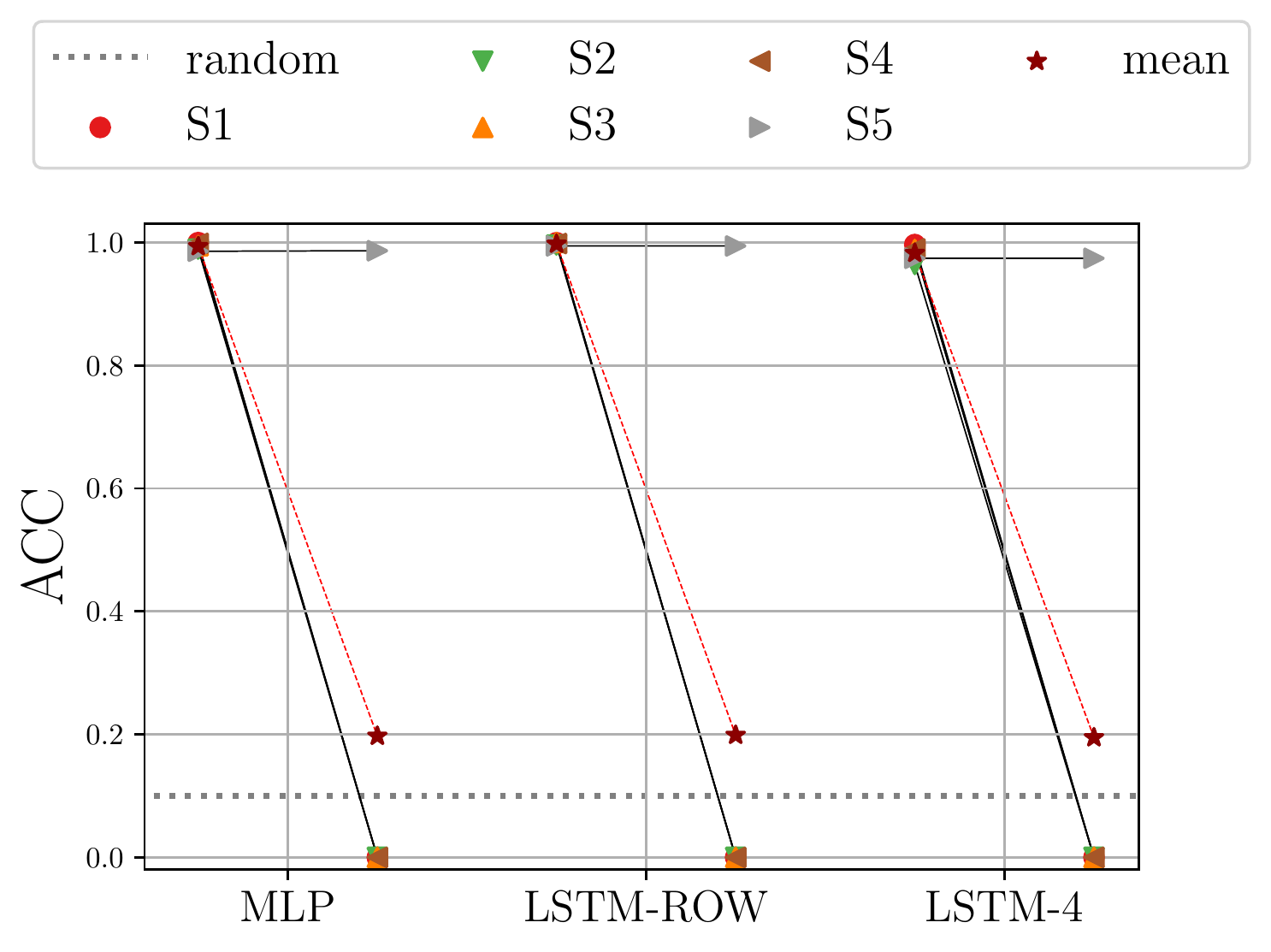}\caption{SMNIST + naive}\end{subfigure}
\begin{subfigure}[t]{0.4\textwidth}\centering\includegraphics[width=\textwidth]{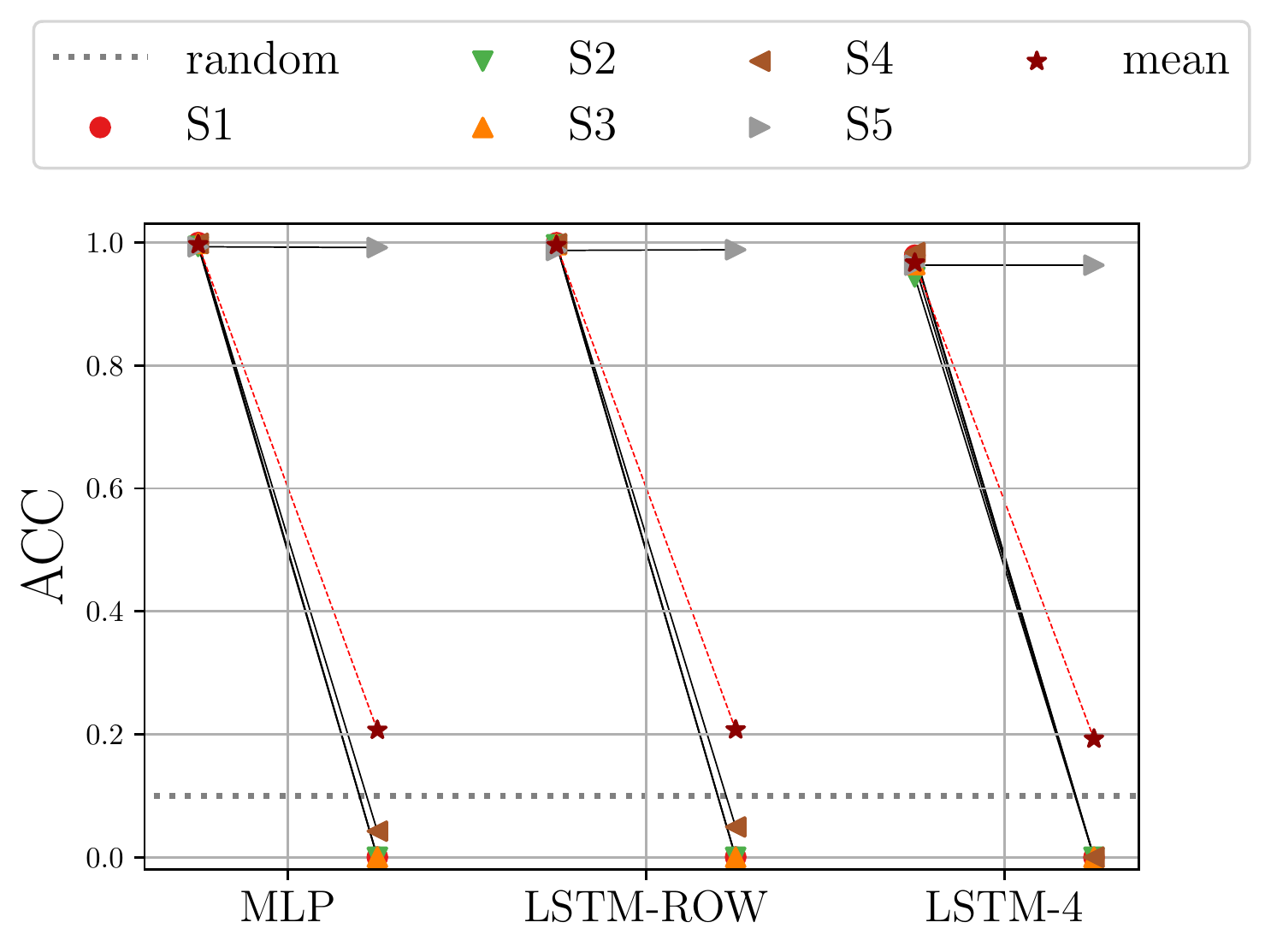}\caption{SMNIST + ewc}\end{subfigure}
\begin{subfigure}[t]{0.4\textwidth}\centering\includegraphics[width=\textwidth]{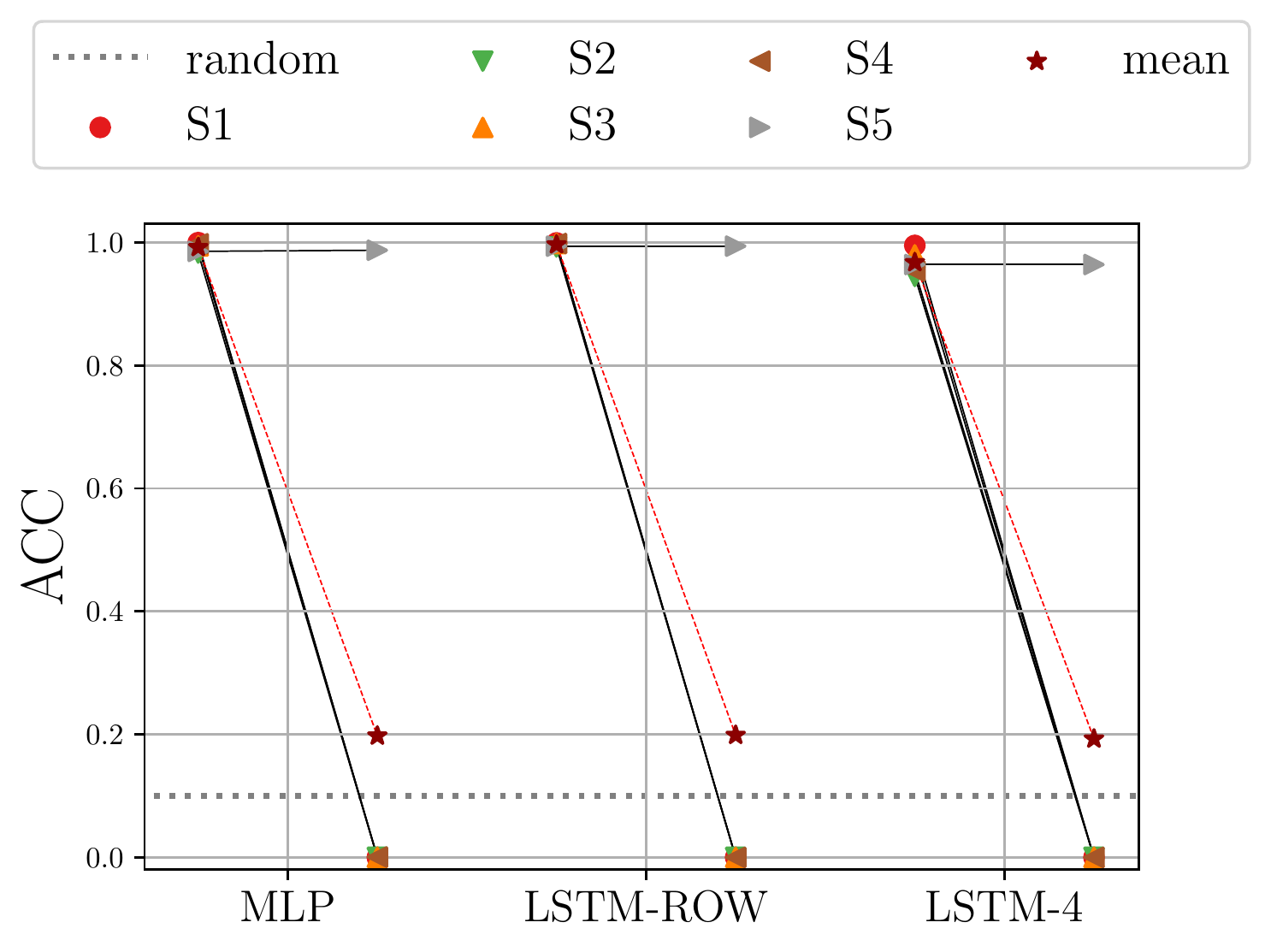}\caption{SMNIST + mas}\end{subfigure}
\begin{subfigure}[t]{0.4\textwidth}\centering\includegraphics[width=\textwidth]{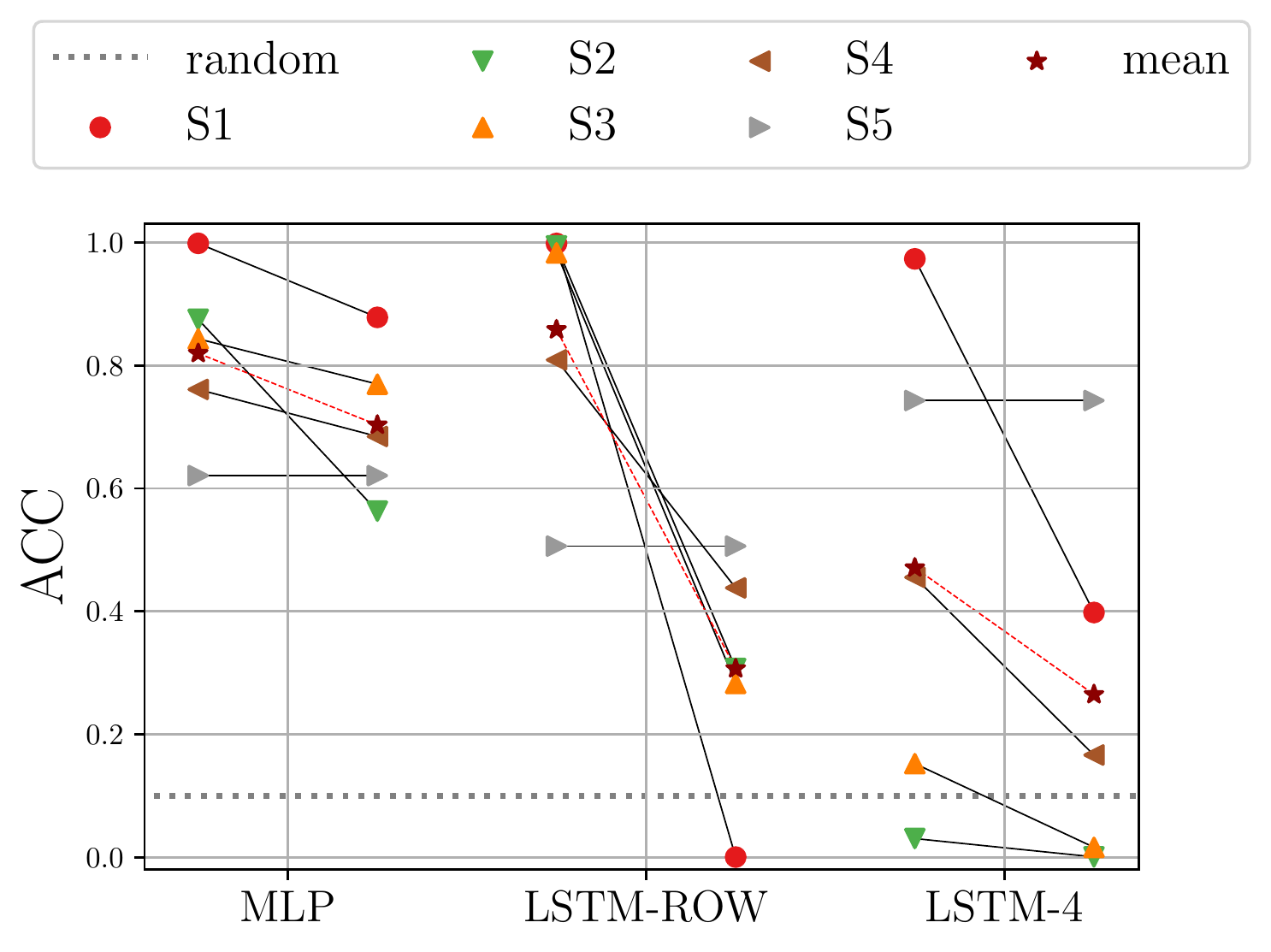}\caption{SMNIST + lwf}\end{subfigure}
\begin{subfigure}[t]{0.4\textwidth}\centering\includegraphics[width=\textwidth]{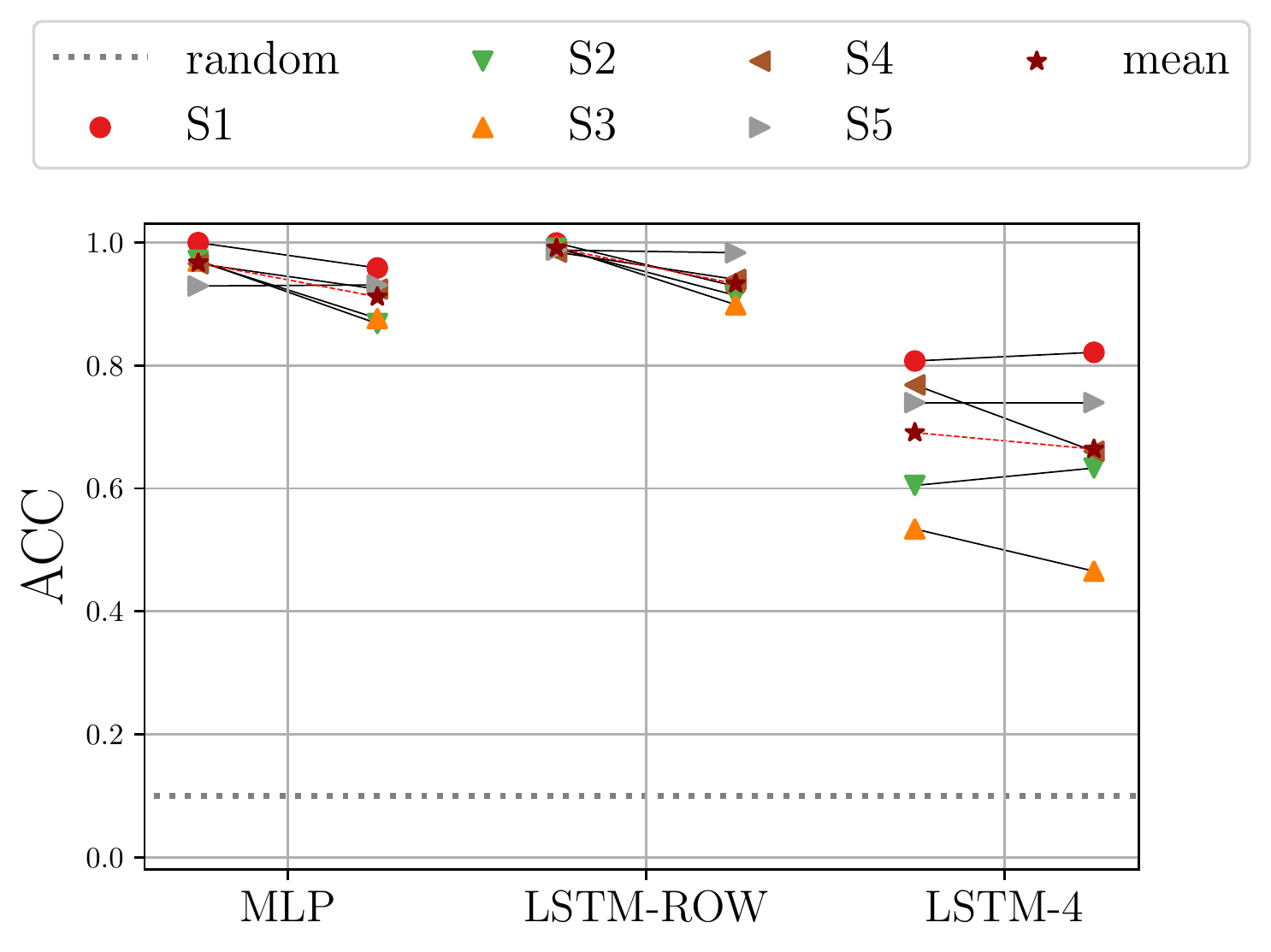}\caption{SMNIST + gem}\end{subfigure}
\begin{subfigure}[t]{0.4\textwidth}\centering\includegraphics[width=\textwidth]{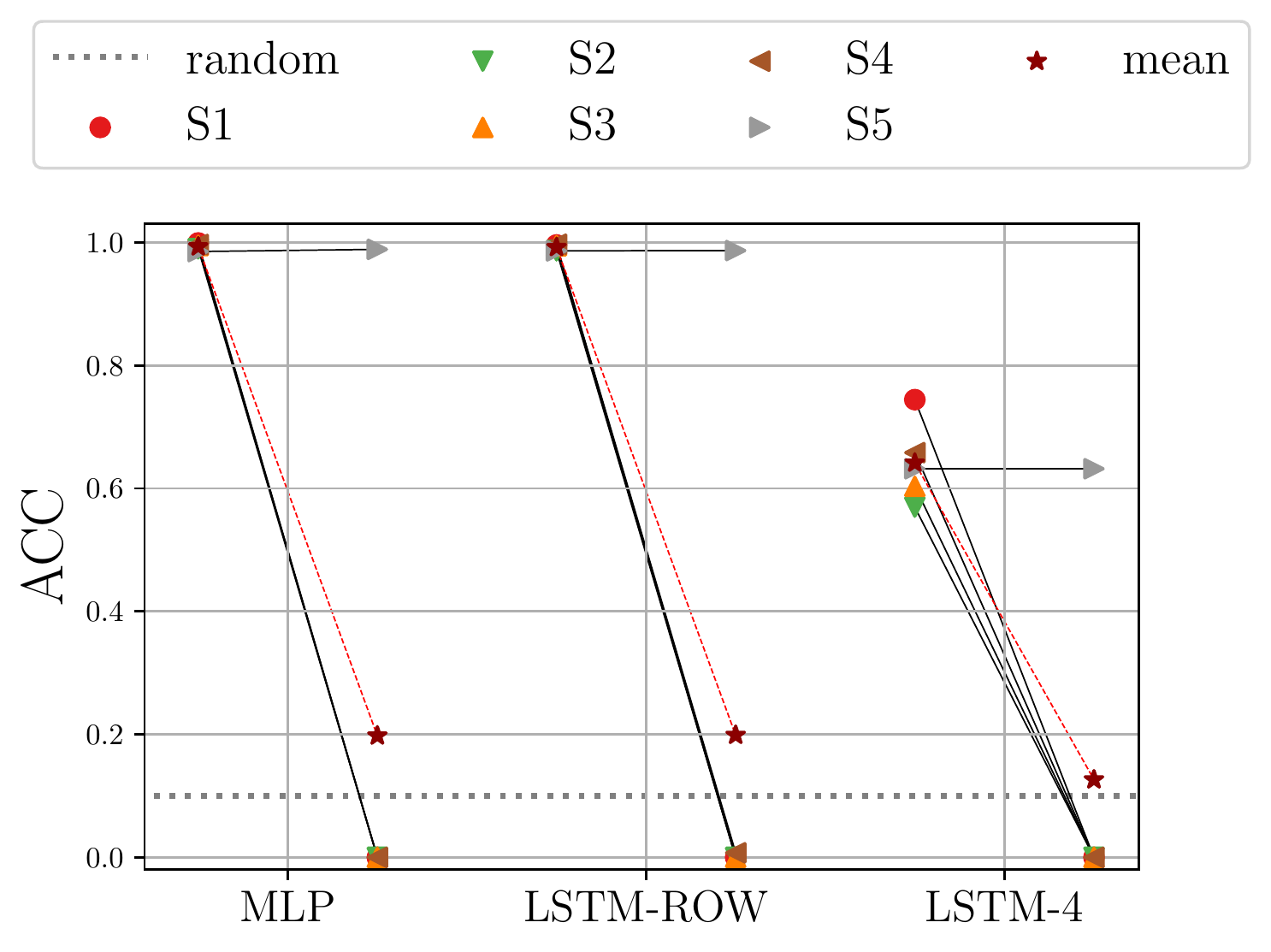}\caption{SMNIST + agem}\end{subfigure}
\begin{subfigure}[t]{0.4\textwidth}\centering\includegraphics[width=\textwidth]{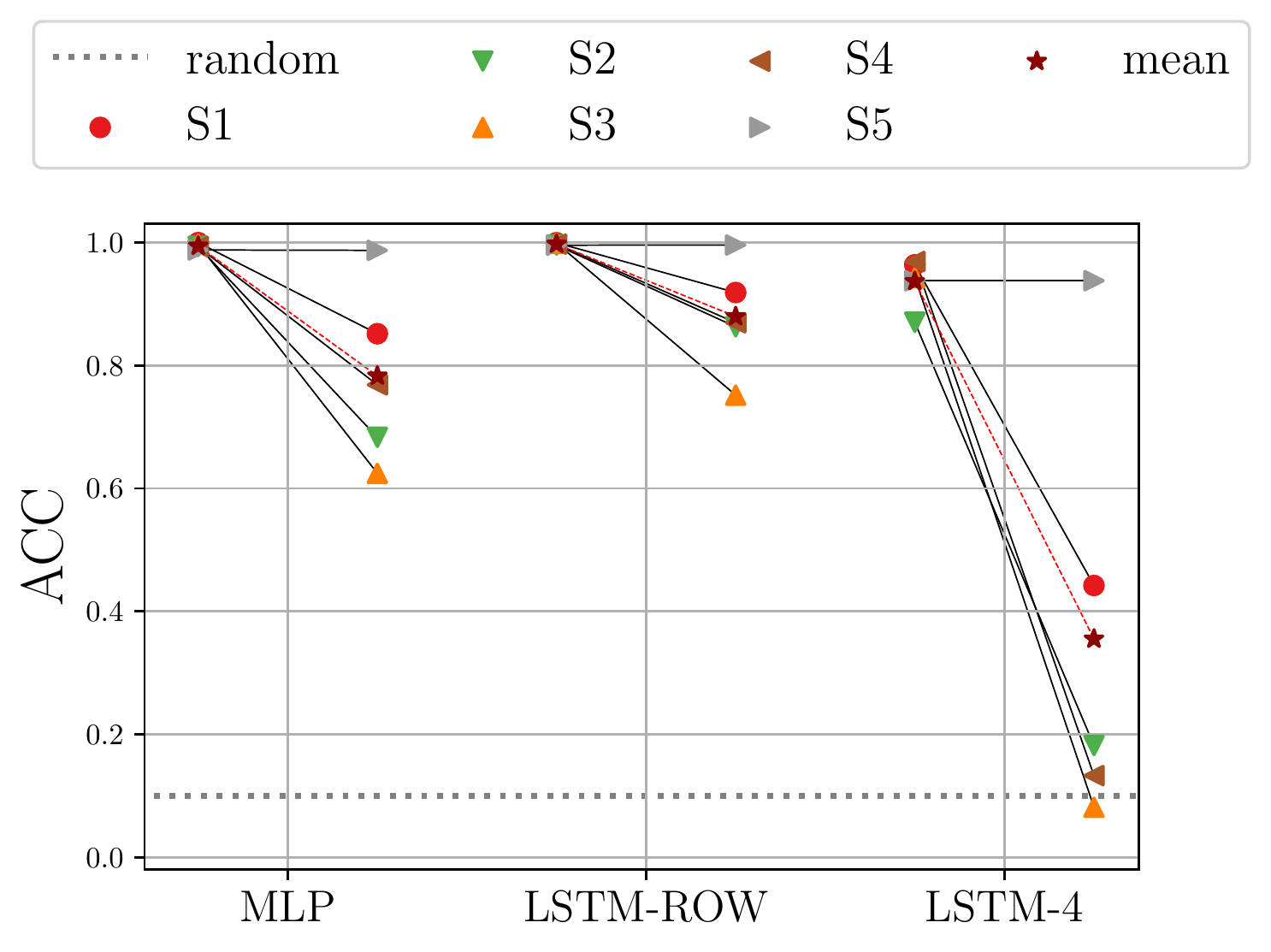}\caption{SMNIST + replay-1}\end{subfigure}
\begin{subfigure}[t]{0.4\textwidth}\centering\includegraphics[width=\textwidth]{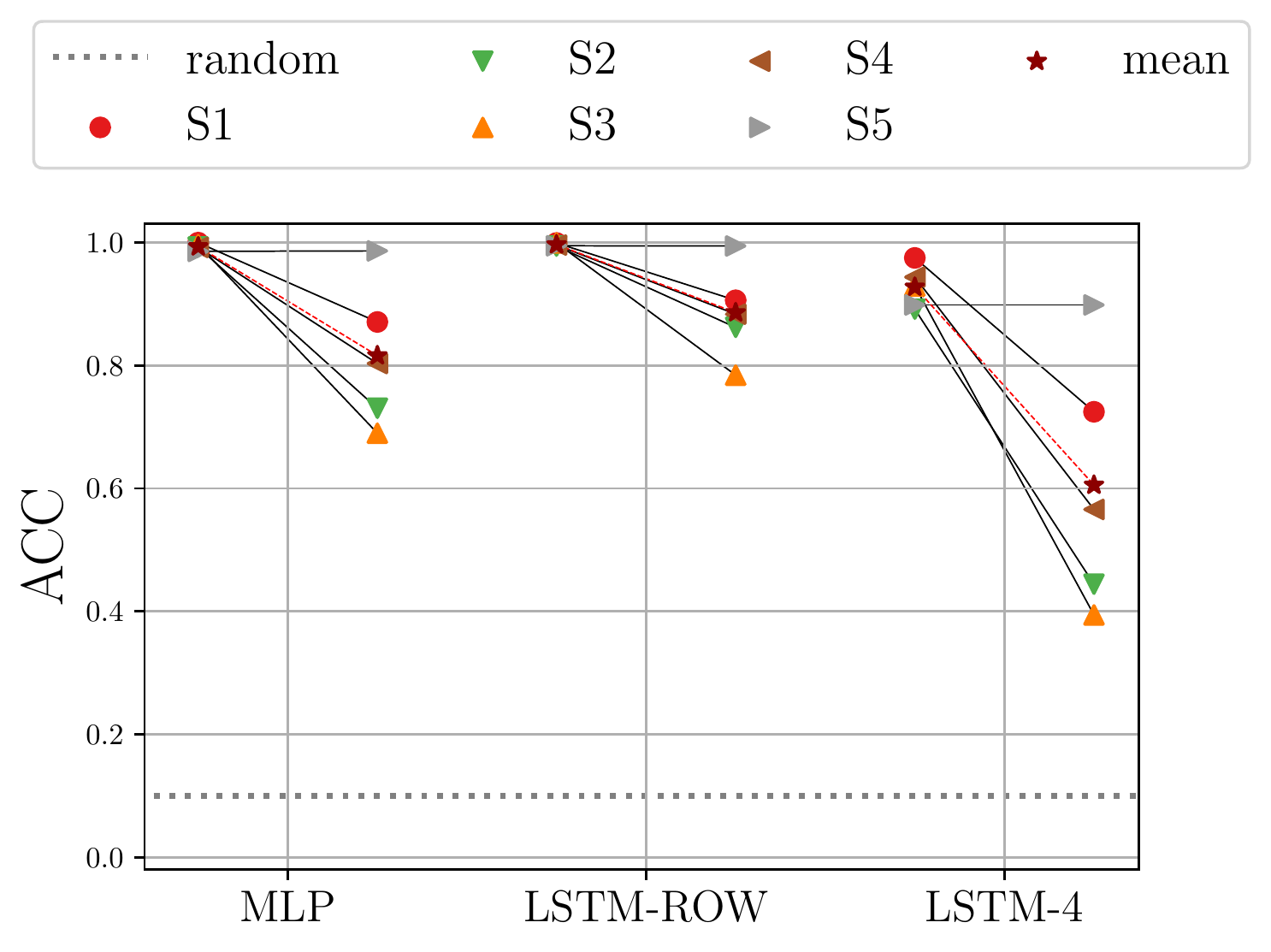}\caption{SMNIST + replay-10}\end{subfigure}
\end{figure}
\begin{figure}
\centering
\ContinuedFloat
\begin{subfigure}[t]{0.4\textwidth}\centering\includegraphics[width=\textwidth]{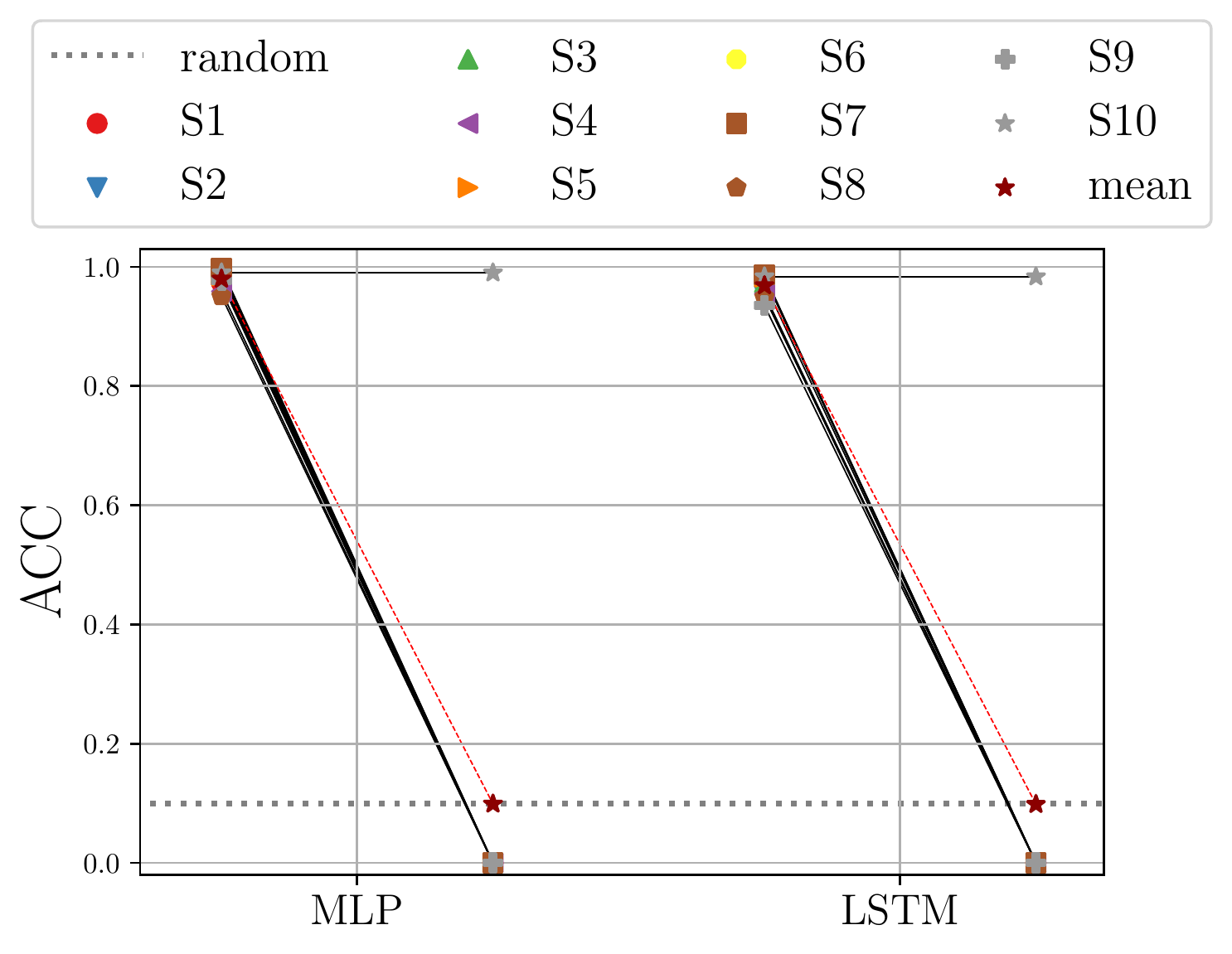}\caption{SSC + naive}\end{subfigure}
\begin{subfigure}[t]{0.4\textwidth}\centering\includegraphics[width=\textwidth]{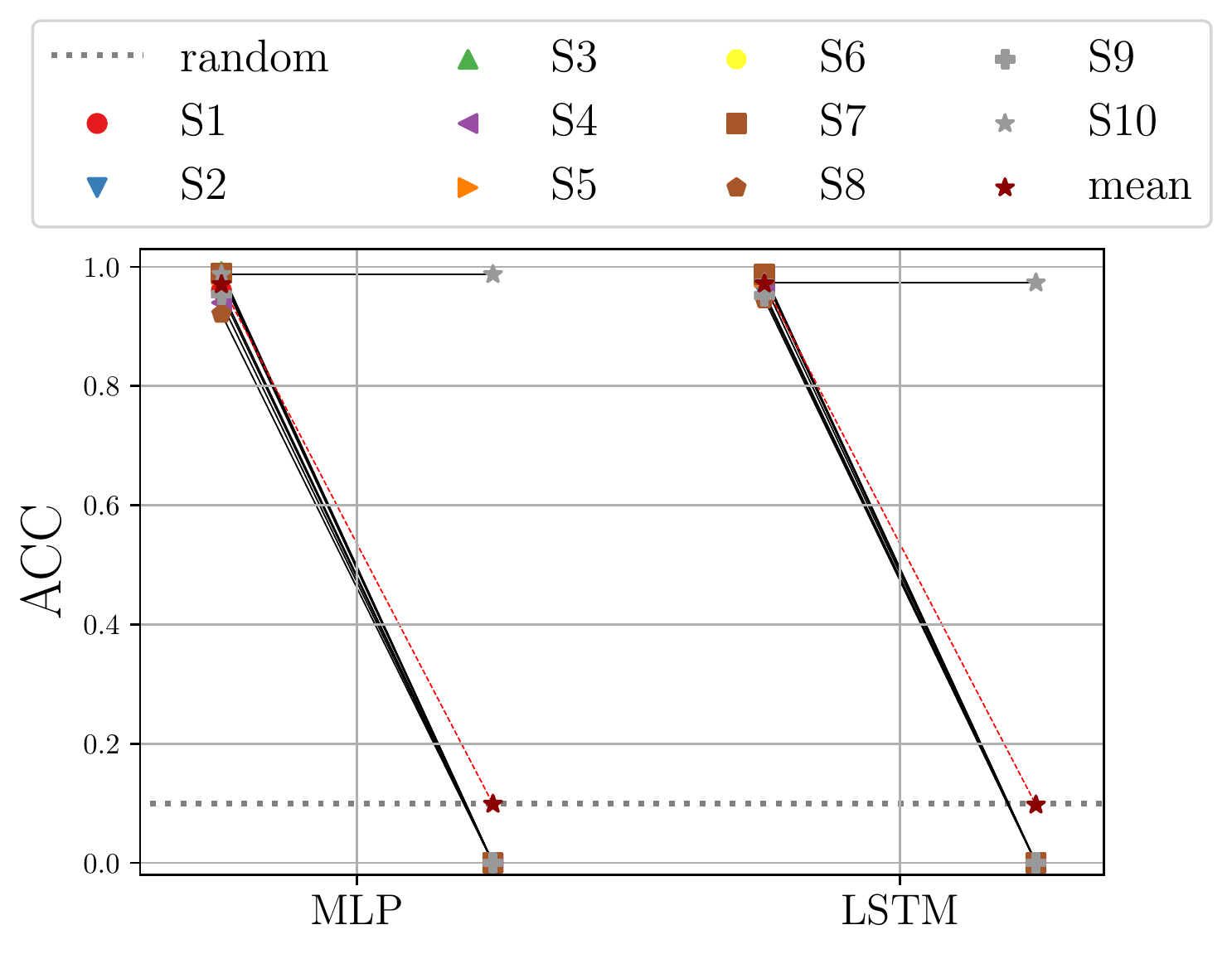}\caption{SSC + ewc}\end{subfigure}
\begin{subfigure}[t]{0.4\textwidth}\centering\includegraphics[width=\textwidth]{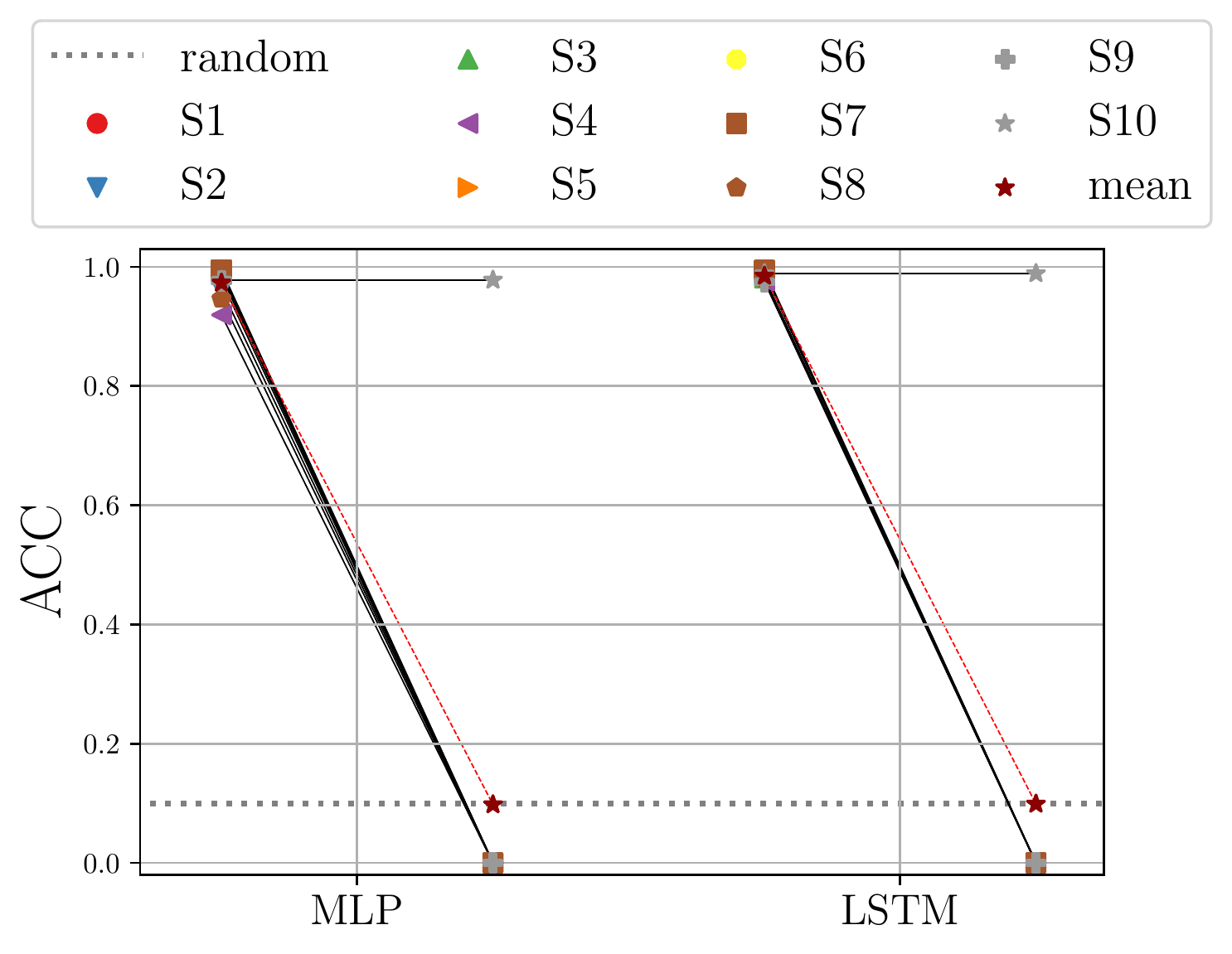}\caption{SSC + mas}\end{subfigure}
\begin{subfigure}[t]{0.4\textwidth}\centering\includegraphics[width=\textwidth]{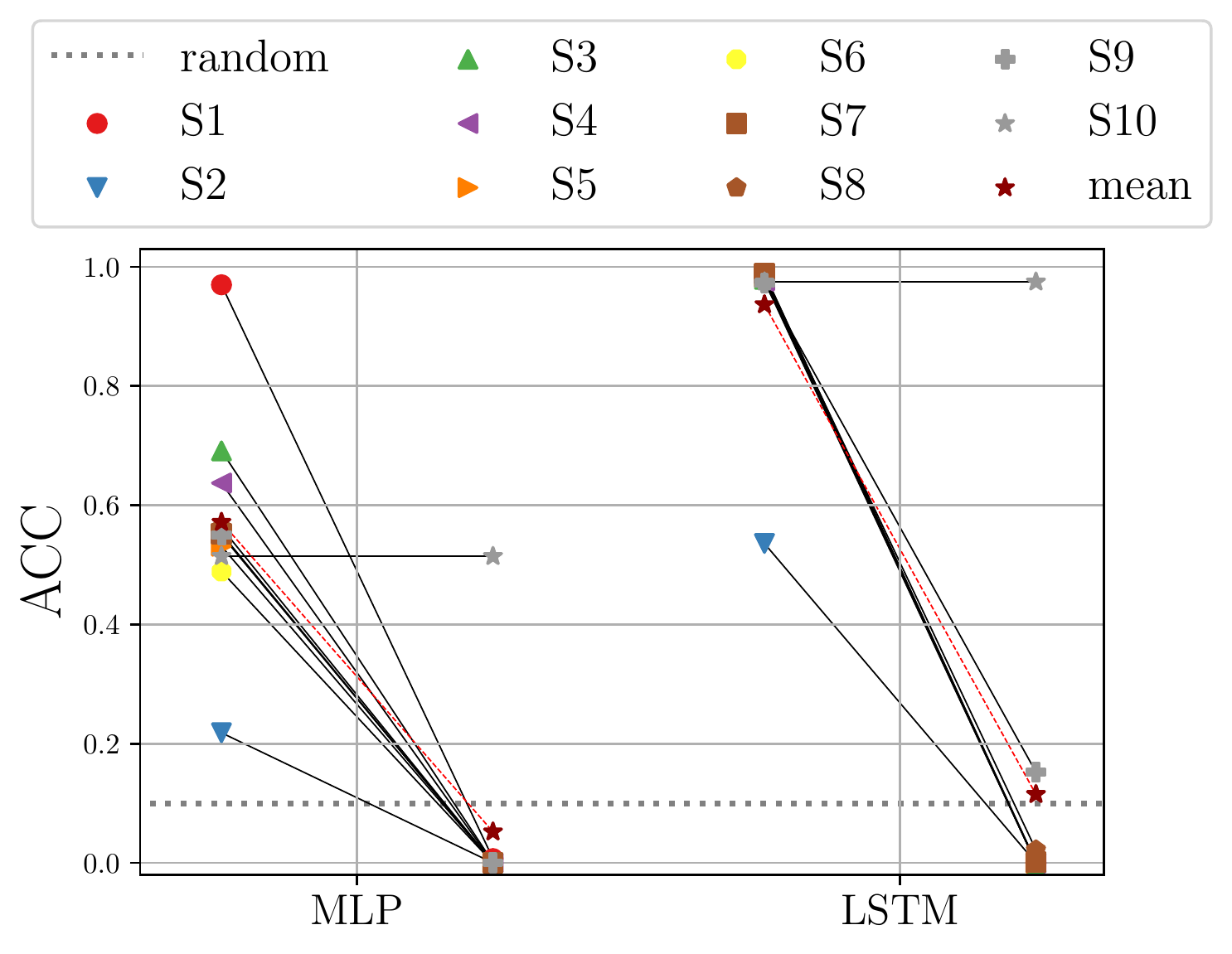}\caption{SSC + lwf}\end{subfigure}
\begin{subfigure}[t]{0.4\textwidth}\centering\includegraphics[width=\textwidth]{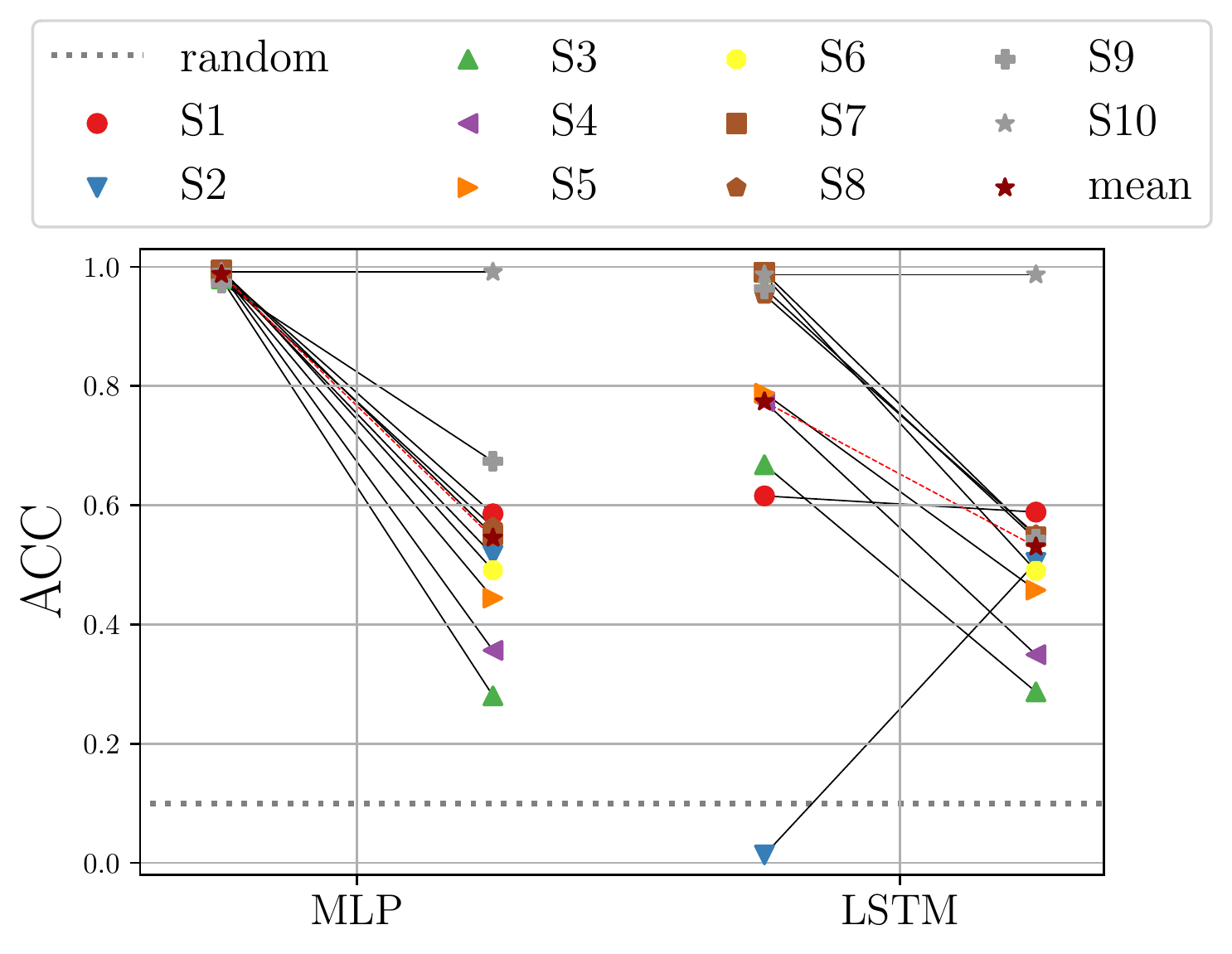}\caption{SSC + gem}\end{subfigure}
\begin{subfigure}[t]{0.4\textwidth}\centering\includegraphics[width=\textwidth]{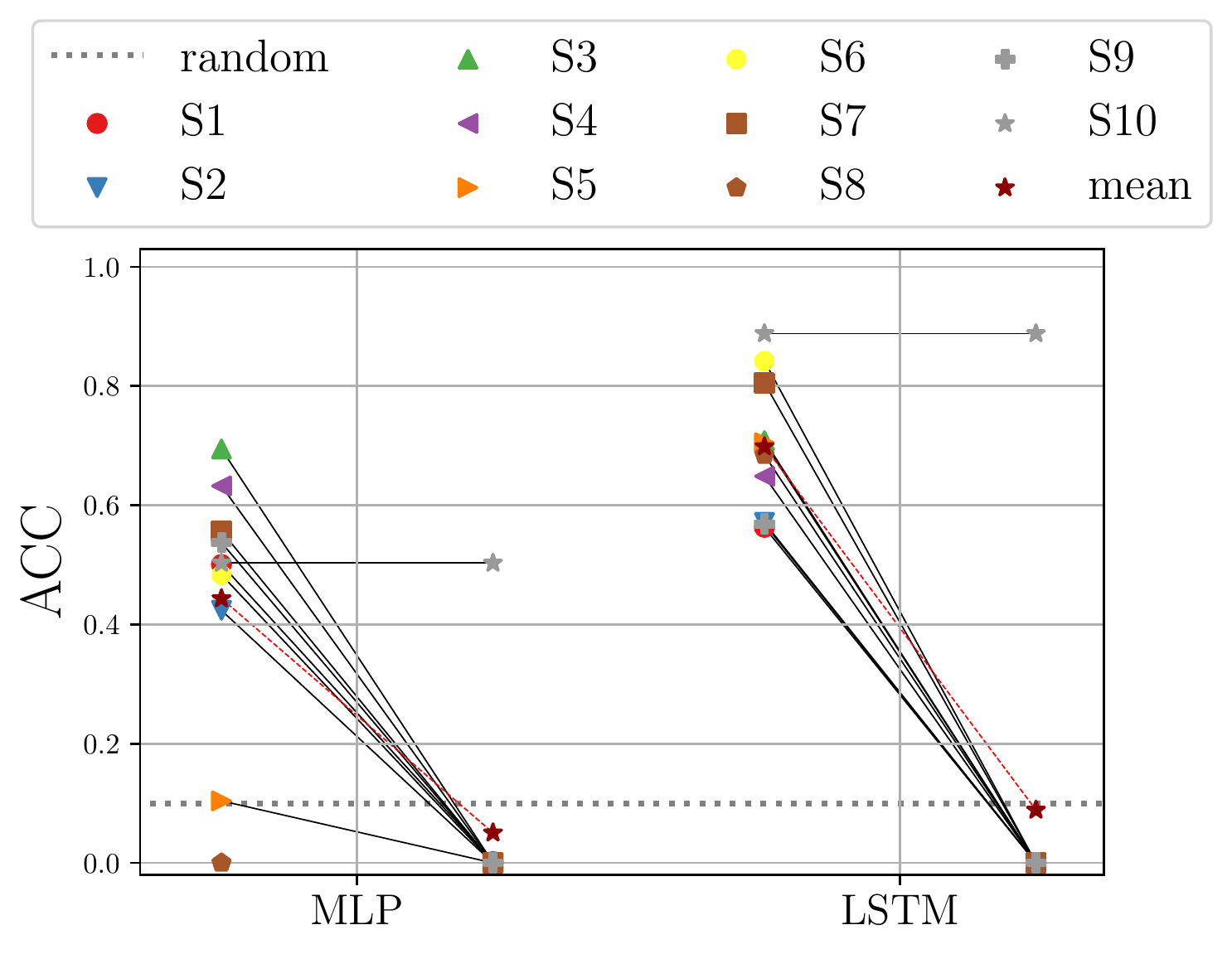}\caption{SSC + agem}\end{subfigure}
\begin{subfigure}[t]{0.4\textwidth}\centering\includegraphics[width=\textwidth]{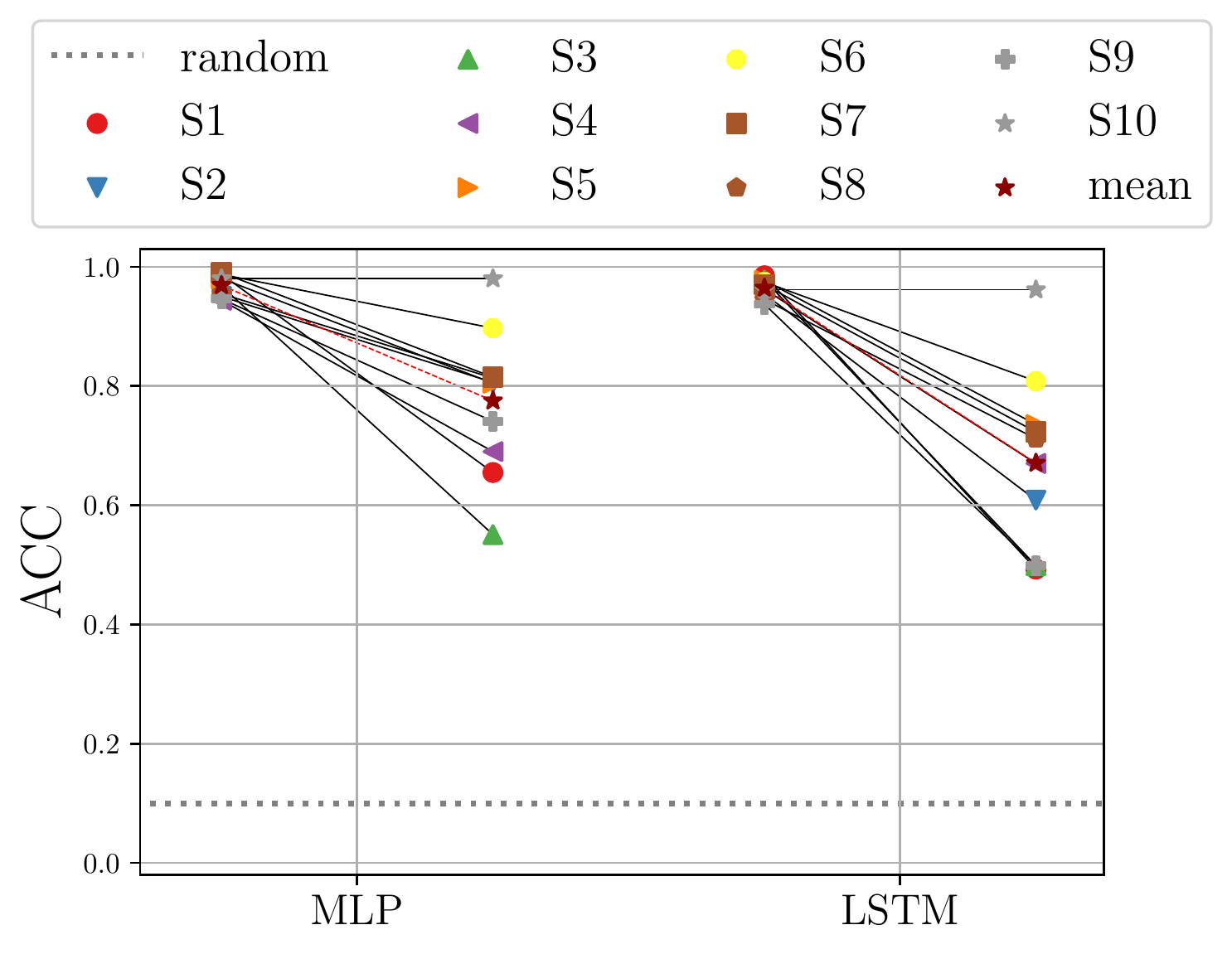}\caption{SSC + replay-1}\end{subfigure}
\begin{subfigure}[t]{0.4\textwidth}\centering\includegraphics[width=\textwidth]{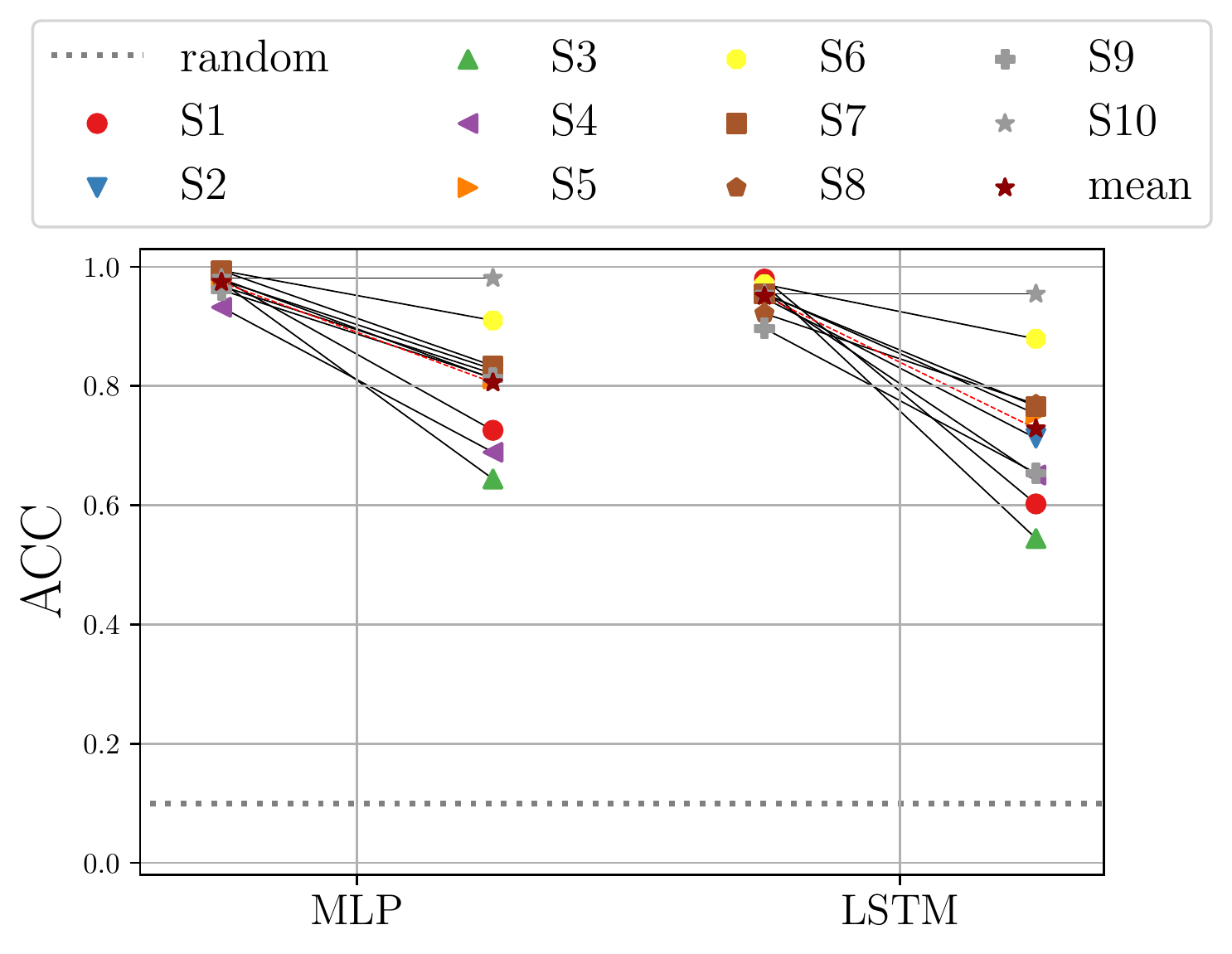}\caption{SSC + replay-10}\end{subfigure}
\end{figure}
\begin{figure}
\centering
\ContinuedFloat
\begin{subfigure}[t]{0.4\textwidth}\centering\includegraphics[width=\textwidth]{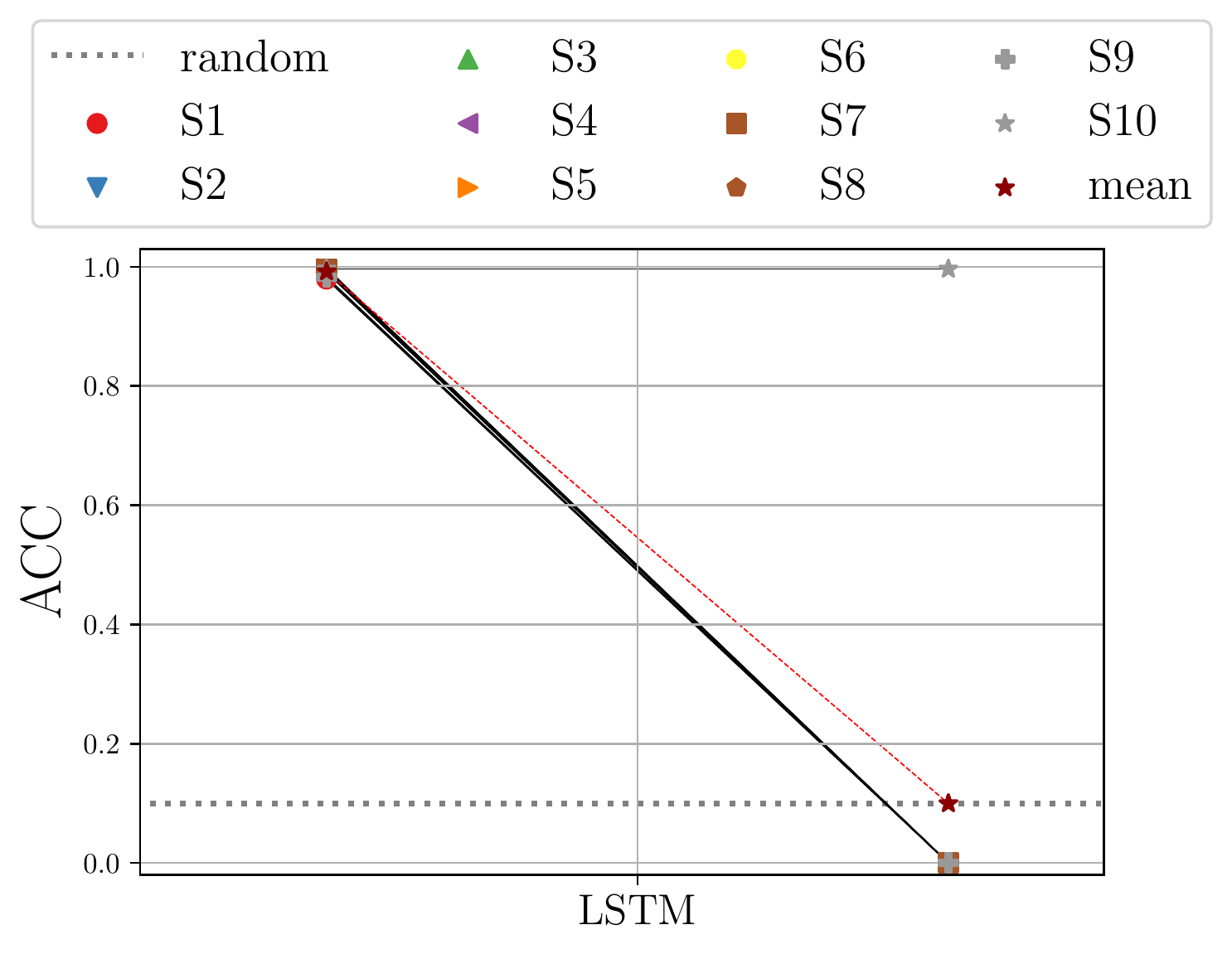}\caption{QD + naive}\end{subfigure}
\begin{subfigure}[t]{0.4\textwidth}\centering\includegraphics[width=\textwidth]{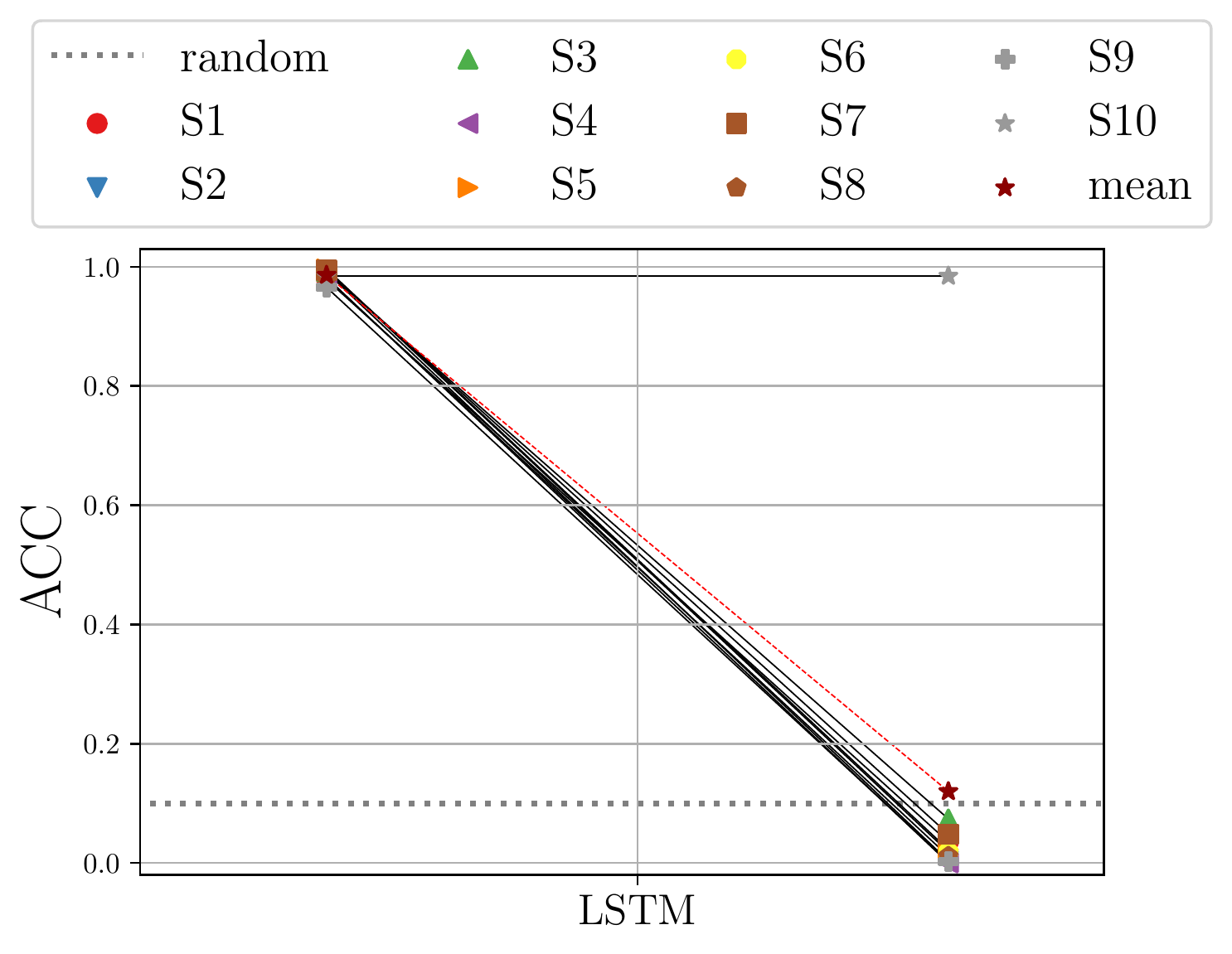}\caption{QD + ewc}\end{subfigure}
\begin{subfigure}[t]{0.4\textwidth}\centering\includegraphics[width=\textwidth]{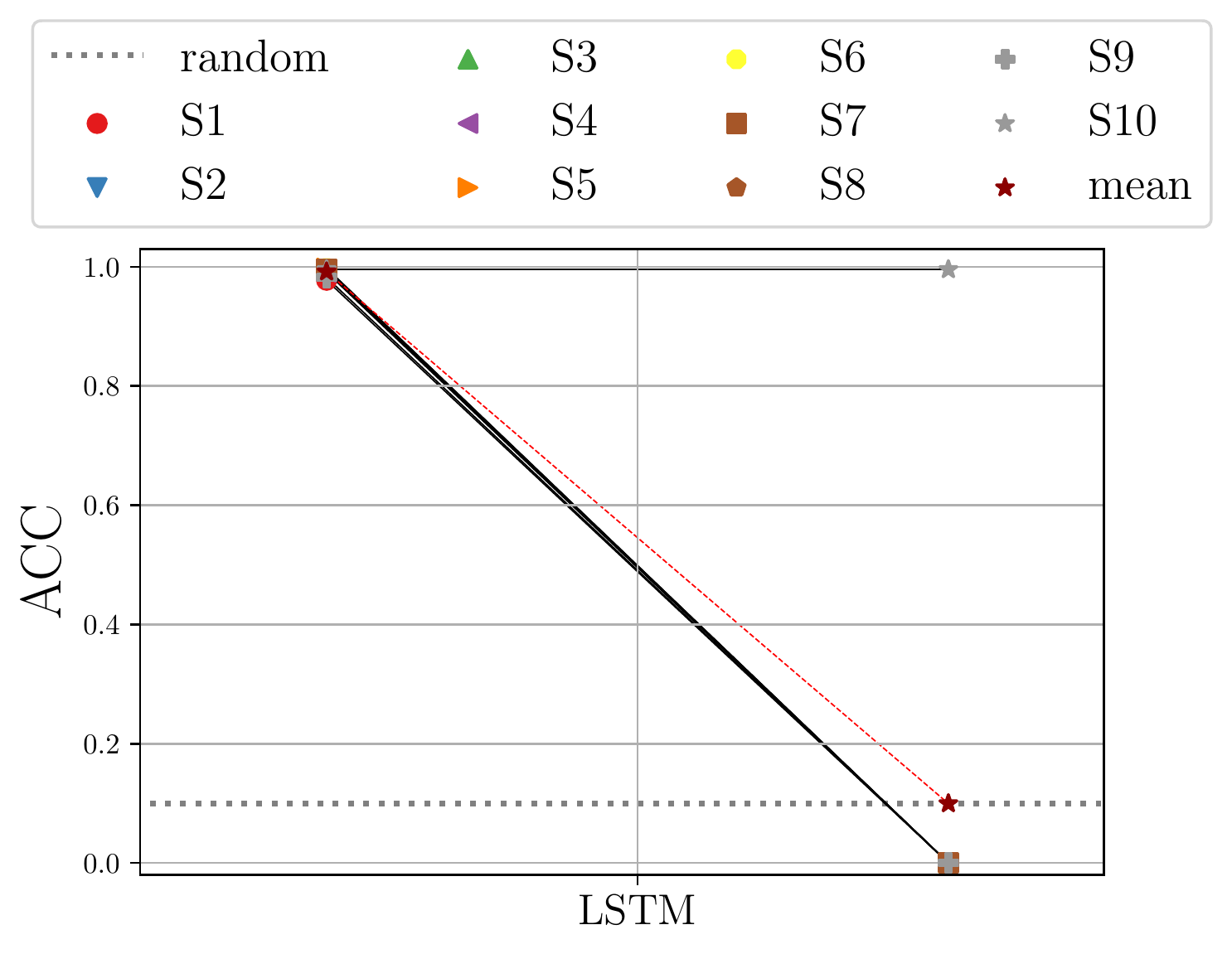}\caption{QD + mas}\end{subfigure}
\begin{subfigure}[t]{0.4\textwidth}\centering\includegraphics[width=\textwidth]{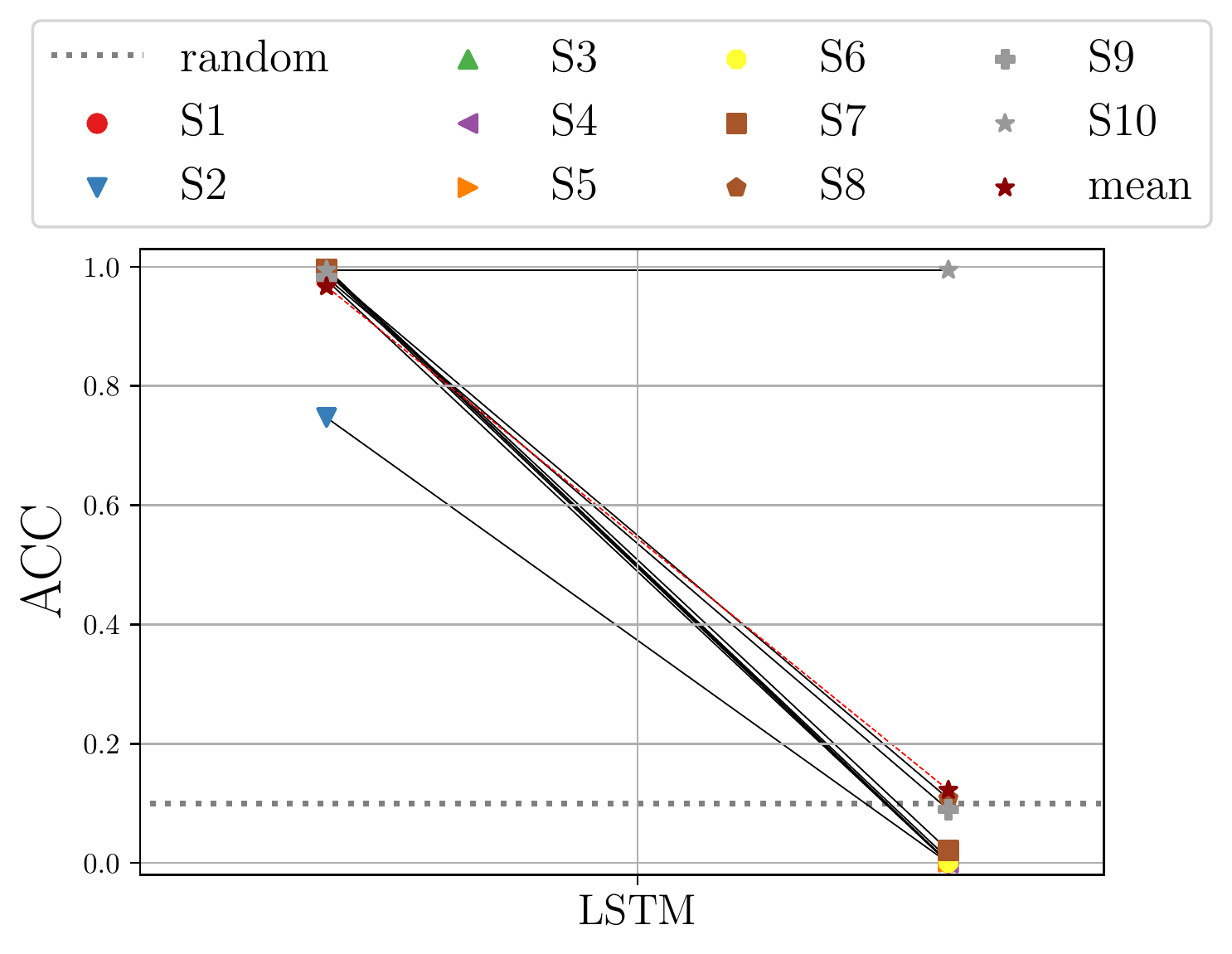}\caption{QD + lwf}\end{subfigure}
\begin{subfigure}[t]{0.4\textwidth}\centering\includegraphics[width=\textwidth]{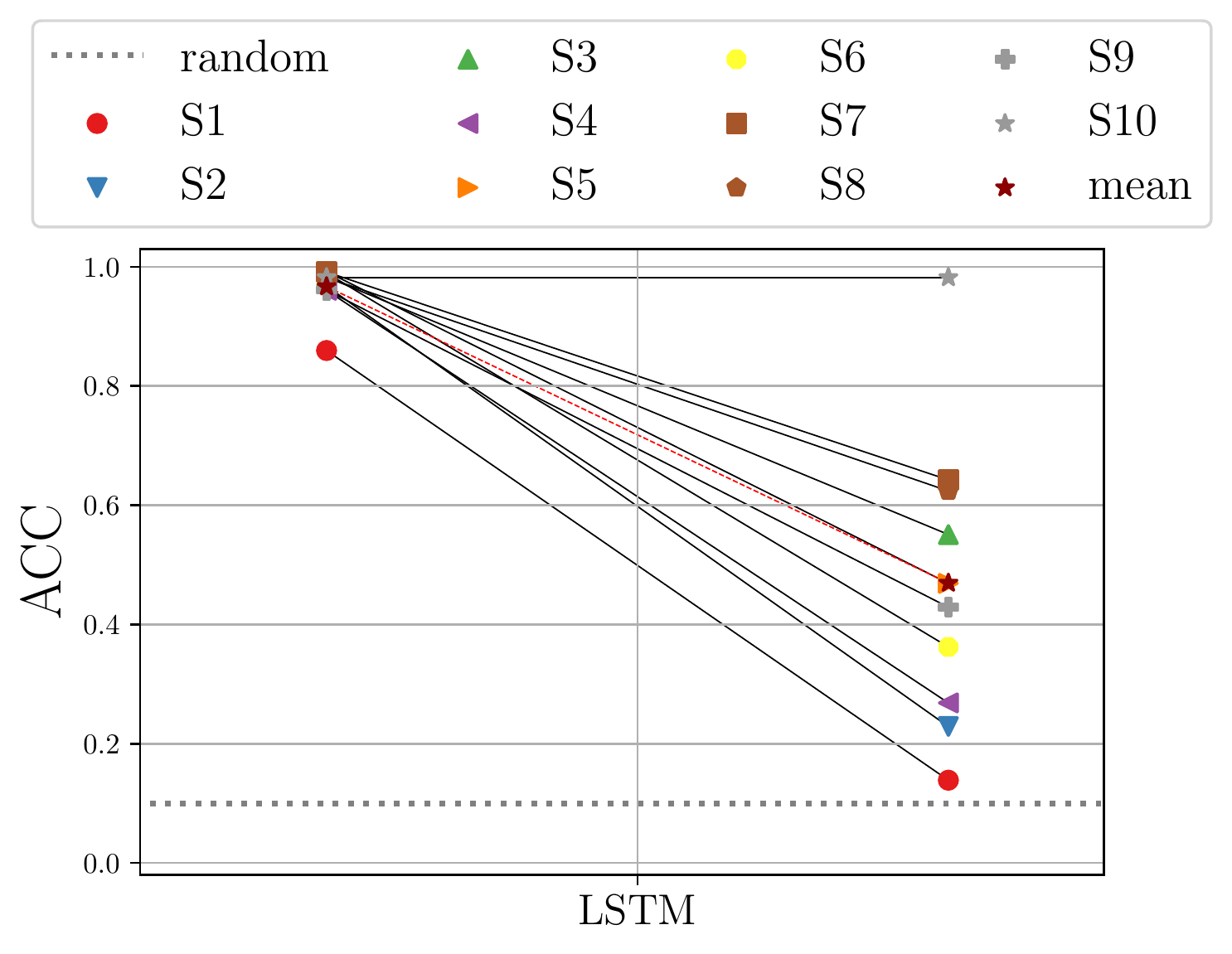}\caption{QD + gem}\end{subfigure}
\begin{subfigure}[t]{0.4\textwidth}\centering\includegraphics[width=\textwidth]{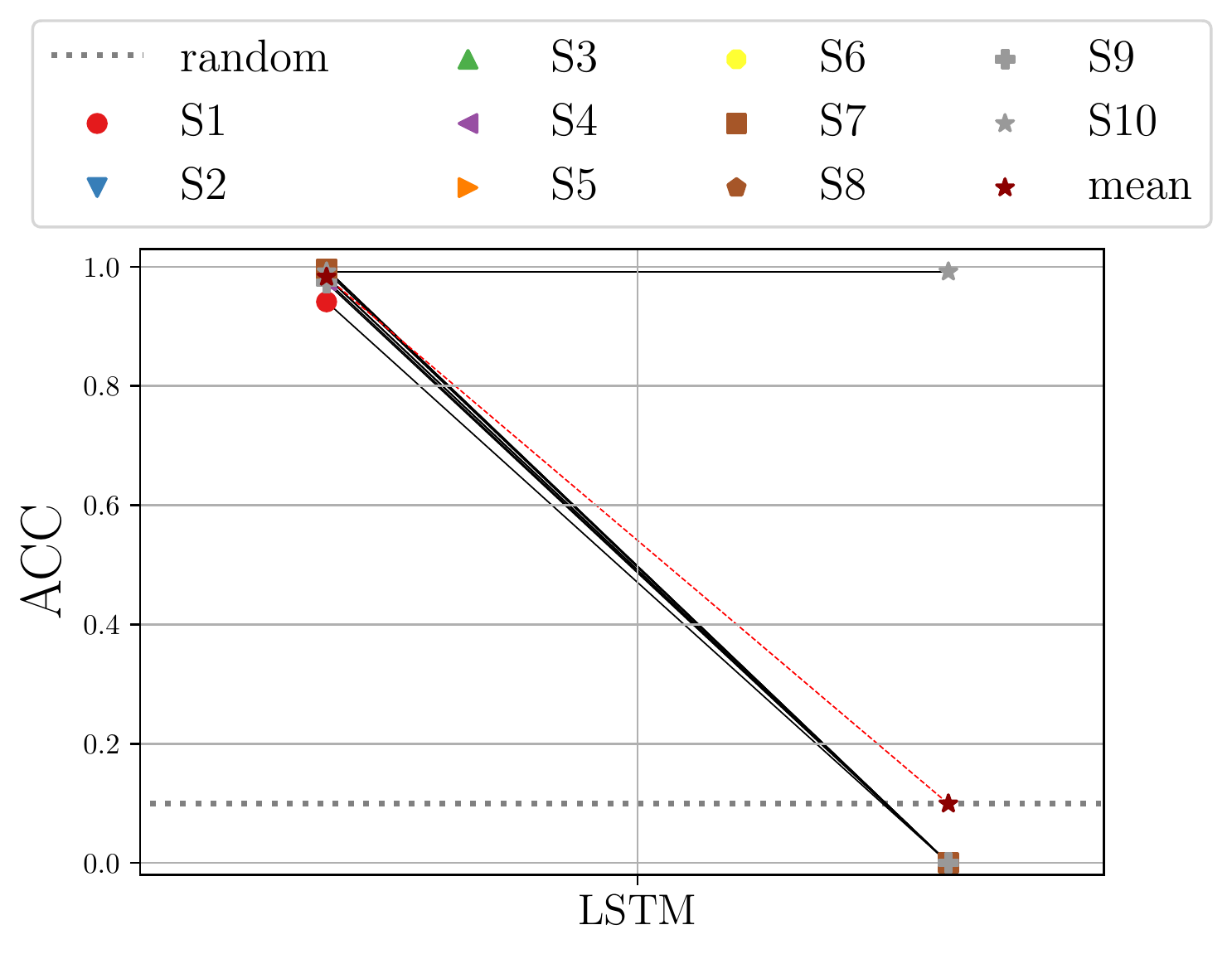}\caption{QD + agem}\end{subfigure}
\begin{subfigure}[t]{0.4\textwidth}\centering\includegraphics[width=\textwidth]{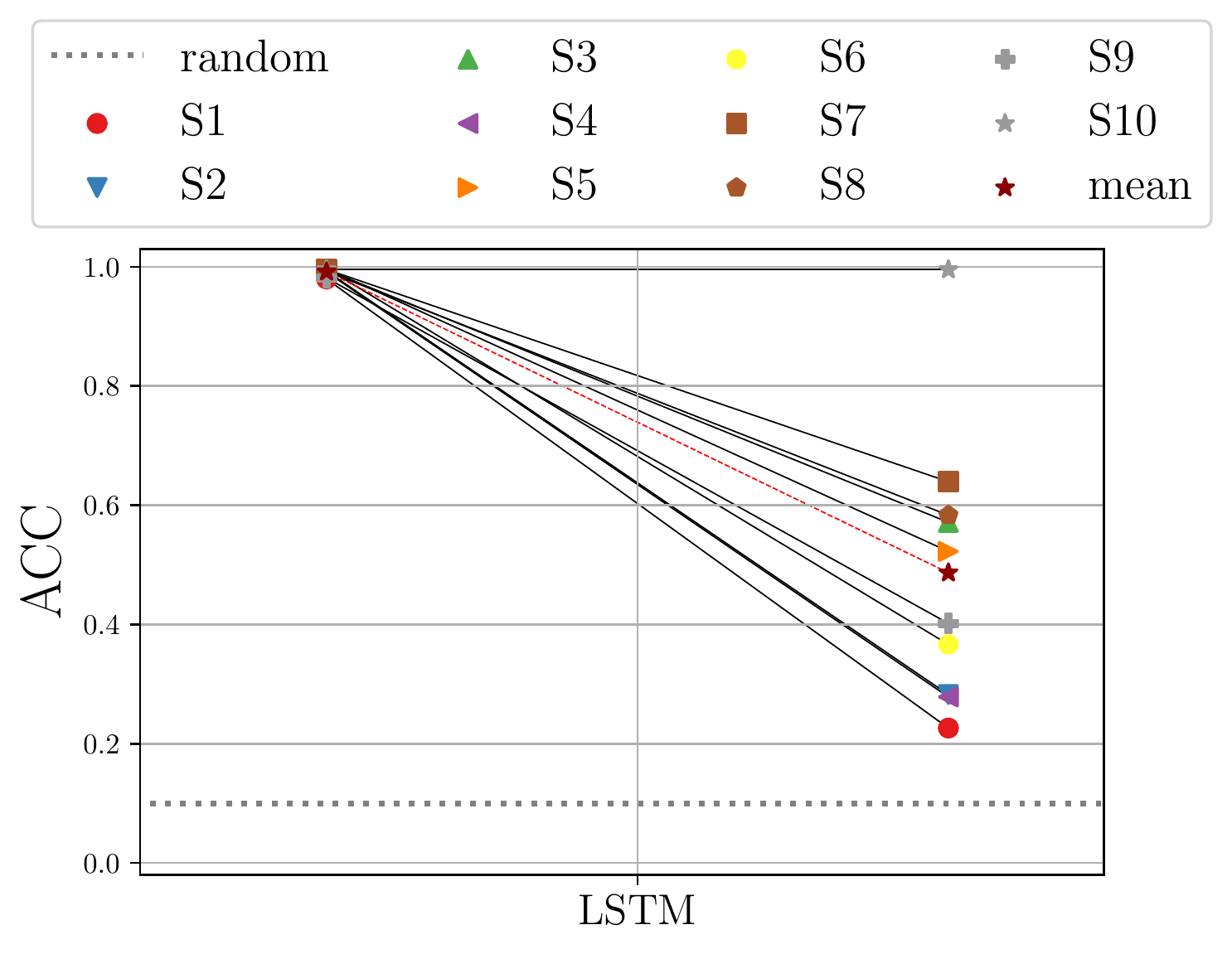}\caption{QD + replay-1}\end{subfigure}
\begin{subfigure}[t]{0.4\textwidth}\centering\includegraphics[width=\textwidth]{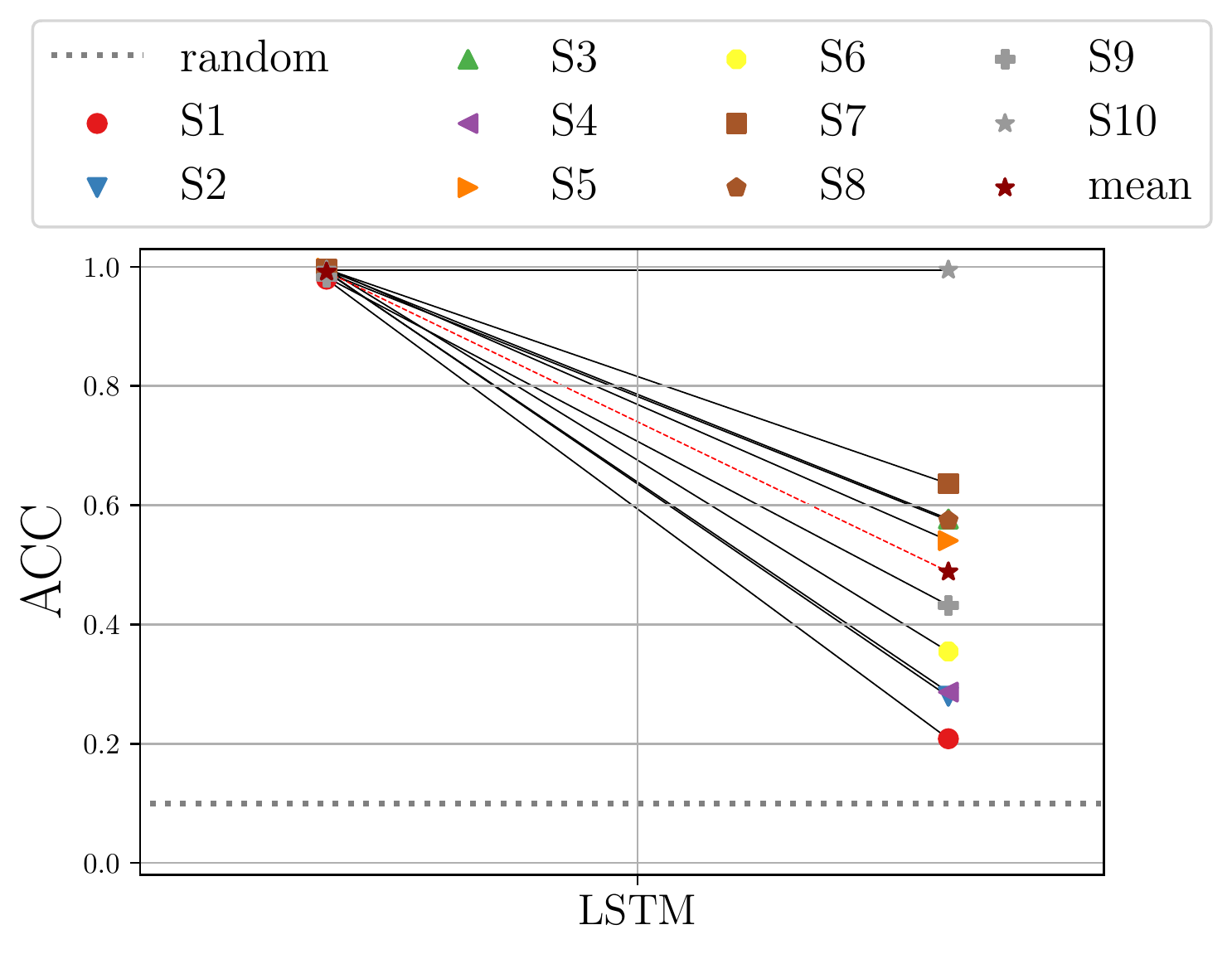}\caption{QD + replay-10}\end{subfigure}
    \caption{Paired plots comparing accuracy at the end of training on each step (left point) and accuracy after training on all steps (right point). Each column represents a different model. Red asterisk represents mean performance across all steps. Horizontal dotted line represents the performance of a random classifier. The more vertical the line, the larger the forgetting.  Best viewed in color.}
 \label{fig:pairedplots}
 \end{figure}
 
\end{document}